\documentclass[lettersize,journal]{style/IEEEtran}
\usepackage{amsmath,amsfonts}
\usepackage{algorithmic}
\usepackage{algorithm}
\usepackage{array}
\ifCLASSOPTIONcompsoc
    \usepackage[caption=false, font=normalsize, labelfont=sf, textfont=sf]{subfig}
\else
\usepackage[caption=false, font=footnotesize]{subfig}
\fi
\usepackage{textcomp}
\usepackage{stfloats}
\usepackage{url}
\usepackage{verbatim}
\usepackage{graphicx}
\usepackage{cite}
\usepackage[nameinlink, capitalize]{cleveref}
\usepackage{color}
\definecolor{CommentPink}{rgb}{1,0.2,0.5}
\definecolor{CommentBlue}{rgb}{0,0,1}
\definecolor{CommentGreen}{rgb}{0,1,0}

\usepackage{bm}
\usepackage{booktabs}

\Crefname{section}{Sec.}{Sec.}
\Crefname{equation}{Eq.}{Eq.}

\DeclareMathOperator*{\argmax}{argmax}

\usepackage[printonlyused,withpage,nolist,nohyperlinks]{acronym}

\begin{acronym}

\acro{2D}{two-dimensional}

\acro{3D}{three-dimensional}

\acro{AHRS}{attitude and heading reference system}

\acro{AUV}{autonomous underwater vehicle}

\acro{CPP}{Chinese Postman Problem}

\acro{DoF}{degree of freedom}
\acrodefplural{DoF}[DoFs]{degrees of freedom}

\acro{DVL}{Doppler velocity log}

\acro{FSM}{finite state machine}

\acro{IMU}{inertial measurement unit}

\acro{LBL}{Long Baseline}

\acro{MCM}{mine countermeasures}

\acro{MDP}{Markov decision process}
\acrodefplural{MDP}[MDPs]{Markov decision processes}

\acro{POMDP}{Partially Observable Markov Decision Process}
\acrodefplural{POMDP}[POMDPs]{Partially Observable Markov Decision Processes}

\acro{PRM}{Probabilistic Roadmap}
\acrodefplural{PRM}[PRM]{Probabilistic Roadmaps}

\acro{ROI}{region of interest}
\acrodefplural{ROI}[ROIs]{regions of interest}

\acro{ROS}{Robot Operating System}

\acro{ROV}{remotely operated vehicle}

\acro{RRT}{Rapidly-exploring Random Tree}
\acrodefplural{RRT}[RRTs]{Rapidly-exploring Random Trees}

\acro{SLAM}{Simultaneous Localisation and Mapping}

\acro{SSE}{sum of squared errors}

\acro{STOMP}{Stochastic Trajectory Optimization for Motion Planning}

\acro{TRN}{Terrain-Relative Navigation}

\acro{UAV}{unmanned aerial vehicle}

\acro{USBL}{Ultra-Short Baseline}

\acro{IPP}{informative path planning}

\acro{FoV}{field of view}
\acrodefplural{FoV}[FoVs]{fields of view}

\acro{CDF}{cumulative distribution function}

\acro{ML}{maximum likelihood}

\acro{RMSE}{Root Mean Squared Error}
\acro{MLL}{Mean Log Loss}

\acro{GP}{Gaussian Process}
\acrodefplural{GP}[GPs]{Gaussian Processes}

\acro{KF}{Kalman Filter}

\acro{IP}{Interior Point}
\acro{BO}{Bayesian Optimization}

\acro{SE}{squared exponential}

\acro{UI}{uncertain input}

\acro{MCL}{Monte Carlo Localisation}
\acro{AMCL}{Adaptive Monte Carlo Localisation}

\acro{SSIM}{Structural Similarity Index}

\acro{MAE}{Mean Absolute Error}
\acro{RMSE}{Root Mean Squared Error}
\acro{AUSE}{Area Under the Sparsification Error curve}

\acro{AL}{active learning}
\acro{DL}{deep learning}
\acro{CNN}{convolutional neural network}

\acro{MC}{Monte-Carlo}

\acro{GSD}{ground sample distance}

\acro{BALD}{Bayesian active learning by disagreement}

\acro{fCNN}{fully convolutional neural network}
\acro{FCN}{fully convolutional neural network}

\acro{CMA-ES}{covariance matrix adaptation evolution strategy}

\acro{FoV}{field of view}

\acro{mIoU}{mean Intersection-over-Union}
\acro{ECE}{expected calibration error}

\acro{GAN}{Generative Adversarial Network}
\acro{POMCP}{Partially Observable Monte-Carlo Planning}
\acro{MCTS}{Monte-Carlo tree search}
\acro{RL}{reinforcement learning}

\end{acronym}

\usepackage[numbers,sectionbib,sort&compress]{natbib}

\begin{document}

\title{An Informative Path Planning Framework for \\ Active Learning in UAV-based Semantic Mapping}

\author{Julius R\"{u}ckin, Federico Magistri, Cyrill Stachniss, Marija Popovi\'{c}
\thanks{This work has been funded by the Deutsche Forschungsgemeinschaft (DFG, German Research Foundation) under Germany's Excellence Strategy, EXC-2070 -- 390732324 (PhenoRob). All authors are with the Cluster of Excellence PhenoRob, Institute of Geodesy and Geoinformation, University of Bonn. Cyrill Stachniss is also with the University of Oxford and Lamarr Institute for Machine Learning and Artificial Intelligence, Germany.
Corresponding: \texttt{jrueckin@uni-bonn.de}.}
}

\markboth{ACCEPTED FOR PUBLICATION IN IEEE TRANSACTIONS ON ROBOTICS, AUGUST 23, 2023}{}

\maketitle

\begin{abstract}
Unmanned aerial vehicles (UAVs) are frequently used for aerial mapping and general monitoring tasks.
Recent progress in deep learning enabled automated semantic segmentation of imagery to facilitate the interpretation of large-scale complex environments.
Commonly used supervised deep learning for segmentation relies on large amounts of pixel-wise labelled data, which is tedious and costly to annotate.
The domain-specific visual appearance of aerial environments often prevents the usage of models pre-trained on publicly available datasets.
To address this, we propose a novel general planning framework for UAVs to autonomously acquire informative training images for model re-training.
We leverage multiple acquisition functions and fuse them into probabilistic terrain maps.
Our framework combines the mapped acquisition function information into the UAV's planning objectives. 
In this way, the UAV adaptively acquires informative aerial images to be manually labelled for model re-training.
Experimental results on real-world data and in a photorealistic simulation show that our framework maximises model performance and drastically reduces labelling efforts. 
Our map-based planners outperform state-of-the-art local planning.
\end{abstract}

\begin{IEEEkeywords}
Informative Path Planning, Active Learning, Bayesian Deep Learning, Semantic Segmentation and Mapping
\end{IEEEkeywords}

\section{Introduction} \label{S:intro}

\Acp{UAV} enable highly agile, low-cost operations in various aerial imaging applications~\citep{Osco2021,Kellenberger2019}, such as precision agriculture~\citep{popovic2020informative,Rodriguez2021}, wildlife conservation~\citep{Kellenberger2019}, and urban planning~\citep{Potsdam2018,Lenczner2022,kemker2018algorithms,Tuia2009}. Combined with advances in deep learning for semantic segmentation through \acp{FCN}~\citep{long2015fully, ronneberger2015u}, deploying \acp{UAV} accelerates automated scene understanding in large-scale and complex aerial environments~\cite{Garcia2017}. Classical deep learning-based semantic segmentation models often used in this context are usually trained on a static curated dataset in a supervised fashion only once before deployment. This leads to two major drawbacks. First, training a semantic segmentation model requires enormous amounts of pixel-wise labelled images, which is a repetitive and time-consuming process often executed by costly domain experts. Second, visual appearance can differ significantly between environments or change over time. Thus, a critical requirement for robot autonomy is the ability to learn about an environment by continuously improving the robot's semantic perception with minimal expert guidance.

In this work, we examine the problem of \ac{AL} in \ac{UAV}-based semantic mapping. Our goal is to improve the robot's vision capabilities in initially unknown environments while minimising the total amount of human-labelled data. To this end, our approach exploits ideas from \ac{AL} research and incorporates them into a new \ac{IPP} framework. The framework replans the \ac{UAV}'s path online as new observations are collected to actively target regions of informative training data. The newly gathered images are labelled by a human annotator and used to re-train an \ac{FCN}, maximising its semantic segmentation performance.

\begin{figure}[!t]
    \captionsetup[subfigure]{labelformat=empty}
    \centering

    \subfloat[]{\includegraphics[width=\columnwidth]{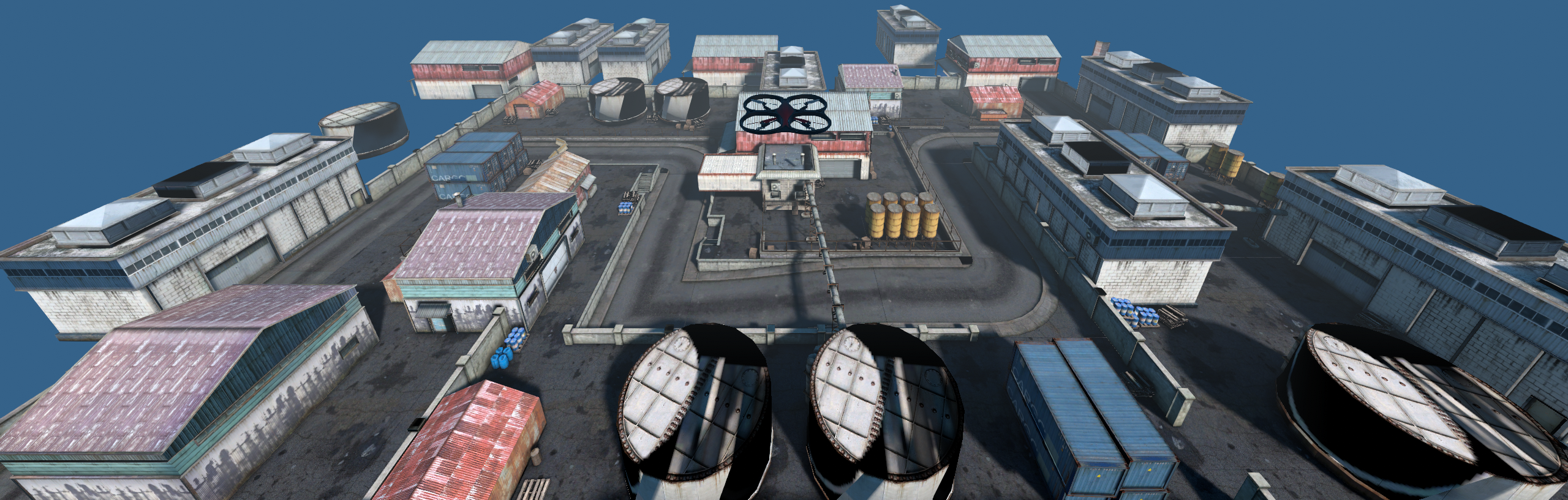}}
    \vspace{-4mm}
    \subfloat[]{\includegraphics[width=\columnwidth]{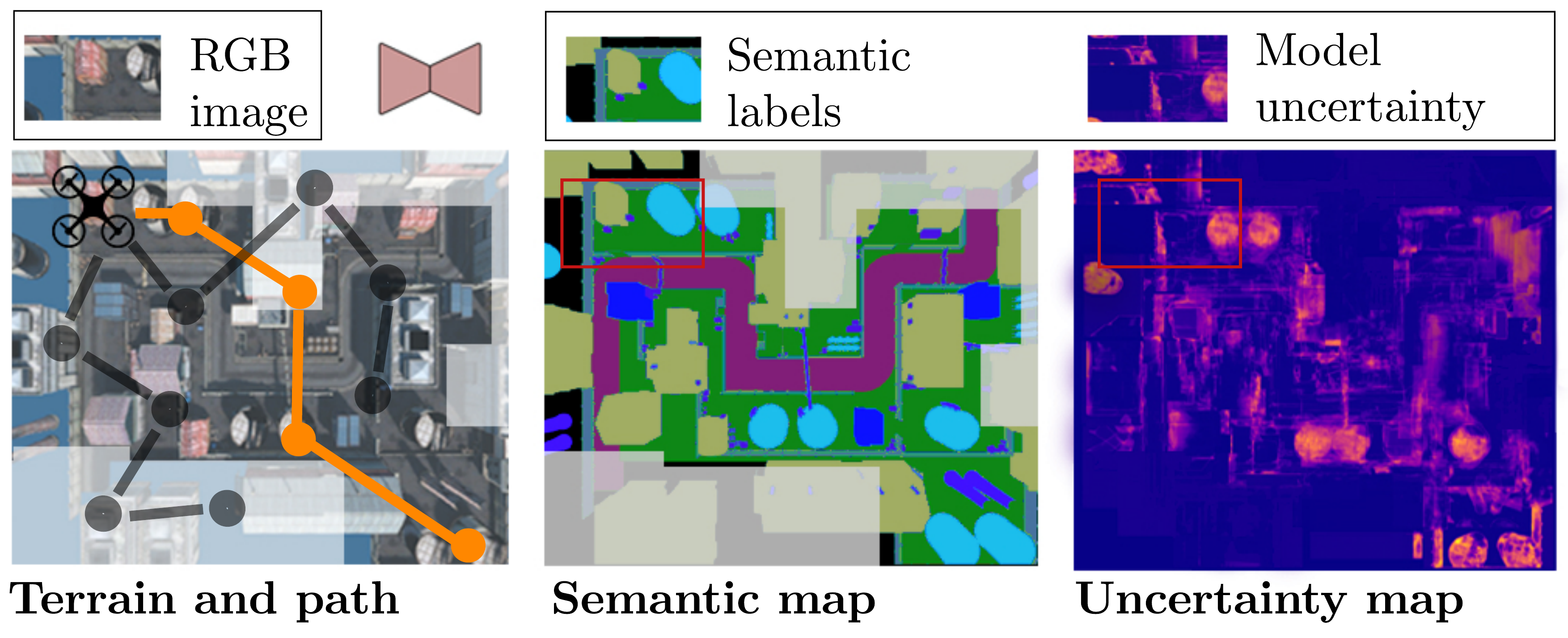}}

    \vspace{-3mm}
    \caption{Our general planning framework for active learning in \ac{UAV}-based semantic mapping deployed in a photo-realistic simulator~\cite{song2020flightmare} (top). We compute an acquisition function, e.g. model uncertainty, and predict semantic segmentation online (centre-right) and fuse both in terrain maps (bottom-right). Our map-based planners replan a \ac{UAV}'s path (orange, bottom-left) to collect the most informative, e.g. most uncertain (yellow), images for network re-training. Our approach reduces the number of images that must be manually labelled to maximise semantic segmentation performance.}
    \label{F:teaser}
\end{figure}

Various \ac{AL} methods for machine learning effectively reduce the requirements for human-labelled training data~\cite{freund1997selective, lewis1994sequential, campbell2000query, tong2001support, kapoor2007active, li2013adaptive, wang2015querying}. Recently, \ac{AL} approaches for deep learning models are gaining attention~\cite{gal2017deep, sener2017active, sinha2019variational, ebrahimi2020minimax, yang2017suggestive, zhdanov2019diverse}. These works develop acquisition functions for selecting to-be-labelled training data to maximise model performance. However, they cannot be directly applied to robotic missions as they assume access to large pre-recorded unlabelled in-domain data pools. An open problem is how to leverage \ac{AL} to improve robot perception with minimal expert guidance when operating in initially unknown environments. More recent \ac{AL} works for aerial imagery consider the \ac{UAV} to be a passive data collection device to record static data pools~\cite{Lenczner2022, Kellenberger2019}. In contrast, we aim to utilise the \ac{UAV}'s decision-making capabilities to improve its perception and, thus, its reasoning about the environment for downstream tasks.

The main contribution of this work is a novel and generally applicable active planning framework linking ideas from \ac{AL} to robotic planning objectives, as illustrated in \cref{F:teaser}. The key benefit of our approach is that it reduces human labelling effort for continuous robotic perception improvement. We exploit various model uncertainty and training data novelty estimation techniques for deep learning models~\cite{gal2016dropout, lakshminarayanan2017simple, postels2021practicality} and apply them to semantic segmentation with a pre-trained \ac{FCN}~\cite{Romera2018}. The inferred pixel-wise semantic labels and estimated model uncertainty and novelty scores are fused sequentially into a probabilistic terrain map as new observations are acquired. As a key feature, our new \ac{IPP} framework iteratively replans the \ac{UAV}'s path to collect the most informative, i.e. the most uncertain or novel, images for labelling and model re-training in a targeted fashion.

This article builds upon our previous conference paper~\cite{ruckin2022informative}. In our previous work~\cite{ruckin2022informative}, we proposed an \ac{IPP} approach linking globally mapped model uncertainties to a robotic planning objective for \ac{AL} in \ac{UAV}-based terrain monitoring. This submission extends our previous method and generalises experimental findings in the following ways. First, we introduce a general \ac{IPP} framework for \ac{AL} in \ac{UAV}-based semantic mapping by linking various uncertainty- and representation-driven acquisition functions to planning objectives, as opposed to just utilising model uncertainties computed via \ac{MC} dropout. Second, we propose new terrain mapping features to improve map-based planning compared to our conference version. Third, we systematically evaluate mapping, planners, and \ac{AL} planning objectives on new datasets from different domains and in a photorealistic simulator. We present a thorough empirical analysis of combining \ac{AL} acquisition functions and \ac{IPP} approaches, giving new insights into how to connect \ac{AL} and autonomous robotic decision-making.

In sum, we make the following four claims. First, our active planning framework for \ac{AL} in \ac{UAV}-based semantic mapping reduces the number of labelled images needed to maximise segmentation performance compared to both traditionally used coverage and random walk data collection. Second, probabilistic global mapping of gathered information enhances map-based planning performance for \ac{AL}. Third, our map-based planners outperform state-of-the-art local planning for \ac{AL}~\cite{Blum2019}. Fourth, we demonstrate the generality of our approach, showing that it significantly reduces labelling effort in largely varying domains irrespective of the used uncertainty estimation methods, planning strategies, and \ac{AL}-based objectives.
We open-source our code for usage by the community at: \url{https://github.com/dmar-bonn/ipp-al-framework}.

\section{Related Work} \label{S:related_work}

Our goal is to collect the most informative images to train a semantic segmentation model with a minimal amount of labelled data using \acp{UAV} in aerial mapping missions. Our approach combines advances in \ac{AL} with \ac{IPP}. This section overviews how our work is placed within these research areas.

\subsection{Active Learning} \label{SS:active_learning}

Active learning aims to maximise model performance while minimising labelled training data. It assumes the existence of a large unlabelled data pool, then iteratively selects a data point from the pool by maximising an acquisition function until a labelling budget is exceeded~\cite{freund1997selective, lewis1994sequential, campbell2000query, tong2001support}. Settles et al.~\cite{settles2009active} provide a comprehensive overview of \ac{AL} approaches for low-dimensional machine learning problems. Recent \ac{AL} approaches focus on training deep learning models from high-dimensional inputs, e.g. images, where a single data point has a negligible effect on model performance. \ac{AL} methods for deep learning collect a batch of data from the pool instead of single data points, called batch-mode \ac{AL}~\cite{guo2007discriminative, gal2017deep, joshi2009multi, sener2017active}. However, these strategies are not applicable in robotic settings since they reason about which images from an existing large data pool should be labelled. In contrast, we propose an \ac{IPP} framework for \ac{AL} collecting new batches of to-be-labelled data directly during a mission in initially unknown environments. We link the \ac{AL} acquisition function to an \ac{IPP} objective, adaptively guiding the \ac{UAV} towards regions of informative training data. Further, we answer the following two open research questions. First, how to incorporate recently proposed acquisition functions~\cite{gal2017deep, beluch2018power, Blum2019} into our \ac{IPP} framework and second, in which ways planners, planning objectives, and terrain mapping influence \ac{AL} performance.

Uncertainty-based \ac{AL} methods select data with the highest model uncertainty~\cite{joshi2009multi, kapoor2007active, gal2017deep, beluch2018power}. Early methods use Gaussian processes~\cite{kapoor2007active} or support-vector machines~\cite{joshi2009multi} to quantify model uncertainty in tasks with low-dimensional inputs. Measuring model uncertainty in deep neural networks is computationally challenging due to their parameter space dimensionality. One approach aims at estimating the model uncertainty deterministically in a single forward pass. Although computationally efficient, these methods are often not well-calibrated in real-world vision tasks~\cite{postels2021practicality}. Alternatively, Gal et al.~\cite{gal2016dropout} propose using dropout at test time, called \ac{MC} dropout, to efficiently approximate the Bayesian posterior over the network parameters. They utilise \ac{MC} dropout in acquisition functions applied to image classification maximising model uncertainty~\cite{gal2017deep}. Other works use neural network ensembles for uncertainty estimation~\cite{lakshminarayanan2017simple, pawlowski2017efficient}. Each network is independently initialised and trained. Ensembles achieve higher prediction performance and better calibration than \ac{MC} dropout~\cite{durasov2021masksembles, beluch2018power}. Further, recent advances make ensemble training computationally more efficient~\cite{huang2017snapshot, durasov2021masksembles}. In this work, we study the applicability of different uncertainty-based \ac{AL} objectives in a robotic planning context.

Representation-based \ac{AL} methods maximise training data diversity by selecting data points with novel representations in feature space~\cite{sener2017active, sinha2019variational, ebrahimi2020minimax}. Generative adversarial network-inspired approaches use a generator learning the joint data representation, while the discriminator distinguishes labelled and unlabelled data~\cite{sinha2019variational, ebrahimi2020minimax}. Sener et al.~\cite{sener2017active} select a number of data points, called a core-set, geometrically covering a data pool in the model's latent space with a minimal number of data points. However, both approaches require large in-domain data pools to learn rich representations of the data-generating distribution. These methods are impractical in our scenario as autonomous robots operate in unknown and visually varying environments. In contrast, Blum et al.~\cite{Blum2019} propose a method for quantifying data novelty in semantic segmentation tasks without access to large in-domain data pools by using kernel-density estimation of unlabelled images in the network's latent space. They use this novelty estimation in a local planning objective and apply it for \ac{AL} in aerial semantic mapping. We integrate their novelty estimation into new global map-based planning objectives. We rigorously analyse its \ac{AL} performance using various global planning schemes and datasets, outperforming their local planning strategy.

\subsection{Informative Path Planning} \label{SS:informative_path_planning}

Informative path planning enables autonomous robots to efficiently and actively explore initially unknown environments subject to platform constraints, such as battery capacity~\cite{dunbabin2012robots}. \ac{IPP} methods have been applied to various environmental monitoring scenarios, including lake monitoring~\cite{hitz2017adaptive}, underwater inspection~\cite{hollinger2013active}, infrastructure surface inspection~\cite{bircher2018receding}, and agricultural monitoring~\cite{popovic2017multiresolution}. We distinguish between non-adaptive approaches, which precompute paths before a mission starts, e.g. coverage planning~\cite{galceran2013survey}, and adaptive approaches, which replan paths online as new data is collected~\cite{hollinger2013active, hitz2017adaptive, popovic2020informative}. We focus on adaptive methods as our goal is to collect informative training data on-the-fly.

Combinatorial approaches solve \ac{IPP} problems in a near-optimal fashion~\cite{chekuri2005recursive, ko1995exact, binney2012branch, singh2009efficient}. However, they exhaustively query the search space scaling exponentially in problem size, which makes most of them impractical for online replanning. In contrast, sampling approaches break the curse of dimensionality to increase the computational efficiency of online \ac{IPP}~\cite{hollinger2014sampling, choudhury2020adaptive}. Hollinger et al.~\cite{hollinger2014sampling} propose receding-horizon rapidly exploring information gathering algorithms to sample motion plans. Choudhury et al.~\cite{choudhury2020adaptive} combine \ac{MC} planning with cost-benefit rollouts to increase sampling efficiency. 

Similarly, optimisation approaches directly optimise \ac{IPP} objectives~\cite{hitz2017adaptive, popovic2020informative, vivaldini2019uav}. Vivaldini et al.~\cite{vivaldini2019uav} utilise Bayesian optimisation to choose a sequence of informative measurement positions for \ac{UAV}-based tree disease monitoring. In a similar problem setup, Hitz et al.~\cite{hitz2017adaptive} leverage the \ac{CMA-ES} to optimise a sequence of measurement positions. Popović et al.~\cite{popovic2020informative} extend this approach by introducing a greedily optimised initial sequence of measurement positions, then using the \ac{CMA-ES} to fine-tune the initial solution resulting in more informative paths.

Geometric approaches collect candidate measurement positions for efficient exploration. The candidate maximising an objective function is chosen as the most informative one~\cite{ghaffari2018gaussian, chen2020autonomous, gonzalez2002navigation}. Gonzalez. et al.~\cite{gonzalez2002navigation} choose the position maximising the potentially visible unexplored space. Ghaffari et al.~\cite{ghaffari2018gaussian} generate candidate positions along probabilistic frontiers of explored space, greedily selecting the one which maximises the expected information gain. Similarly, Cheng et al.~\cite{chen2020autonomous} train an agent choosing frontiers to minimise localisation uncertainty and maximise information gain.

The above-mentioned works consider adaptive \ac{IPP} for mapping environmental phenomena. In contrast, our framework applies planning algorithms to the problem setting of improving robot vision with minimal human labelling effort. We design new \ac{IPP} objective functions to replan paths towards informative training data as new observations are collected and demonstrate their integration into map-based planners.

\subsection{Informative Path Planning for Active Learning} \label{SS:informative_path_planning_for_al}

Using autonomous robots to reduce manual labelling effort for training deep learning models is a relatively unexplored research area. Georgakis et al.~\cite{Georgakis2021} propose a framework for active semantic goal navigation which uses ensembles to estimate model uncertainty in their planning objective. Other methods introduce self-supervised approaches to improve or adapt the robot's perceptions to new environments without the need for manual labelling. Frey et al.~\cite{frey2021continual} introduce a self-improving continual learning framework for semantic segmentation in indoor scenes without manual labelling by generating pseudo-labels from 3D maps. Zurbrügg et al.~\cite{zurbrugg2022embodied} extend this approach to an embodied agent autonomously navigating towards high training data novelty viewpoints. Chaplot et al. \cite{chaplot2021seal} suggest a similar self-supervised approach for semantic segmentation in indoor scenes training an exploration policy with \ac{RL} to target uncertain 3D map parts. The policy training depends on the simulation environment and the currently trained network at the same time. As \ac{RL} performance degrades with simulation to real-world gaps, this method requires the availability of realistic domain-specific simulators and introduces policy re-training costs after each network re-training. Further, as discussed by Chaplot et al. \cite{chaplot2021seal}, the approaches~\cite{frey2021continual, zurbrugg2022embodied, chaplot2021seal} rely on large labelled indoor datasets for pre-training a semantic segmentation model to produce high-quality pseudo-labels in new indoor scenes. If the pre-trained model misclassifies objects, these errors not only prevent learning semantics but could even be reinforced in the case of over-confident predictions. Zurbrügg et al.~\cite{zurbrugg2022embodied} experimentally show that the expected model improvement strongly depends on the chosen pre-training dataset and environment the robot is deployed in. Aerial mapping missions, as in our problem setting, present much more visual variability, with very little and often small pre-training datasets being available, further exacerbating these issues. As the environment and domain are initially unknown, these purely self-supervised methods require enormous engineering work to relax the above-mentioned assumptions and are not directly applicable to our use case.

Most similar to our work is the local planning approach of Blum et al.~\cite{Blum2019} for \ac{AL} in semantic mapping. Their planning objective aims to promote training data novelty in semantic prediction tasks. We combine their ideas on novelty estimation for \ac{AL} with our previous work on \ac{IPP} for \ac{AL}~\cite{ruckin2022informative}. In contrast to Blum et al.~\cite{Blum2019}, we propose a general and unified \ac{IPP} framework supporting probabilistic semantic mapping, various acquisition functions, planning objectives, and map-based planning algorithms. Further, we provide in-depth empirical analyses and show that our map-based planners outperform existing methods~\cite{Blum2019, ruckin2022informative}.

\section{Our Approach} \label{S:approach}

\begin{figure*}[!t]
    \centering
    \includegraphics[width=0.9\linewidth]{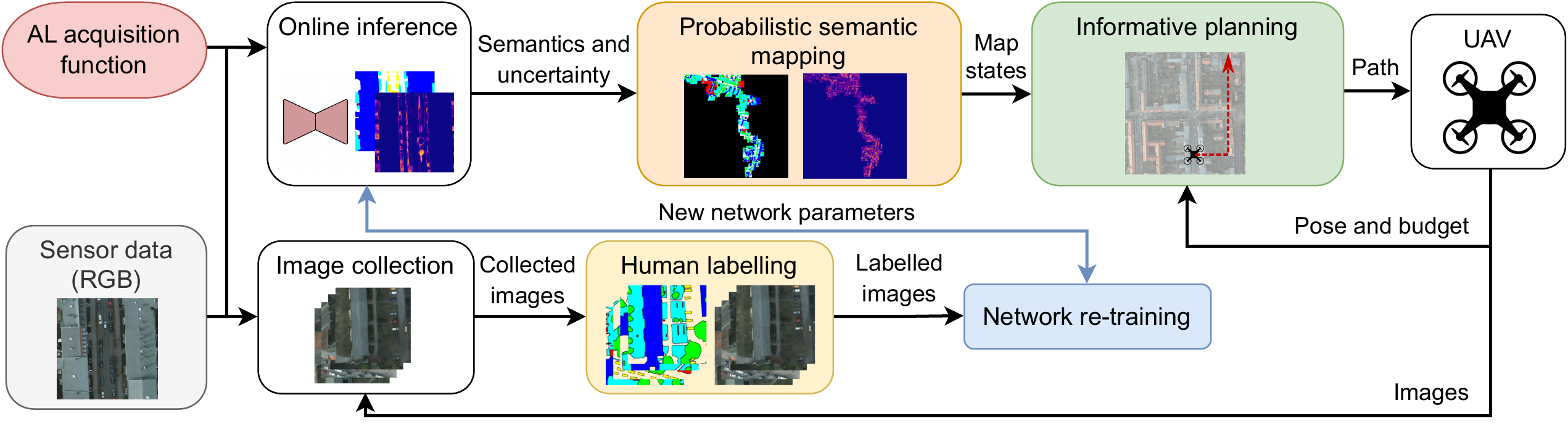}
    \vspace*{-2mm}
    \caption{Overview of our approach. We start with a pre-trained semantic segmentation network deployed on a \ac{UAV}. During a mission, the network processes RGB images to predict pixel-wise semantic labels, model uncertainties (\cref{SSS:uncertainty_acquisition_functions}), and novelty scores (\cref{SSS:representation_acquisition_functions}), which are projected onto the terrain to build global maps capturing these variables (\cref{SS:mapping}). Based on the current UAV position, budget, and posterior map state, our algorithm plans paths for the UAV to collect informative training data for improving the network performance (\cref{SS:path_planning}). After the mission, the collected images are labelled by an annotator and used for network re-training. By guiding the UAV to collect informative training data, our pipeline reduces the human labelling effort.}
    \label{F:approach_overview}
\end{figure*}

We present our general \ac{IPP} framework for \ac{AL} in \ac{UAV}-based semantic mapping. Our setup considers a \ac{UAV} collecting images of a flat terrain using a downwards-facing RGB camera. Assuming no further prior knowledge about the terrain, the goal is to autonomously collect informative training data to improve the robot's perception with minimal human labelling effort. As shown in \cref{F:approach_overview}, our framework links \ac{AL} with planning objectives guiding the \ac{UAV} to regions of informative training data. As new data is collected, we utilise a lightweight \ac{FCN} to predict pixel-wise semantics. Further, we estimate the pixel-wise model uncertainty associated with the prediction and training data novelty of the collected image and then fuse them into probabilistic terrain maps. The \ac{UAV} position, its remaining budget, and the current map state are combined into new \ac{AL}-based information objectives used to replan the future path towards informative training data. A key feature of our framework is its general applicability, as it is agnostic to the chosen network and supports different uncertainty estimation techniques, mapping methods, and map-based planners. The following subsections detail the framework's individual modules and the specific methods we investigate in this work.

\subsection{Active Learning Acquisition Functions} \label{SS:al_acquisition_functions}

We first derive measures for an image's information value when a network is re-trained on this data. To this end, \ac{AL} works propose two main paradigms, uncertainty-based and representation-based acquisition functions. We demonstrate the generality of our approach using either paradigm. 

\begin{figure}[!t]
    \centering
    \includegraphics[width=0.87\columnwidth]{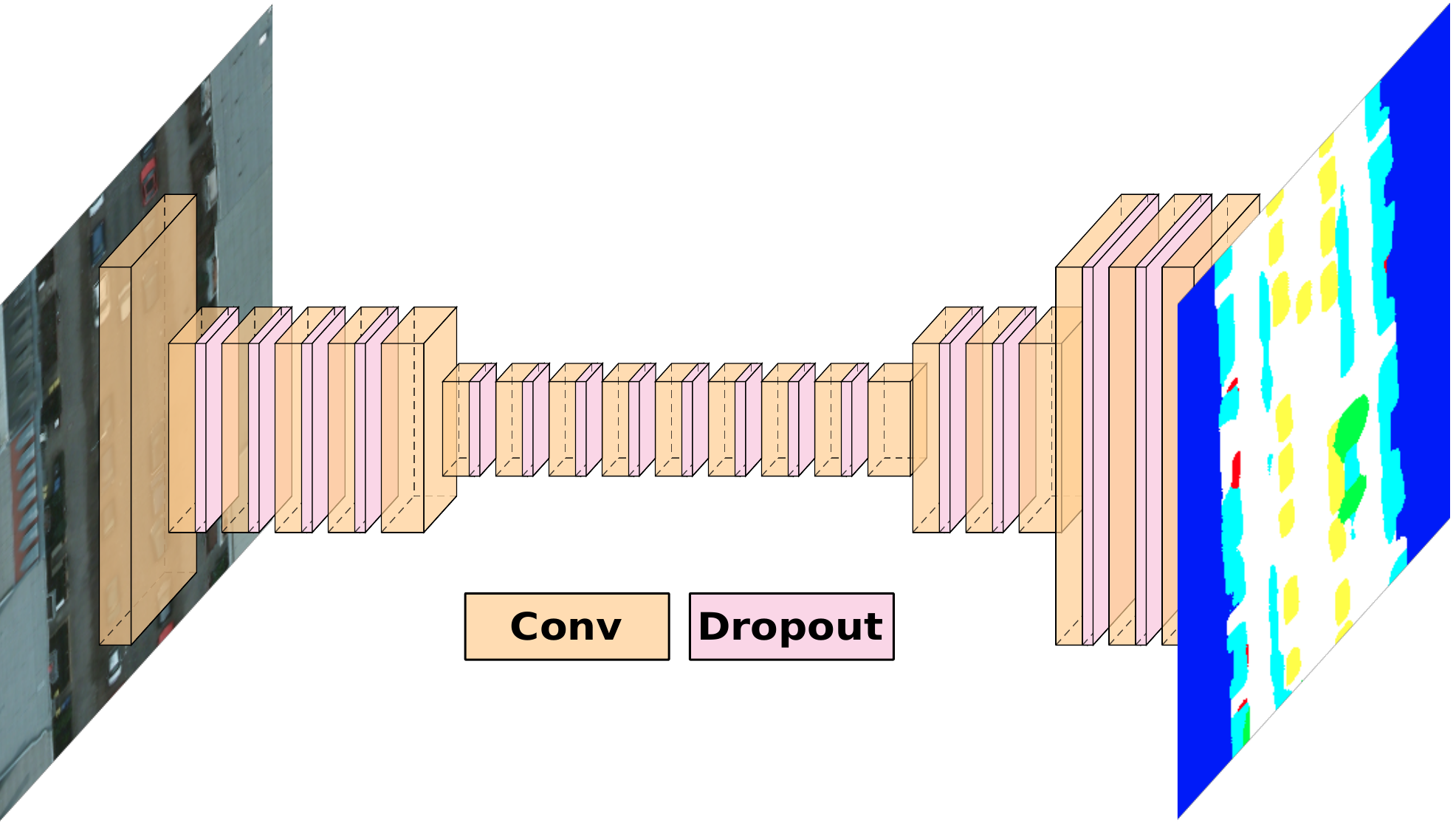}
    \caption{ERFNet architecture proposed by Romera et al.~\cite{Romera2018}. The network takes an RGB image (left) as input and outputs semantic labels (right). We utilise the network in our ensemble method to predict model uncertainty.}
    \label{F:network_architecture}
\end{figure}

We adapt the ERFNet encoder-decoder architecture proposed by Romera et al.~\cite{Romera2018} depicted in \cref{F:network_architecture} to our \ac{AL} use case. Although our framework is agnostic to the chosen network architecture, the lightweight ERFNet is particularly suitable for online robot deployment with limited computational resources. In the following, the model $f^{\bm{W}}(\cdot)$ is parameterised by weights $\bm{W}$ and outputs a probability tensor $p(\bm{y} \mid \bm{f}^{\bm{W}}(\bm{z})) = \text{softmax}(\bm{f}^{\bm{W}}(\bm{z})) \in [0, 1]^{K \times w \times h}$, where $\bm{z}$ is the input RGB image with width $w$ and height $h$, and $\bm{y}$ is the pixel-wise semantic label over the $K$ classes. The training set contains $N$ images $\bm{Z} = \{\bm{z}_1, \ldots, \bm{z}_N\}$ and semantic labels $\bm{Y} = \{\bm{y}_1, \ldots, \bm{y}_N\}$. Our network is trained to minimise cross-entropy with weight decay regularisation factor $\lambda$:
\begin{equation} \label{eq:loss}
    \mathcal{L}(\theta) = - \frac{1}{N} \sum_{i = 1}^{N} \text{log} \, p(\bm{y}_i \,|\,\bm{f}^{\bm{W}}(\bm{z}_i)) + \lambda \lVert \bm{W} \rVert_{2}^{2}\,.
\end{equation}
The following subsections describe different methods to estimate the information value of a candidate image for \ac{AL}.

\subsubsection{Bayesian Uncertainty-based Methods} \label{SSS:uncertainty_acquisition_functions}

We estimate pixel-wise model uncertainty over the prediction $p(\bm{y}\,|\,\bm{f}^{\bm{W}}(\bm{z}))$ as a measure for the informativeness of image $\bm{z}$ for re-training~\cite{gal2017deep, kendall2017uncertainties, gal2016dropout, beluch2018power, houlsby2011bayesian}. We leverage advances in Bayesian deep learning, transforming the deterministic ERFNet into a probabilistic version~\cite{ruckin2022informative}. We consider using two alternative methods: \acf{MC} dropout~\cite{gal2016dropout} and ensembles~\cite{beluch2018power}. To measure model uncertainty, we utilise Bayesian active learning by disagreement~\cite{houlsby2011bayesian}, which computes the mutual information between the unknown labels $\bm{y}$ and the posterior distribution over weights $p(\bm{W}\,|\,\bm{Z}, \bm{Y})$. However, the weights' posterior is intractable for \acp{FCN}~\cite{gal2016dropout}. Thus, we approximate the true posterior prediction~\cite{kendall2017uncertainties}:
\begin{equation} \label{eq:predictive_dist}
    \hat{p}(\bm{y}\,|\,\bm{z}, \bm{Z}, \bm{Y}) = \frac{1}{T} \sum_{i=1}^{T} \text{softmax}(\bm{f}^{\hat{\bm{W}}_i}(\bm{z}))\,,
\end{equation}
where we independently sample $T$ weights $\hat{\bm{W}_i}$ from a prior weight distribution $q(\bm{W})$ performing \ac{MC} integration.

\ac{MC} dropout and ensemble methods provide two alternative approaches to construct the prior $q(\bm{W})$. In \ac{MC} dropout, dropout is applied independently to the weights $\bm{W}$ before each of the $T$ forward passes at test time. In the ensemble method, we train $T$ independently randomly initialised ERFNet models with stochastic mini-batch gradient descent. For further details on \ac{MC} dropout and ensembles, we refer to~\cite{gal2016dropout, ruckin2022informative} and~\cite{beluch2018power}, respectively. Following Gal et al.~\cite{gal2017deep}, we approximate the mutual information using \cref{eq:predictive_dist}:
\begin{multline}  \label{eq:mutual_information}
    \mathbb{I}(\bm{y}, \bm{W}\,|\,\bm{z}, \bm{Z}, \bm{Y}) ~ \approx ~ -\hat{p}(\bm{y}\,|\,\bm{z}, \bm{Z}, \bm{Y})^{T}\text{log}\big(\hat{p}(\bm{y}\,|\,\bm{z}, \bm{Z}, \bm{Y})\big) \\
    ~+~ \frac{1}{T} \sum_{i=1}^{T} ~ p(\bm{y}\,|\,\bm{z}, \hat{\bm{W}}_i)^{T}\text{log}\big(p(\bm{y}\,|\,\bm{z}, \hat{\bm{W}}_i)\big)\,,
\end{multline}
where $\text{log}(\cdot)$ is applied element-wise. Intuitively, model uncertainty is high whenever the posterior prediction entropy is high, while single prediction entropy is low, but disagreeing with each other. We exploit this measure to guide the \ac{UAV} towards more informative areas, i.e. regions of high model uncertainty. Note that our framework is agnostic to both, the model uncertainty estimation method and the chosen network.

\subsubsection{Representation-based Method} \label{SSS:representation_acquisition_functions}

Inspired by recent \ac{AL} works~\cite{sener2017active, sinha2019variational, ebrahimi2020minimax}, we study a representation-based planning objective as an alternative to uncertainty-based objectives. We deterministically quantify the network's confidence in its prediction by estimating the image's novelty to the network $f^{\bm{W}}$ given training images and labels $\bm{Z}, \bm{Y}$~\cite{papernot2018deep, mandelbaum2017distance, Blum2019, postels2021practicality}. Intuitively, the image's novelty is high whenever the network's latent representation of a new image $\bm{z}$ and training images $\bm{Z}$ is dissimilar. Although confidence measures for classification are well-known~\cite{papernot2018deep, mandelbaum2017distance}, they are not directly applicable for semantic segmentation as they do not provide pixel-wise scores and are not invariant to object locations. Thus, we utilise the novelty measure for semantic segmentation proposed by Blum et al.~\cite{Blum2019}.

\begin{figure}[!t]
    \centering
    \includegraphics[width=\columnwidth]{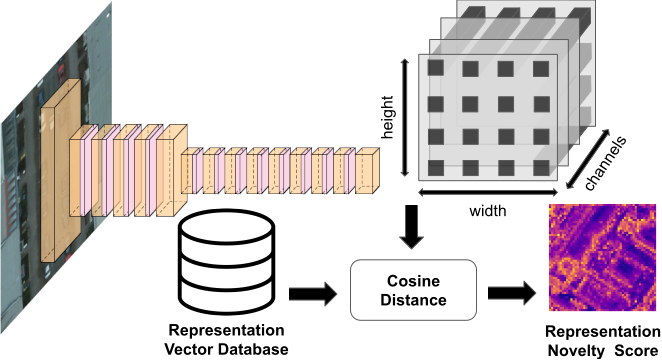}
    \caption{Representation-based image novelty score~\cite{Blum2019}. An RGB image $\bm{z}$ is passed through our ERFNet encoder and its latent vectors $\bm{r}^{\bm{z}}_{i,j} \in \bm{r}^{\bm{z}}$ are extracted along the channel dimension. We compute the cosine distance between each $\bm{r}^{\bm{z}}_{i,j}$ and its $k$-nearest neighbors from the training representation vector database. The resulting novelty image is upsampled to the spatial dimensions of $\bm{z}$. Last, we add all $\bm{r}^{\bm{z}}_{i,j}$ to the representation vector database. Lighter colours indicate higher novelty, i.e. higher informativeness for \ac{AL}.}
    \label{F:representation_score}
\end{figure}

We perform kernel-density estimation in the network latent space by computing the average cosine distance between the latent representations of image $\bm{z}$ and its $k$-nearest latent representations of training images $\bm{Z}$. We exploit the \ac{FCN}'s architecture, where the network $f^{\bm{W}}(\cdot) = d^{\bm{W}_d}(e^{\bm{W}_e}(\cdot))$ consists of an encoder $e^{\bm{W}_e}$ parameterised by $\bm{W}_e$, a decoder $d^{\bm{W}_d}$ parameterised by $\bm{W}_d$, and $\bm{W} = \{\bm{W}_e, \bm{W}_d\}$. Specifically, we extract representations $\bm{e}^{\bm{W}_e}(\bm{z}) = \bm{r}^{\bm{z}} \in \mathbb{R}^{\frac{w}{8} \times \frac{h}{8} \times C}$ after the encoder's last convolutional layer with spatial dimensions downsampled by a factor of $8$ compared to the image, and $C$ channel dimensions as induced by the ERFNet architecture. Hence, $\bm{r}^{\bm{z}}_{i,j}$ is a $C$-dimensional latent vector of the $(i, j)$-th $8 \times 8$ pixels patch of image $\bm{z}$. After model training, we generate a database $R = \{\bm{r}^{\bm{z}_1}_{1,1}, ..., \bm{r}^{\bm{z}_{N}}_{\frac{w}{8},\frac{h}{8}}\}$ of $\frac{w}{8} \cdot \frac{h}{8} \cdot N$ patch-wise representations of the training images $\bm{Z}$. Given an image $\bm{z}$ at inference time, its $(i,j)$-th novelty score is:
\begin{equation} \label{eq:novelty_score}
    r(\bm{z})_{i,j} = \frac{1}{k} \sum_{\bm{r} \in NN(\bm{r}_{i,j}^{\bm{z}})} 1 - \bigg | \frac{\bm{r}^{\top} \bm{r}_{i,j}^{\bm{z}}}{\lVert \bm{\bm{r}_{i,j}^{\bm{z}}} \rVert_2 \, \lVert \bm{r} \rVert_2} \bigg |\, ,
\end{equation}
where $NN(\bm{r}_{i,j}^{\bm{z}})$ is the set of $k$-nearest neighbors of $\bm{r}_{i,j}^{\bm{z}}$ in $R$ with respect to the cosine distance. Intuitively, higher novelty indicates higher informativeness of image $\bm{z}$ for re-training. \cref{F:representation_score} provides a schematic of an image's novelty score computation. For more details, we refer to Blum et al.~\cite{Blum2019}.

A key feature of our framework is that it can easily be adapted to other acquisition functions and \acp{FCN}. This work shows its generality using the uncertainty- and representation-based objectives with ERFNet, as described above.

\subsection{Probabilistic Semantic Mapping} \label{SS:mapping}

An important basis for our new planning objective functions is our 2D multi-layer terrain map. This map captures global semantics, model uncertainties, representation novelties, and training data statistics to provide different sources of information for informative planning. We propose a probabilistic mapping module updating this information online as the \ac{UAV} collects new images of the terrain. To achieve this, we utilise sequential probabilistic occupancy grid mapping~\cite{Elfes1989} to update each map layer when a new measurement arrives. We discretise the terrain into three 2D maps $\mathcal{G}_{S}: G \to \{0,1\}^{K \times W \times L}, ~ \mathcal{G}_{U}: G \to [0,1]^{W \times L}, ~ \mathcal{G}_{R}: G \to [0,1]^{W \times L}$ defined over a grid lattice $G$ with $W \times L$ spatially independent cells capturing the discrete semantic classes, continuous model uncertainties, and continuous novelty scores.

The semantic map $\mathcal{G}_S$ consists of $K$ independent layers $\mathcal{G}_{S_i}: G \to \{0,1\}^{W \times L}$ to map $i \in [K] > 2$ classes. Each grid cell's $G^{m,n}$ random state follows a uniform prior distribution $\mathcal{G}_{S_i}^{m,n} \sim p(\mathcal{G}_{S_i}^{m,n}=1) = \frac{1}{K}$. When a new image $\bm{z}_t$ arrives at time step $t$, the semantic predictions $\hat{p}(\bm{y} \mid \bm{f}^{\bm{W}}(\bm{z}_t))$, see \cref{eq:predictive_dist}, are projected to the flat terrain given the \ac{UAV} position $\bm{p}_t \in \mathbb{R}^3$ and camera intrinsics. We utilise standard occupancy grid mapping for each layer $i$ and cell $G^{m,n}$ computing the posterior belief $\mathcal{G}_{S_i}^{m,n} \sim p(\cdot \,|\, \bm{z}_{1:t}, \bm{p}_{1:t})$:
\begin{equation} \label{eq:semantic_map_update}
    \begin{aligned}
        l(\mathcal{G}_{S_i}^{m,n} \,|\, \bm{z}_{1:t}, \bm{p}_{1:t}) &= l(\mathcal{G}_{S_i}^{m,n} \,|\, \bm{z}_t, \bm{p}_{t}) + \\
        &l(\mathcal{G}_{S_i}^{m,n} \,|\, \bm{z}_{1:t-1}, \bm{p}_{1:t-1}) - l(\mathcal{G}_{S_i}^{m,n})\,,
    \end{aligned}
\end{equation}
where $l(\cdot)$ are the log odds of the binary random variable, $p(\mathcal{G}_{S_i}^{m,n} \,|\, \bm{z}_t, \bm{p}_{t})$ is given by the projected semantic predictions, $p(\mathcal{G}_{S_i}^{m,n} \,|\, \bm{z}_{1:t-1}, \bm{p}_{1:t-1})$ is the recursive map belief, and $p(\mathcal{G}_{S_i}^{m,n})$ is the map prior.

The model uncertainties and novelties are stored in the maps $\mathcal{G}_U$ and $\mathcal{G}_R$ with prior means $\mu_{U,0}$ and $\mu_{R,0}$ respectively. We fuse projected uncertainties $\bm{u}_t$ given by \cref{eq:mutual_information} and novelty scores $r(\bm{z}_t)$ given by \cref{eq:novelty_score} using maximum likelihood estimation assuming normally distributed $\mathcal{G}_U$ and $\mathcal{G}_R$. We maintain a hit map $H: G \to \mathbb{N}^{W \times L}$ counting the total number of times a grid cell was updated during a mission. Then, we update the means $\mu_{U,t}^{m,n}$ and $\mu_{R,t}^{m,n}$ for a grid cell $G^{m,n}$ by:
\begin{equation} \label{eq:occupancy_map_update}
    \begin{aligned}
        \mu_{U,t}^{m,n} &= \mu_{U,t-1}^{m,n} + \frac{1}{H(G^{m,n})} (\bm{u}_{t}^{m,n} - \mu_{U,t-1}^{m,n})\,, \\
        \mu_{R,t}^{m,n} &= \mu_{R,t-1}^{m,n} + \frac{1}{H(G^{m,n})} (r(\bm{z}_t)^{m,n} - \mu_{R,t-1}^{m,n})\,.
    \end{aligned}
\end{equation}

Last, we store a map $T: G \to \mathbb{N}^{W \times L}$ to count how often grid cells occur in the training data set to foster data diversity in our proposed planning objectives. Note that the maps $H(\cdot)$ and $T(\cdot)$ are different as the camera could provide a high-frequency image stream for mapping while images only at the planned measurement position are collected for training.

A key feature of our mapping approach is that we accumulate and update the information between missions by updating the map prior. After each \ac{UAV} mission, the network is re-trained on the collected training data. Re-training changes the semantic predictions, model uncertainty, and representation novelty estimates. Thus, we store all previously collected data and corresponding \ac{UAV} positions. After re-training, we predict semantics, model uncertainties, and representation novelties of the stored data and sequentially fuse them according to \cref{eq:semantic_map_update} and \cref{eq:occupancy_map_update}. Our informed map prior strategy enhances map-based planning by avoiding exploring from scratch or replanning with outdated terrain knowledge.

\subsection{Informative Path Planning} \label{SS:path_planning}

We develop \ac{IPP} algorithms to guide a UAV to adaptively collect useful training data for our \ac{FCN}. Our key idea is to link acquisition functions introduced in \cref{SS:al_acquisition_functions} to planning objective functions. Our planning strategies use the probabilistic terrain maps presented in \cref{SS:mapping} to guide the \ac{UAV} online towards informative training data in an unknown terrain. 

In general, \ac{IPP} algorithms optimise an information criterion $I: \Psi \to \mathbb{R}_{\geq 0}$ over paths $\psi = (\bm{p}_1, \ldots, \bm{p}_{P}) \in \Psi$ defined by $P$ measurement positions $\bm{p}_i \in \mathbb{R}^3$:
\begin{equation} \label{eq:ipp_problem}
    \psi^* = \argmax_{\psi \in \Psi} I(\psi),\, \mathrm{s.t.}\, C(\psi) \leq B\,,
\end{equation}
where $B \geq 0$ is the mission budget, e.g. flight time, and $\Psi$ is the set of all possible paths of length $P$. The function $C: \Psi \to \mathbb{R}_{\geq 0}$ defines the cost of executing a path $\psi$:
\begin{equation} \label{eq:path_costs}
    C(\psi) = \sum_{i=1}^{P-1} c(\bm{p}_{i}, \bm{p}_{i+1})\,,
\end{equation}
where $c: \mathbb{R}^3 \times \mathbb{R}^3 \to \mathbb{R}_{\geq 0}$ computes the flight time between two measurement positions assuming constant acceleration and deceleration $\pm a$, and maximum velocity $v$. The key insight of our work is to couple the \ac{AL} acquisition functions with \ac{IPP} information criteria $I(\cdot)$. This allows us to maximise model performance and minimise the labelling effort resulting from collecting images along a planned path $\psi$. 

\begin{figure}[!t]
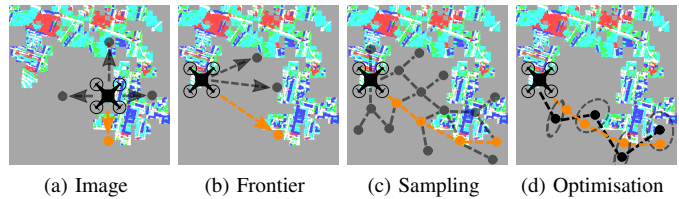

    \centering

    \subfloat[Image \label{SF:image_planner}]{\includegraphics[width=0.24\columnwidth]{figures/approach/planning/image_planner.pdf}}
    \hfill
    \subfloat[Frontier \label{SF:frontier_planner}]{\includegraphics[width=0.24\columnwidth]{figures/approach/planning/frontier_planner.pdf}}
    \hfill
    \subfloat[Sampling \label{SF:mcts_planner}]{\includegraphics[width=0.24\columnwidth]{figures/approach/planning/mcts_planner.pdf}}
    \hfill
    \subfloat[Optimisation \label{SF:cmaes_planner}]{\includegraphics[width=0.24\columnwidth]{figures/approach/planning/cmaes_planner.pdf}}
    \hfill
    \caption{Our planning strategies for training data collection. Dark gray dots and lines indicate candidate measurements and paths evaluated based on their estimated information adding these images to the training set. Orange dots and lines indicate the most informative chosen measurements and paths. Light gray depicts unexplored terrain. In (c), black indicates the greedy initialisation, and ellipses indicate the optimisation of candidate paths in continuous space.}
    \label{F:planning_strategies}
\end{figure}

We propose four different replanning strategies in our framework, one local image-based and three global frontier, sampling and optimisation schemes with information criteria $I(\cdot)$ optimising \cref{eq:ipp_problem} given the current terrain map states. The planners are illustrated in \cref{F:planning_strategies}. In our experimental evaluation, we compare each planner's performance in terms of segmentation performance over the total labelling cost.

In the following, we exemplarily present our planning objectives with respect to the globally mapped model uncertainties $\mathcal{G}_{U, t}$ at a time step $t$ (\cref{SSS:uncertainty_acquisition_functions}). In case of the representation-based objective (\cref{SSS:representation_acquisition_functions}), we substitute the uncertainties $\mathcal{G}_{U, t}$ with novelties $\mathcal{G}_{R, t}$, see \cref{eq:occupancy_map_update}. This variable can also be changed to capture other \ac{AL} acquisition functions.

\noindent \textbf{Local planner}. Our local image-based planner follows the direction of the highest estimated training data information in the image recorded at the current \ac{UAV} position. Specifically, we choose the direction of the image edge $e_{\bm{z}_t}^{*}$ with the highest average \ac{AL} value normalised by the current training data counts $T_{t}(\bm{p}_t)$ in the grid's subset spanned by the camera field of view from position $\bm{p}_t$ projected to the flat terrain. This way, we select neighboring informative images while locally fostering training data diversity. Then, $\bm{p}_{t+1}^{*}$ is reached by taking a predefined step size towards the direction of edge $e_{\bm{z}_t}^{*}$ at a fixed altitude. This resembles the planner proposed by Blum et al.~\cite{Blum2019} and generalises it to any \ac{AL} objective.

\noindent \textbf{Frontier-based planner.} Our global geometric planner guides the \ac{UAV} towards frontiers of the explored terrain with the highest \ac{AL} objective in the terrain map. We use the hit map $H(\cdot)$ to identify exploration frontiers. Particularly, we greedily choose the next-best measurement position $\bm{p}^{*}_{t+1}$ from a set of candidate positions $\bm{p}^{c}_{t+1}$ equidistantly sampled along the frontiers at a fixed altitude. As the planner acts greedily, optimising \cref{eq:ipp_problem} reduces to selecting the path $\psi^{*} = (\bm{p_{t+1}^{*}})$:
\begin{equation} \label{eq:frontier_objective}
    \bm{p}_{t+1}^{*} = \argmax_{\bm{p}_{t+1}^{c}} I((\bm{p}_{t+1}^{c})) = \argmax_{\bm{p}_{t+1}^{c}} \frac{\lVert \mathcal{G}_{U, t}(\bm{p}_{t+1}^{c}) \rVert_{1}}{\lVert T_{t}(\bm{p}_{t+1}^{c}) \rVert_{1}}\,,
\end{equation}
where $\mathcal{G}_{U, t}(\bm{p}_{t+1}^{c})$ and $T_{t}(\bm{p}_{t+1}^{c})$ are the globally mapped model uncertainties and training data counts within the camera field of view from position $\bm{p}_{t+1}^{c}$, and $\lVert \cdot \rVert_1$ is the norm summing all elements in these subsets. This way, our frontier planner trades off both exploration of unknown space for data diversity and focusing on regions potentially valuable for \ac{AL}. 

\noindent \textbf{Optimisation-based planner.} Our optimisation-based planner selects a path $\psi_{t+1}^{*}$ over a fixed horizon of multiple time steps. We utilise a two-step approach for efficient online replanning inspired by Popovi{\'c} et al.~\cite{popovic2020informative}. First, we greedily select a path $\psi^{g}_{t+1}$ of length $P$ over a grid above the terrain. Second, we use an optimisation procedure to fine-tune $\psi^{g}_{t+1}$ in the continuous \ac{UAV} workspace and return the next-best path $\psi_{t+1}^{*}$.

First, we iteratively select a path $\psi^{g}_{t+1} = (\bm{p}_{t+1}^{g}, \ldots, \bm{p}_{t+P}^{g})$, where each measurement position $\bm{p}_{t+i}^{g}$, $i \in \{1, \ldots, P\}$, is greedily chosen over a sparse lattice $F$ of discrete candidate positions $\bm{p}^{c}$ at a fixed altitude:
\begin{equation} \label{eq:greedy_cmaes_init}
    \bm{p}_{t+i}^{g} = \argmax_{\bm{p}^{c} \in F} \frac{\lVert \mathcal{G}_{U,t}(\bm{p}^{c}) \rVert_1}{c(\bm{p}_{t+i-1}, \bm{p}^c) \lVert T_{t+i-1}(\bm{p}^c) \rVert_1}\,,
\end{equation}
where $T_{t+i-1}(\bm{p}^c)$ is the subset of the forward-simulated training data count map given by the camera field of view at position $\bm{p}^c$. The forward simulation of the current map $T_{t}$ based on the previously selected positions $(\bm{p}_{t+1}^{g}, \ldots, \bm{p}_{t+i-1}^{g})$ is crucial as one cannot forward-simulate model uncertainties. Forward-simulating $T_{t}$ linearly decreases uncertainty with the number of a grid cell's training set occurrences. This fosters data diversity and terrain exploration. 

Second, we refine the greedy positions of $\psi_{t+1}^{g}$ in parallel in the continuous \ac{UAV} workspace. To this end, we initialise an optimisation procedure with the greedy solution $\psi_{t+1}^{g}$ and extend \cref{eq:greedy_cmaes_init} to an information criterion $I(\cdot)$ evaluating candidate paths $\psi_{t+1}^{o} = (\bm{p}_{t+1}^{o}, \ldots, \bm{p}_{t+P}^{o})$:
\begin{equation} \label{eq:cmaes_fine_tuning}
    I(\psi_{t+1}^{o}) = \frac{\sum_{i=1}^{P} \lVert \mathcal{G}_{U,t}(\bm{p}_{t+i}^{o}) \rVert_1}{c(\bm{p}_{t+i-1}^{o}, \bm{p}_{t+i}^{o}) \lVert T_{t+i-1}(\bm{p}_{t+i}^{o}) \rVert_1}.
\end{equation}

The candidate path $\psi_{t+1}^{*} = (\bm{p}_{t+1}^{*}, \ldots, \bm{p}_{t+P}^{*})$ maximising \cref{eq:cmaes_fine_tuning} is chosen and measurement position $\bm{p}_{t+1}^{*}$ is executed. We found that normalising \ac{AL} information of a path by its execution costs leads to more efficient budget allocation. This planning strategy supports any optimisation algorithm, which can optimise objective function \cref{eq:cmaes_fine_tuning}.

\noindent \textbf{Sampling-based planner.} Our sampling-based planner utilises \ac{MCTS}~\cite{browne2012survey} to optimise a next-best measurement position $\bm{p}_{t+1}^{*}$ in a non-myopic fashion. We simulate a number of future paths $\psi_{t+1} = (\bm{p}_{t+1}^{n_1}, \ldots, \bm{p}_{t+P}^{n_P})$ of length $P$ at a fixed altitude. Each tree node $n_i$ at depth $i \in \{0, \ldots, P\}$ is uniquely defined by its state $S^{n_i} = \{\bm{p}_{t+i}^{n_i}, T_{t+i}^{n_i}, \Bar{B}^{n_i}\}$ consisting of a measurement position $\bm{p}_{t+i}^{n_i}$, forward-simulated training data count map $T_{t+i}^{n_i}$ along the tree's traversed path to node $n_i$, and remaining mission budget $\Bar{B}^{n_i}$. The tree's root node $n_0$ is defined by $S^{n_0} = \{\bm{p}_{t}, T_{t}, B\}$, where $\bm{p}_{t}$, $T_{t}$ and $B$ are the current \ac{UAV} position, training data count map, and mission budget. At each node, the planner selects the next position from a discrete set of actions with different step sizes and orientations. While traversing the search tree, we use the upper confidence bound bandit algorithm~\cite{browne2012survey} to choose a child node. When reaching a leaf node, we roll out the path by sampling actions uniformly at random in each subsequent node until the remaining budget is exceeded or path length $P$ is reached. A simulated path's information $I(\psi_{t+1})$ is computed by summing rewards along subsequent parent and child nodes $n_i, n_{i+1}$ given by:
\begin{equation} \label{eq:mcts_reward}
    R(n_i, n_{i+1}) = \frac{\lVert \mathcal{G}_{U,t}(\bm{p}_{t+i+1}^{n_{i+1}}) \rVert_1}{c(\bm{p}_{t+i}^{n_{i}}, \bm{p}_{t+i+1}^{n_{i+1}}) \lVert T_{t+i}^{n_i}(\bm{p}_{t+i+1}^{n_{i+1}}) \rVert_1}.
\end{equation}

\begin{table*}[!t]
\caption{Environment, sensor, and \ac{UAV} mission parameters for the three datasets in our experiments}
\label{T:dataset_information}
\centering
\begin{tabular}{@{}cccccccccc@{}}
\toprule
\textbf{Dataset} & \textbf{Type} & \textbf{Classes} & \textbf{Task} & \textbf{Area} [m$\times$m] & \textbf{FoV} [px$\times$px] & \textbf{GSD} [cm/px] & \textbf{Altitude} [m] & \textbf{Budget} [s] & \textbf{Test Images} \\ \midrule
Potsdam~\cite{Potsdam2018}          & Orthomosaic   & 7                & Urban         & 900$\times$900     & 400$\times$400    & 15.0         & 30                & 1800            & 3500              \\
RIT-18~\cite{kemker2018algorithms}           & Orthomosaic   & 6                & Land Cover    & 261$\times$568     & 400$\times$400    & 8.0         & 15                & 400             & 3500              \\
Flightmare~\cite{song2020flightmare}       & Unity/Gazebo  & 10               & Industrial    & 150$\times$130     & 480$\times$720    & 8.3        & 20                & 150             & 1000              \\ \bottomrule
\end{tabular}
\end{table*}

Note that $T_{t+i}^{n_i}(\bm{p}_{t+i+1}^{n_{i+1}})$ are the training data occurrences at the child node's position assuming the training data count after collecting a measurement at the parent's node position. This way, the reward estimates the next position's information value given the map state at replanning time $t+i$. After simulating a number of paths $\psi_{t+1}$, we select the root's child node $n_{1}$ with the highest average information value and the \ac{UAV} moves to its associated measurement position $\bm{p}_{t+1}^{n_1}$.

To show that our approach supports various planning algorithms, we proposed the four diverse planners above and showcase their integration into our modular framework. Further, we highlight that our planning strategies are agnostic to the acquisition functions introduced in \cref{SS:al_acquisition_functions}.

\section{Experimental Results} \label{S:experimental_results}

Our experiments evaluate our proposed method and show the benefits of our mapping module (\cref{SS:mapping_results}), Bayesian ensemble (\cref{SS:bayesian_ensemble_results}), and \ac{AL} planning objective functions (\cref{SS:planning_objectives_results}). We verify our planning framework's generality and analyse its \ac{AL} performance on vastly differing real-world aerial datasets and in a photo-realistic simulator against classical coverage and random walk exploration data collection (\cref{SS:rit18_and_flightmare_results}). We conduct a sensitivity analysis validating the framework's robustness to the choice of model architecture, pre-training schemes, and \ac{UAV} starting positions (\cref{SS:sensitivity_analysis}). Our framework consistently maximises semantic segmentation performance while minimising human labelling effort. Notably, our experiments show that our map-based planners outperform the local planning strategy for \ac{AL} proposed by Blum et al.~\cite{Blum2019}, which, to the best of our knowledge, is the only directly comparable approach to date.

\subsection{Experimental Setup} \label{SS:experiment_setup}

\noindent \textbf{Baselines}. We compare our planning framework against three baselines: a traditionally-used coverage-based collection strategy~\cite{galceran2013survey}, and two random walk-based exploration planners.

\begin{figure}[!t]
    \centering
    \subfloat[Coverage pattern \label{SF:potsdam_coverage_paths}]{\includegraphics[width=0.32\columnwidth]{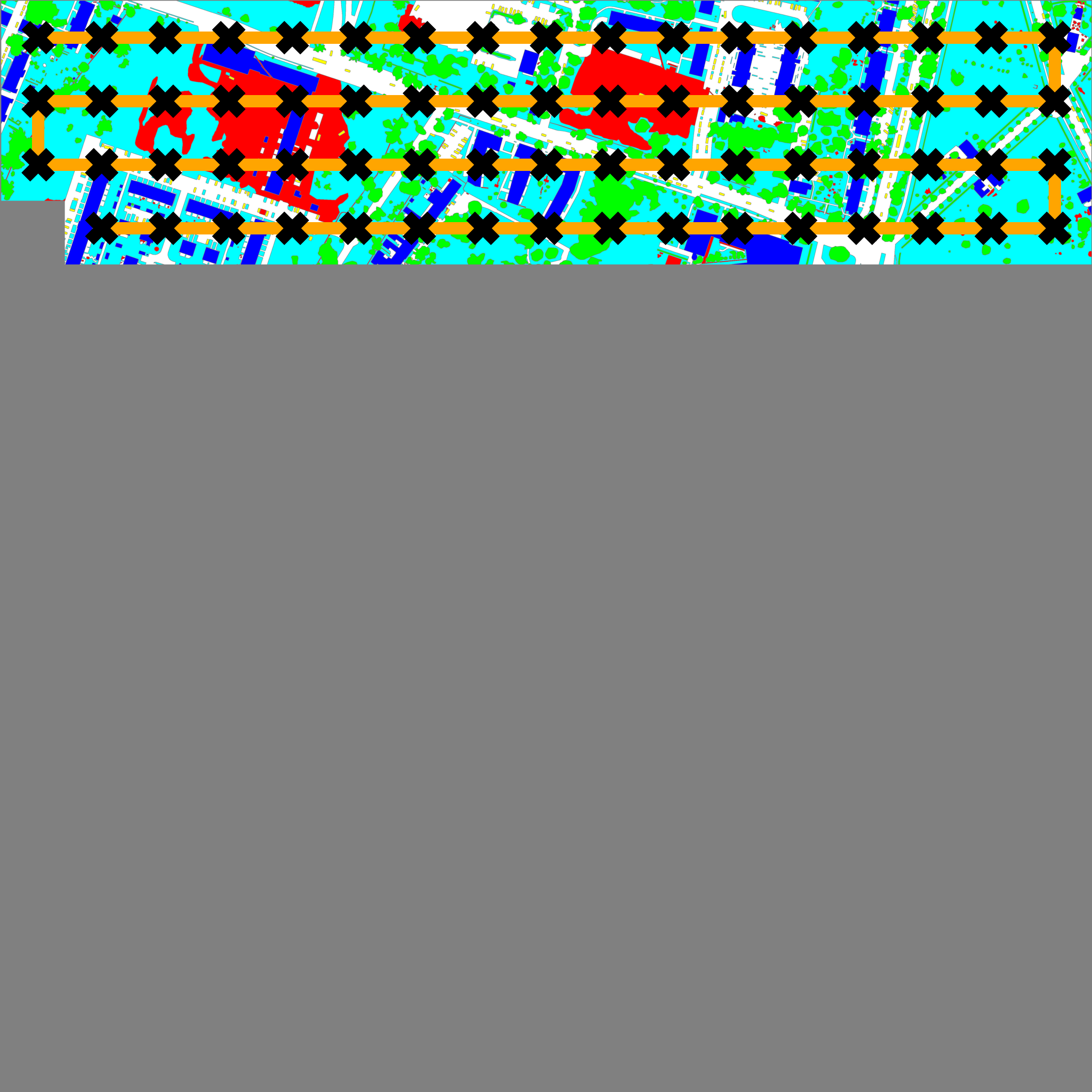}}
    \hfill
    \subfloat[Global random walk \label{SF:potsdam_global_random_paths}]{\includegraphics[width=0.32\columnwidth]{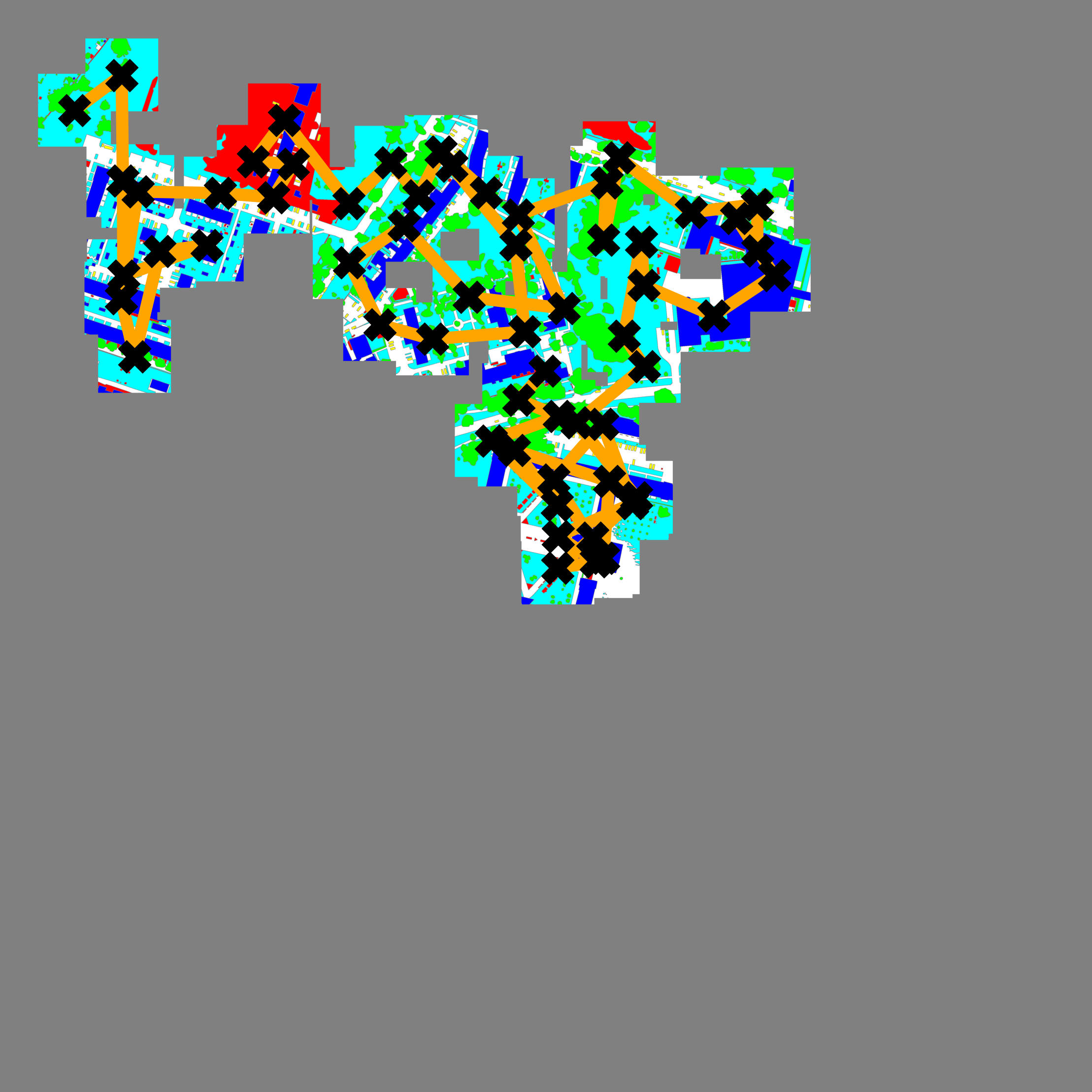}}
    \hfill
    \subfloat[Local random walk \label{SF:potsdam_local_random_paths}]{\includegraphics[width=0.32\columnwidth]{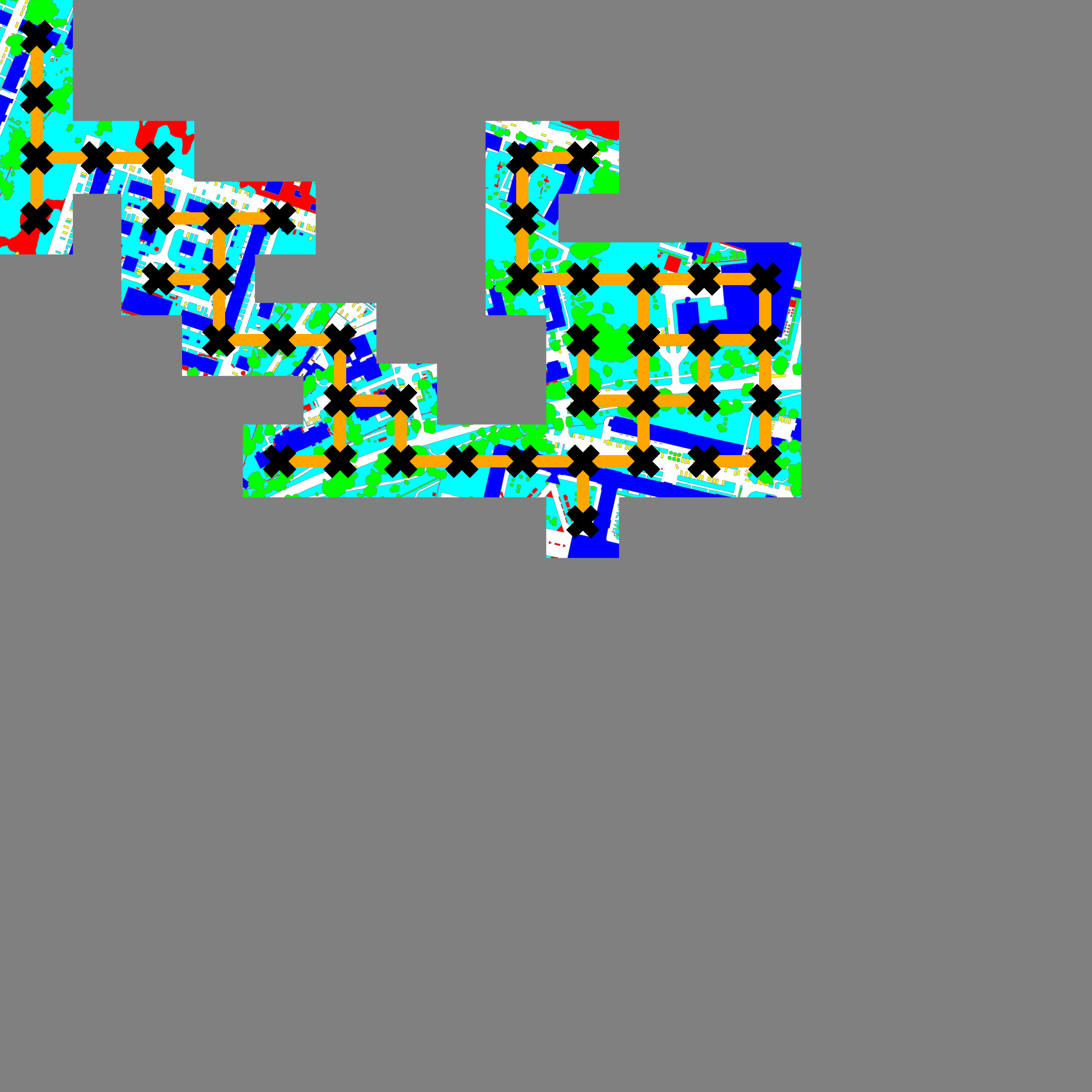}}
\caption{Examples of paths planned by the baseline strategies on ISPRS Potsdam~\citep{Potsdam2018}. Orange lines show paths planned in one mission, black crosses indicate collected training images, and gray areas depict unexplored terrain.}
\label{F:potsdam_baselines_paths}
\end{figure}

The coverage strategy precomputes a static path maximising the area covered by the \ac{UAV} to foster spatial coverage of training data. We precompute lawnmower-like patterns before each mission starts, alternate the pattern's orientations, and vary the step size between measurement positions. 

We consider two random walk exploration planners, local and global planning. Similar to the local planner, the local exploration planner chooses for a given \ac{UAV} position one of the four image edges at random and follows the edge direction with predefined step size. The global exploration planner randomly selects a \ac{UAV} position in the continuous space above the terrain, similar to our map-based planners. For better budget management, we sample a step size uniformly at random between a minimum and maximum radius around the \ac{UAV}, then also set its heading uniformly at random. 

This way, both exploration planners aim to foster data diversity while handling the budget properly. As they resemble the action spaces of the planners introduced in \cref{SS:path_planning}, we can study the influence of our action space design and verify that our active planners maximise \ac{AL} performance beyond random effects. \cref{F:potsdam_baselines_paths} exemplifies the paths planned by all three baselines on the ISPRS Potsdam dataset~\cite{Potsdam2018}.

\begin{figure*}[!t]
    \centering
    \subfloat[ISPRS Potsdam \label{SF:results_potsdam_baselines}]{\includegraphics[width=0.32\textwidth]{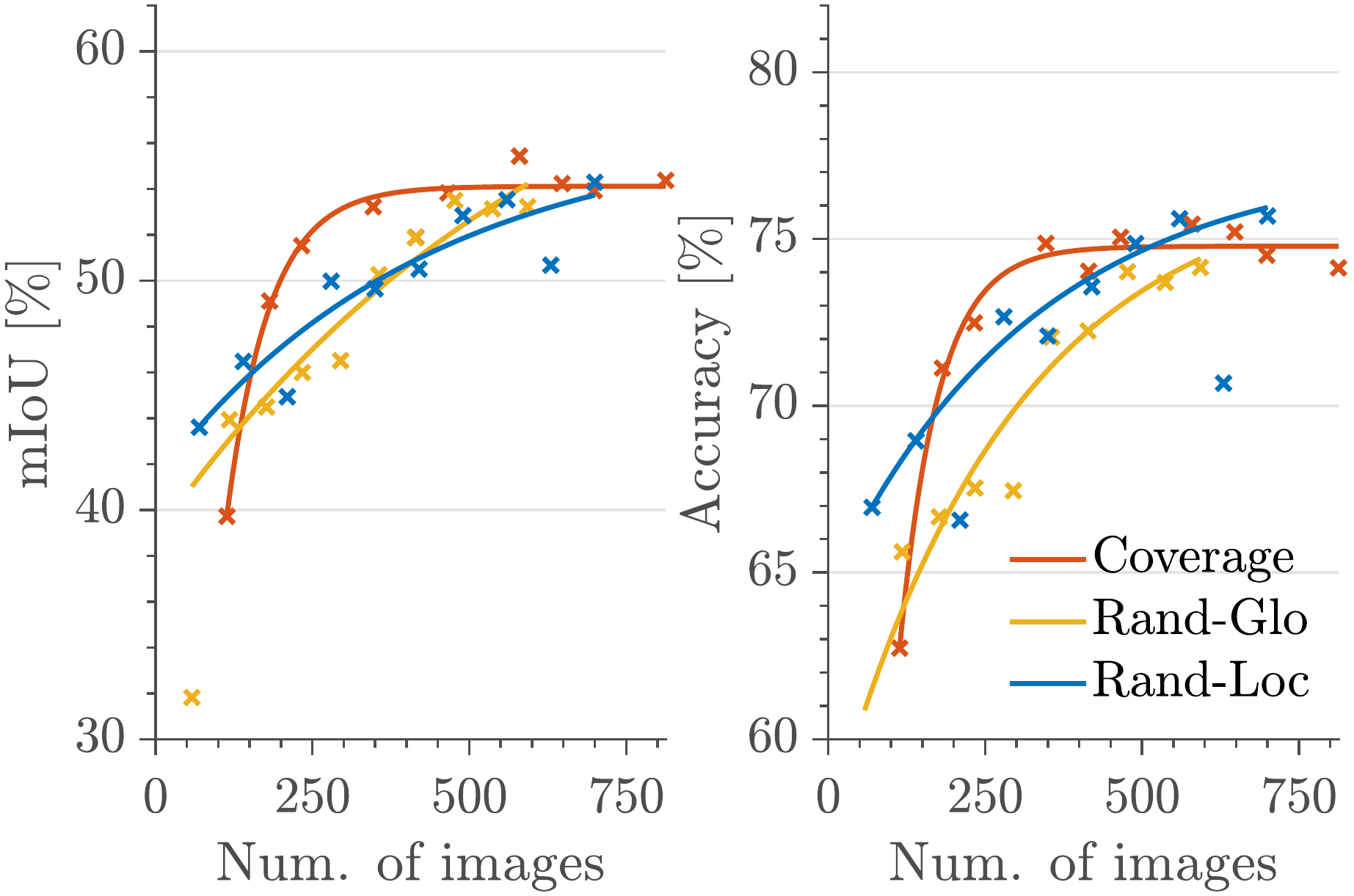}}
    \hfill
    \subfloat[RIT-18 \label{SF:results_rit18_baselines}]{\includegraphics[width=0.32\textwidth]{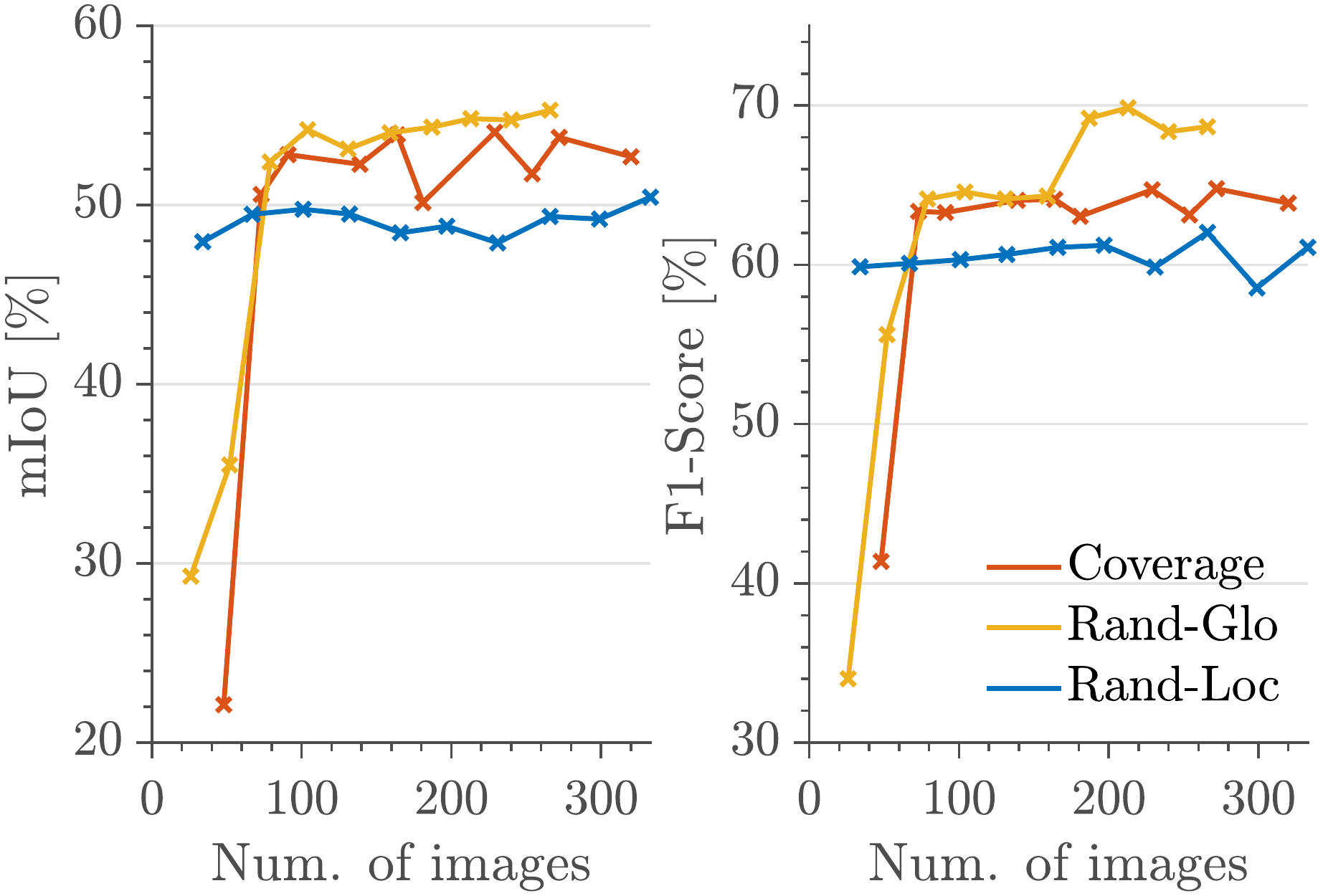}}
    \hfill
    \subfloat[Flightmare \label{SF:results_flightmare_baselines}]{\includegraphics[width=0.32\textwidth]{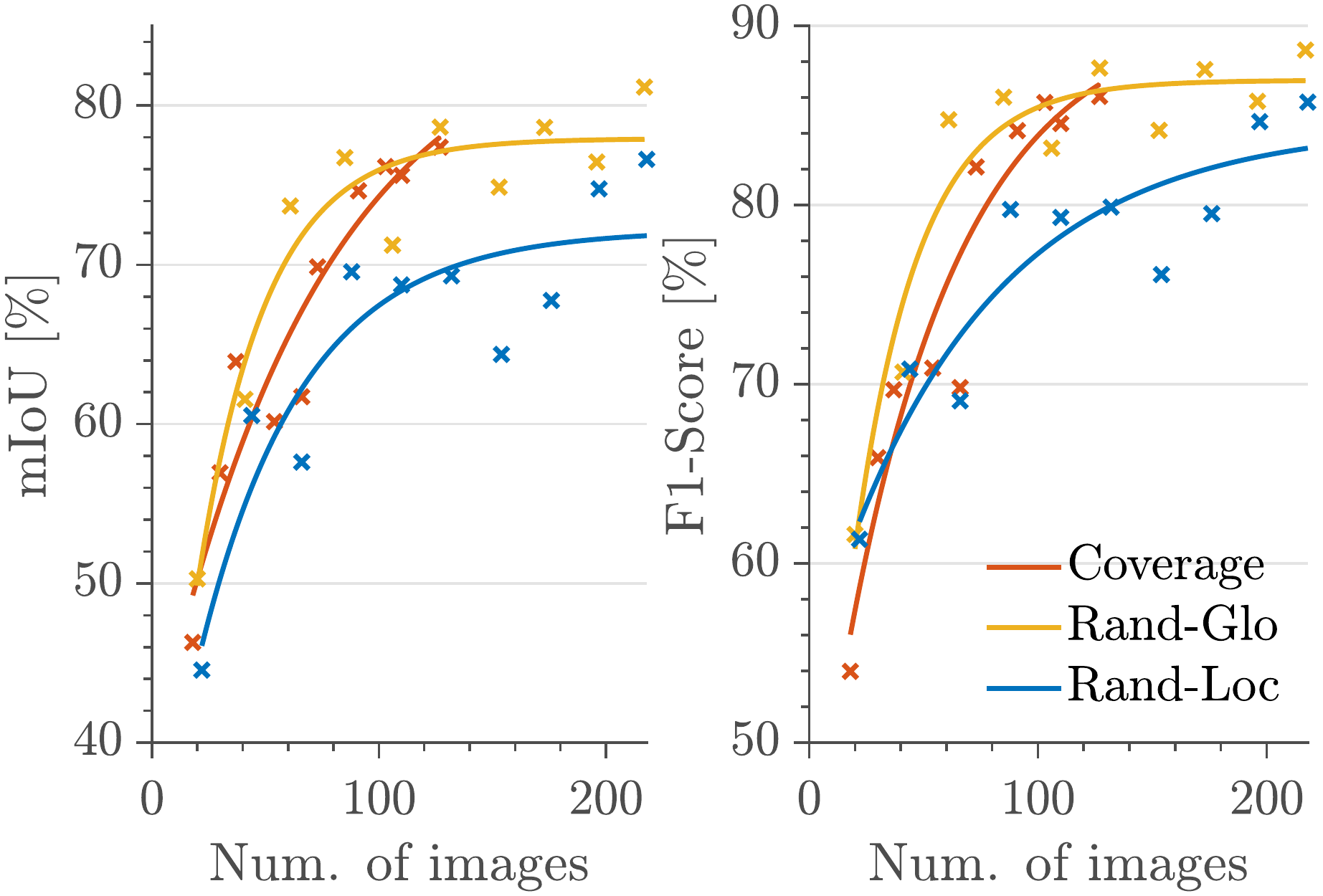}}

    \caption{Comparison of \ac{AL} performance of our three baseline planning approaches with \ac{MC} dropout inference (a) on the ISPRS Potsdam dataset~\cite{Potsdam2018}, (b) on the RIT-18 dataset~\cite{kemker2018algorithms}, and (c) in the photo-realistic Flightmare \ac{UAV} simulator~\cite{song2020flightmare}. Steeper curves indicate better \ac{AL} performance. We compare our planning strategies to the best-performing baseline in each setting: coverage on ISPRS Potsdam and the global random walk on RIT-18 and in Flightmare.}
    \label{F:results_baselines}
\end{figure*}

\noindent \textbf{Datasets.} We evaluate our planning framework on two real-world orthomosaic datasets and in a photorealistic physics-based \ac{UAV} simulator resembling real-world deployment conditions. Detailed environment, sensor, and \ac{UAV} mission settings are shown in \cref{T:dataset_information}. Below, we highlight the key differences between the three scenarios.

First, we use the large 7-class urban aerial ISPRS Potsdam orthomosaic dataset~\cite{Potsdam2018}. This dataset is characterised by a dense spatial distribution of classes, such that the coverage and exploration baselines can collect visually and semantically different features easily. We sample 4000 train, 1000 validation, and 3500 test images uniformly at random from non-overlapping regions in the orthomosaic. We use the ISPRS Potsdam dataset for the main experiments evaluating our mapping module (\cref{SS:mapping_results}), Bayesian ensemble (\cref{SS:bayesian_ensemble_results}), and planning objectives (\cref{SS:planning_objectives_results}).

Second, we evaluate our approach on the land cover RIT-18 orthomosaic dataset~\cite{kemker2018algorithms} consisting of semantics covering large connected areas, e.g. asphalt, vegetation, and lake, and local regions, e.g. building, with six classes in total. As the RIT-18 dataset does not provide different orthomosaics for training and testing, we evaluate the \ac{UAV}'s vision capabilities by sampling the test set from the same area. In contrast to the ISPRS Potsdam dataset, this does not allow us to draw conclusions about the model's generalisability, but about its performance in the deployed environment only. This is still a crucial skill for autonomous robot deployment. Our evaluation protocol on RIT-18 resembles that of Blum et al.~\cite{Blum2019}.

Last, we test our framework in Flightmare, a photorealistic simulator with a physics engine for emulating \ac{UAV} dynamics simulation~\cite{song2020flightmare}. We deploy a \ac{UAV} in the provided `Industrial' environment introducing 10 semantic classes of different spatial distributions, e.g. hangar, container, road, fence, and pipe. The scene covers a dense area leading to compactly distributed semantics easily explorable by the baseline approaches. As the `Industrial' terrain is small, we evaluate the \ac{UAV}'s semantic segmentation performance in the deployed environment only.

We perform a study comparing the \ac{AL} performance of the baseline strategies in \cref{F:results_baselines}. On the ISPRS Potsdam dataset, the coverage pattern is the superior baseline, while the global random walk exploration performs best on the RIT-18 dataset and in the Flightmare simulator. Note that, while \ac{MC} dropout is used in \cref{F:results_baselines} to predict semantic segmentation, we found that similar results hold true for deterministic network and ensemble inference. For visual clarity, we only compare our framework to the baselines with the strongest \ac{AL} performance. 

\noindent \textbf{Evaluation Metrics.} Our \ac{AL} planning pipeline aims to maximise semantic segmentation performance with minimal human labelling effort, i.e. minimal training data. In line with the standard in \ac{AL} literature~\cite{gal2017deep, Blum2019, yang2017suggestive, Tuia2009, tong2001support, wang2015querying, sener2017active, guo2007discriminative, joshi2009multi, beluch2018power, houlsby2011bayesian}, our key evaluation metrics assess semantic segmentation performance (dependent variable) over the number of collected training images (independent variable). Higher semantic segmentation performance thanks to newly added images indicate better \ac{AL}, and thus, planning performance. We choose \ac{mIoU}, per-pixel accuracy, and per-pixel F1-score to access semantic segmentation performance. \ac{mIoU} is used in popular semantic segmentation benchmarks~\cite{everingham2010pascal, cordts2016cityscapes}. It is defined as $mIoU = \frac{TP}{TP+FP+FN}$, where $TP$, $FP$, $TN$, and $FN$ are the true and false positives, and true and false negatives. Per-pixel accuracy $acc$ and F1-score $f1$ are typically used in classification benchmarks~\cite{deng2009imagenet}. They are defined as $acc = \frac{TP}{TP+FP+TN+FN}$ and $f1 = \frac{2TP}{2TP + FP + FN}$. RIT-18 and Flightmare have strongly imbalanced class distributions. Thus, we use the F1-score instead of accuracy for these scenarios. Note that, as training datasets are incrementally collected while exploring an initially unknown environment, the training image distribution changes during deployment as new visual features or semantics are discovered. Hence, the training image distribution could differ from the true image distribution, which could lead to non-monotonic model improvement. To make model performance trends easier to follow, we additionally fit trend lines for the experiments conducted on the ISPRS Potsdam and Flightmare datasets. As performance trends are less regular on the RIT-18 dataset due to the more challenging exploration of semantics, we show piecewise linear line plots for these experiments.

\noindent \textbf{Training Procedure.} We utilise a lightweight Bayesian ERFNet for semantic segmentation as described in \cref{SS:al_acquisition_functions}. The model is pre-trained on the Cityscapes dataset~\citep{cordts2016cityscapes} to start experiments and training after each of the $10$ subsequent data collection missions from the same checkpoint. This also avoids catastrophic forgetting and accumulating train time. We re-train the model until convergence with batch size $8$ and weight decay $\lambda = (1 - p) / 2N$ in \cref{eq:loss}, where $p = 0.5$ is the dropout probability, and $N$ is the number of training images~\cite{gal2017deep}. All other model hyperparameters follow the standard ERFNet~\cite{Romera2018}, not tuned for maximal performance in our setting, and kept fixed with changing datasets and planners.

\noindent \textbf{Planning Hyperparameters.} Our optimisation-based planner leverages the \ac{CMA-ES} procedure as it has been shown to yield competitive performance in terrain monitoring tasks~\cite{hitz2017adaptive, popovic2020informative}. We fix a set of hyperparameters for all planners with reasonable length scales on the ISPRS Potsdam dataset, i.e. \ac{UAV} step sizes, minimum and maximum action space radii, grid discretisation, and initial \ac{CMA-ES} covariance. Only these hyperparameters dependent on the aerial dimensions are scaled accordingly with changing environment sizes. The scale-independent hyperparameters, e.g. number of \ac{MCTS} simulations, are set in line with prior works~\cite{popovic2020informative, browne2012survey}. We fix the \ac{UAV}'s starting position to the top-left corners of each terrain.

\noindent \textbf{Planning Strategies.} We outline our planning strategies in detail in \cref{SS:path_planning}. In our experiments, we refer to the planners in the legends as follows: the local planner is named \textit{Local}, the frontier-based planner is named \textit{Frontier}, the optimisation-based planner is named \textit{Optimisation}, and the sampling-based planner is named \textit{Sampling}. The baseline approaches are referred to as follows: the coverage pattern is called \textit{Coverage}, and the local and global random walk exploration strategies are abbreviated with \textit{Rand-Glo} and \textit{Rand-Loc}.

\subsection{Informative Mapping} \label{SS:mapping_results}

The first set of experiments analyses the performance of our approach. It (i) verifies the superior \ac{AL} performance of our planning framework over the baselines; (ii) shows that our global map-based planners outperform state-of-the-art local planning; and (iii) quantifies the benefit of our mapping module for the global map-based planners. The experiments are evaluated on the ISPRS Potsdam dataset~\cite{Potsdam2018}. To focus on evaluating the effect of our mapping module (\cref{SS:mapping}) on the map-based planners, we fix Bayesian model uncertainty as our planning objective estimated with \ac{MC} dropout (\cref{eq:mutual_information}). Map priors are recomputed before each mission starts to allow for maximally informed global planning. 

\begin{figure}[!t]
    \includegraphics[width=\linewidth]{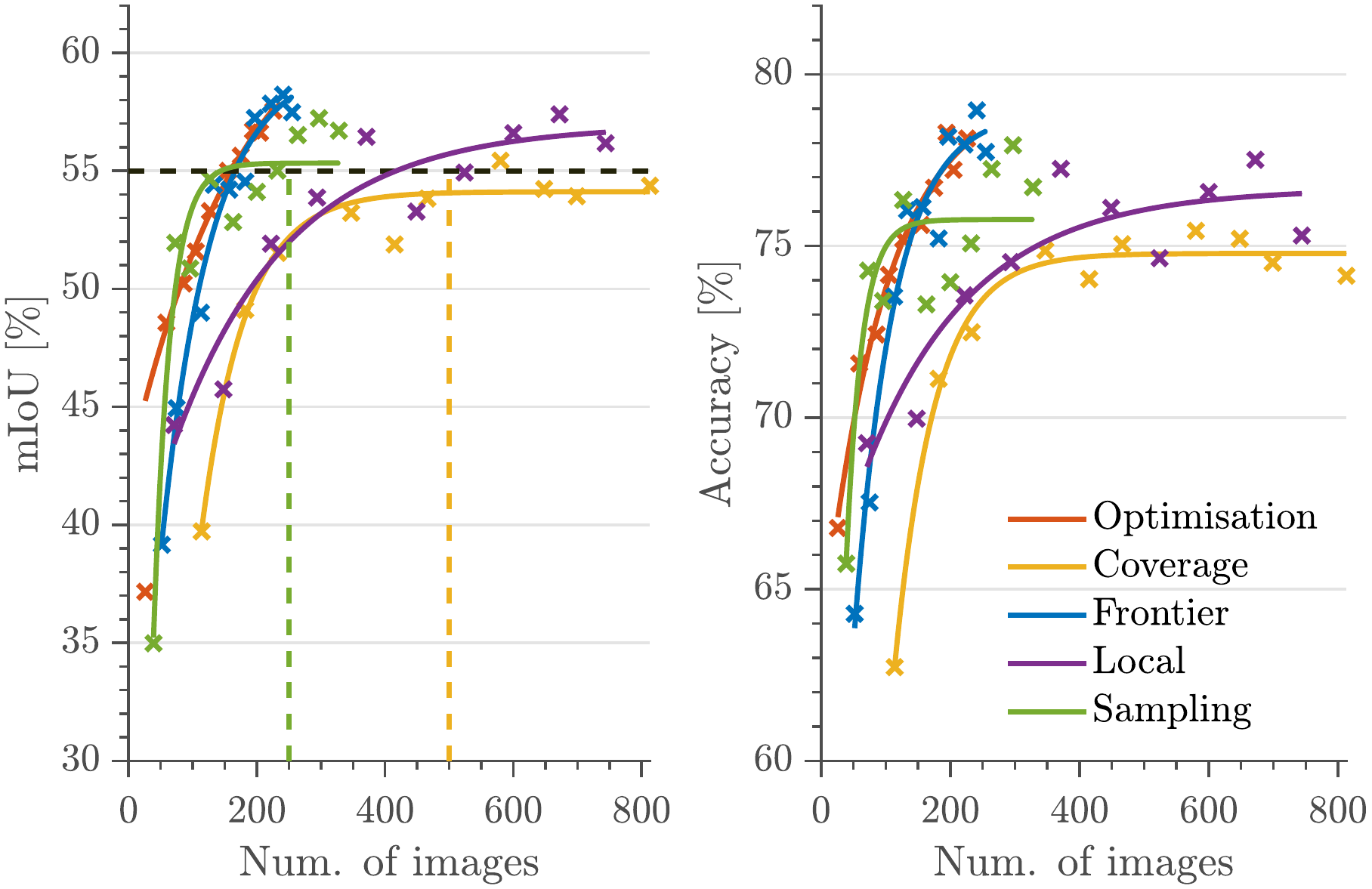}
    \caption{Comparison of \ac{AL} performance with a Bayesian model uncertainty-based planning objective estimated by \ac{MC} dropout and computing informative prior maps before each mission starts. All active planners exceed the coverage baseline's performance (yellow) with less training data as shown by the dashed lines. Our map-based planners outperform the local planner (purple).}
    \label{F:results_potsdam_informed_priors_vs_baselines}
\end{figure}

\begin{figure}[!t]
    \vspace{-2mm}
    \captionsetup[subfigure]{labelformat=empty}
    \centering

    \subfloat[]{\includegraphics[width=0.24\columnwidth]{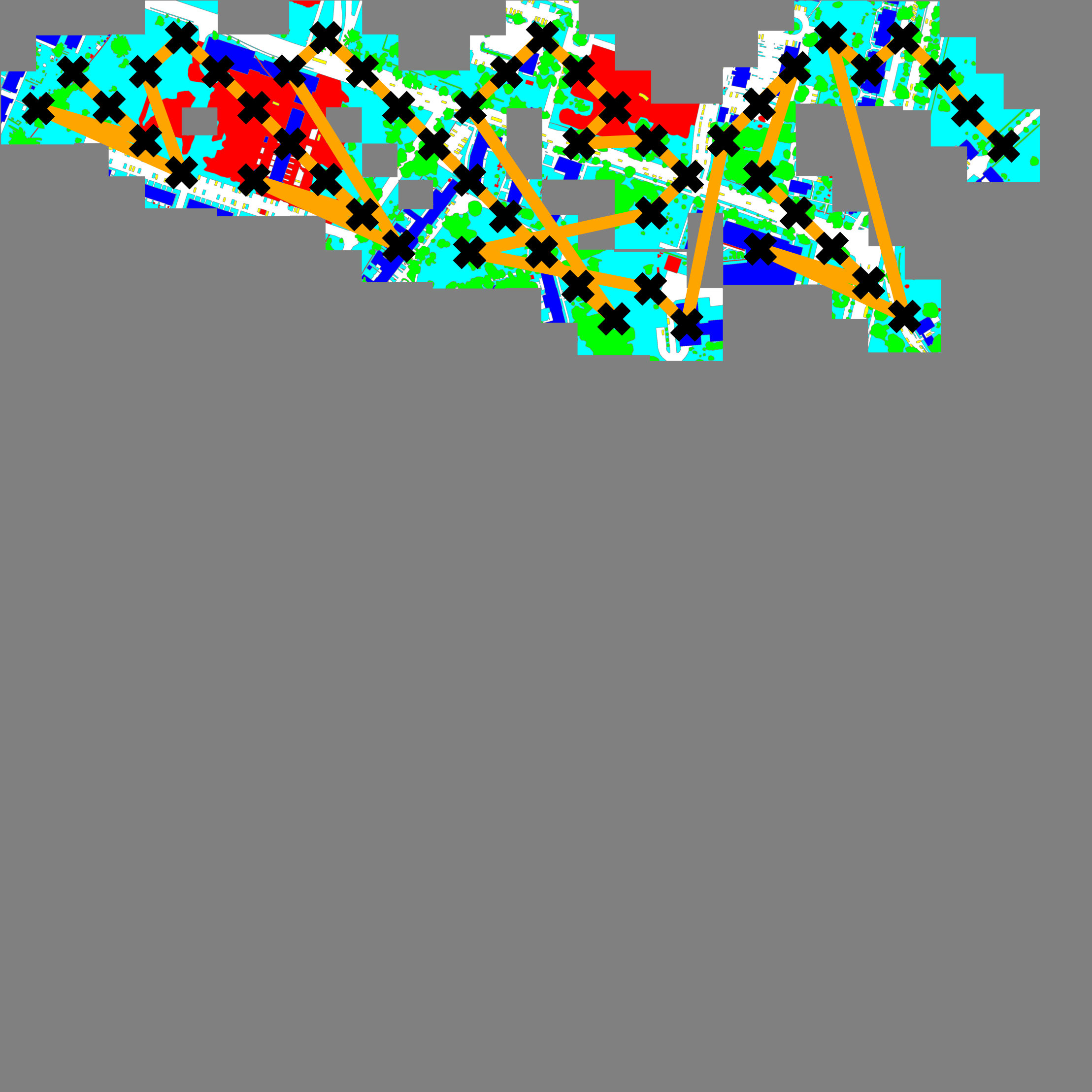}}
    \hfill
    \subfloat[]{\includegraphics[width=0.24\columnwidth]{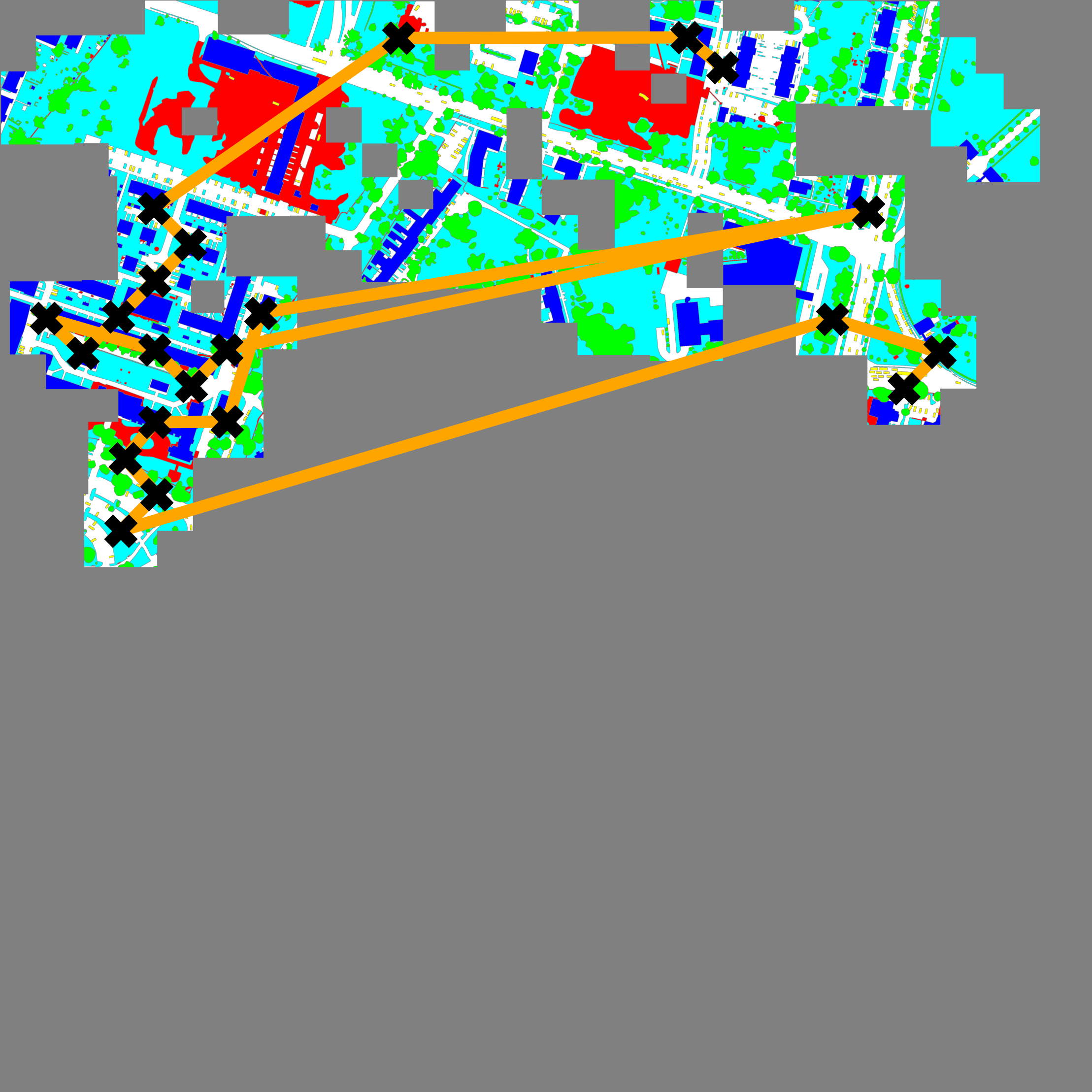}}
    \hfill
    \subfloat[]{\includegraphics[width=0.24\columnwidth]{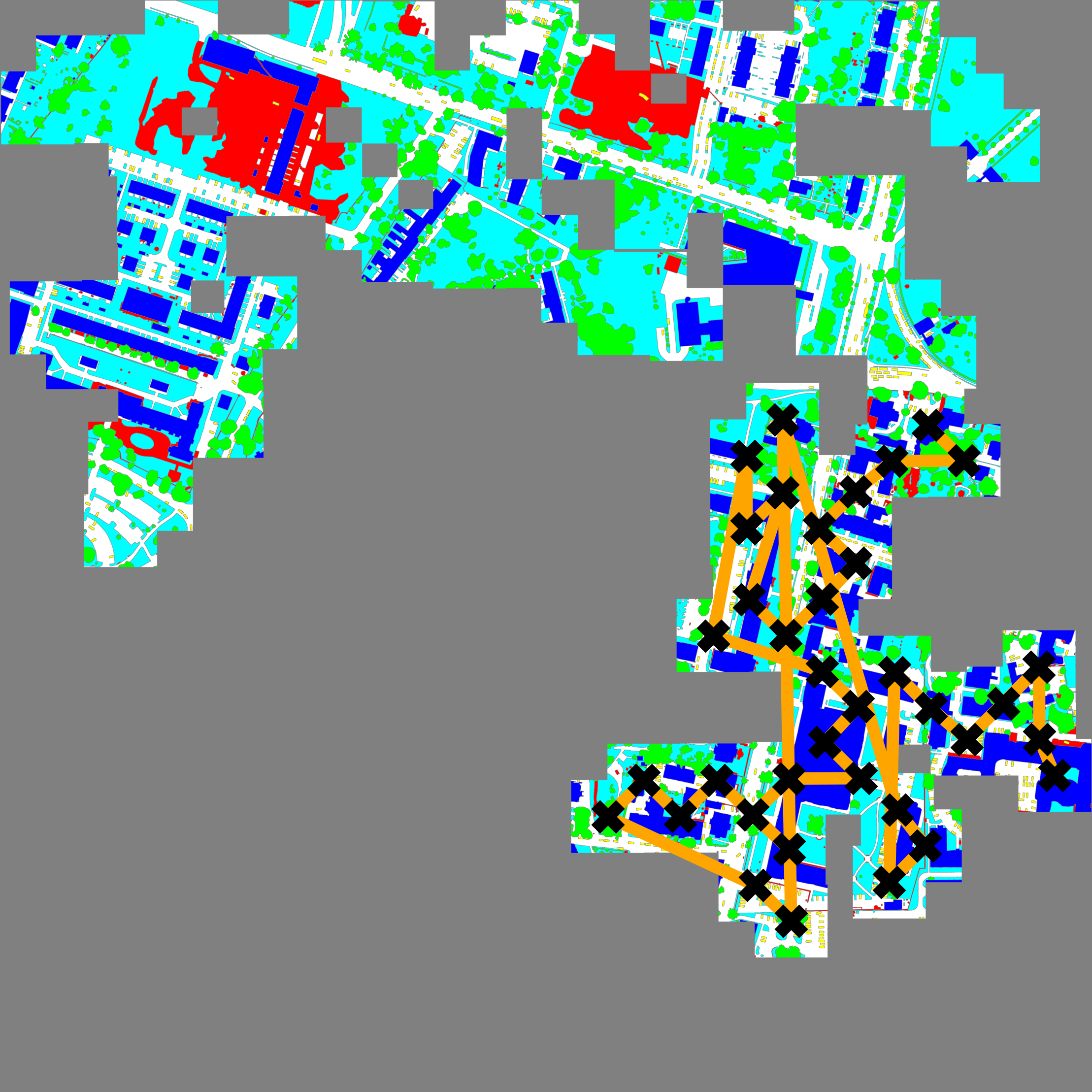}}
    \hfill
    \subfloat[]{\includegraphics[width=0.24\columnwidth]{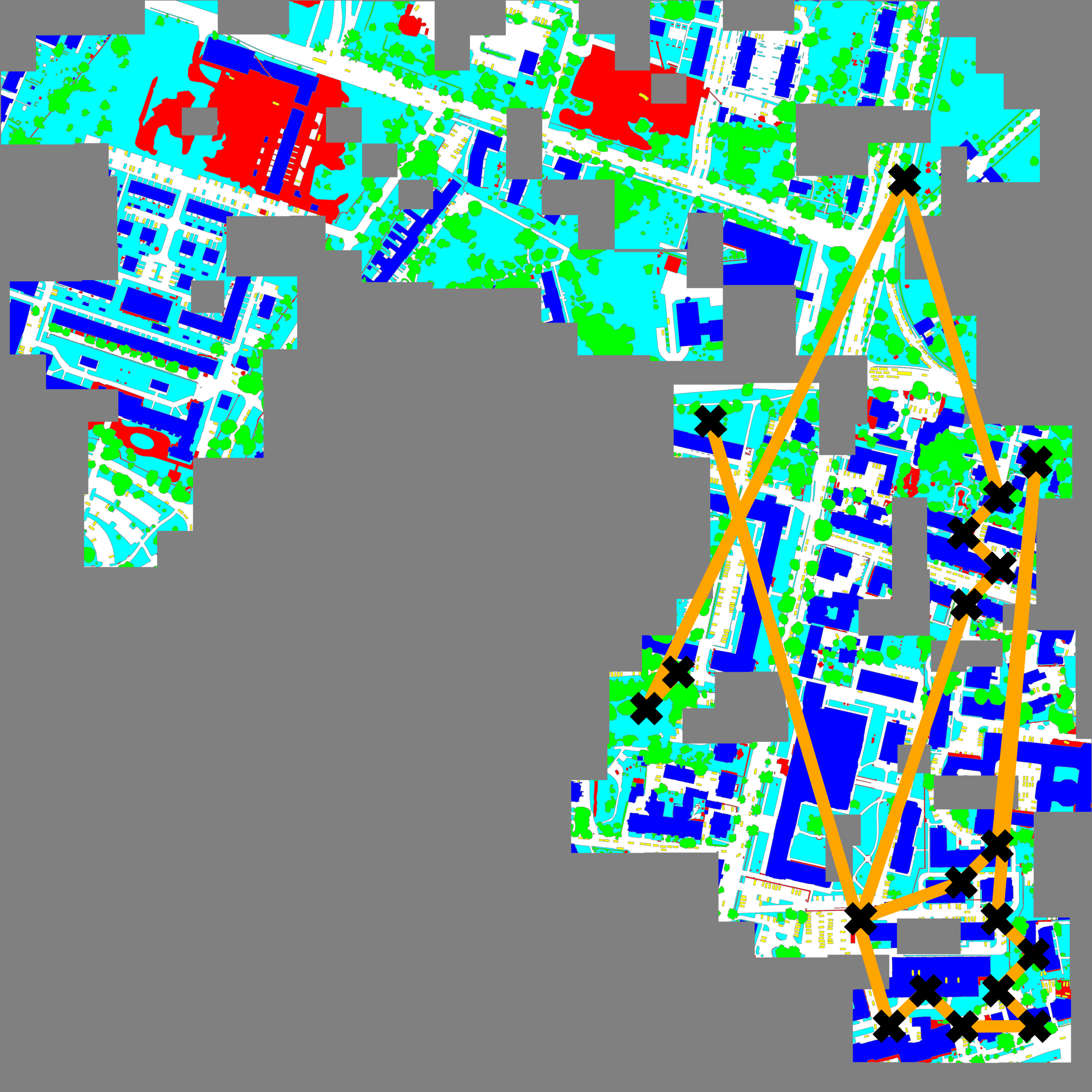}}

    \vspace{-7mm}

    \subfloat[$t=1$]{\includegraphics[width=0.24\columnwidth]{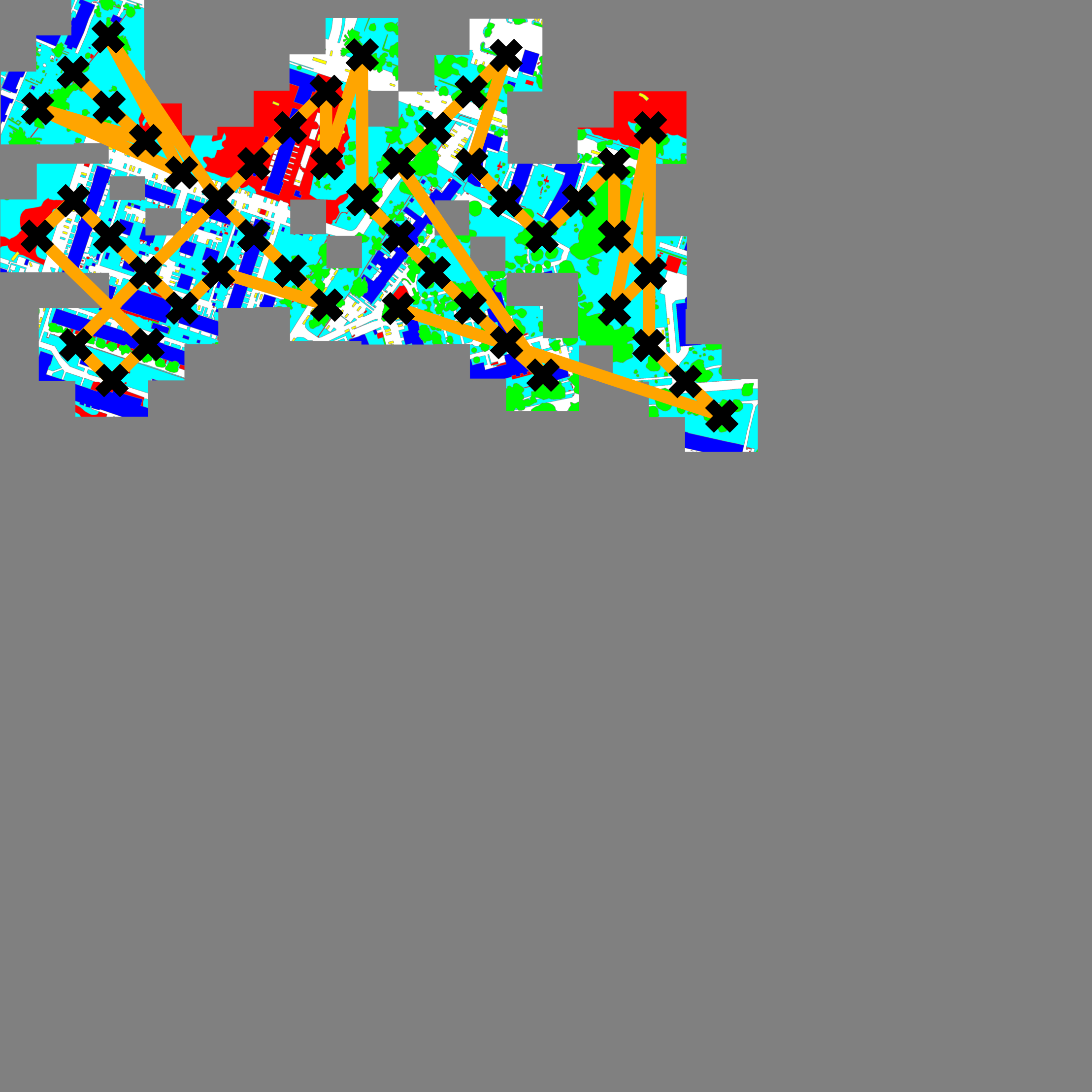}}
    \hfill
    \subfloat[$t=2$]{\includegraphics[width=0.24\columnwidth]{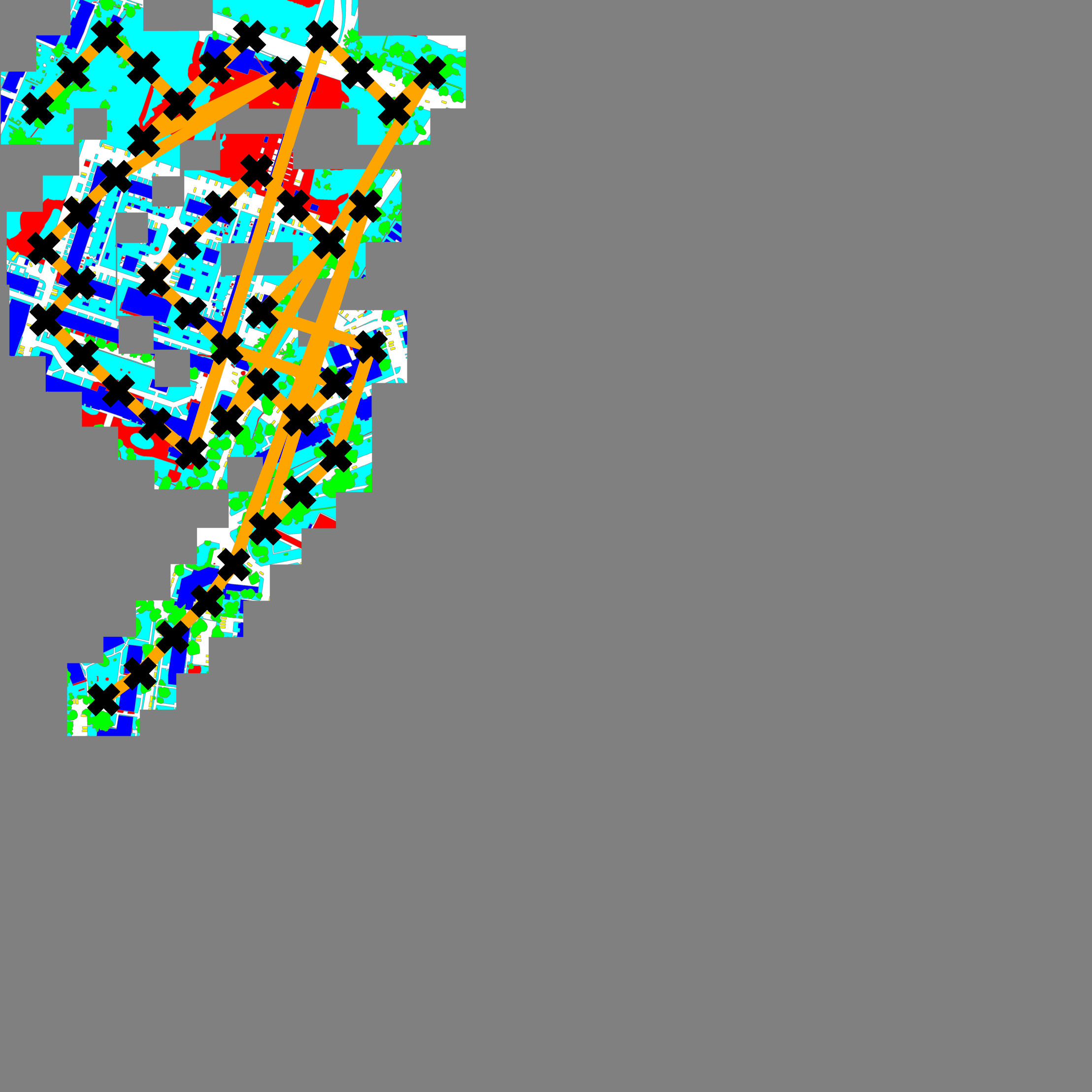}}
    \hfill
    \subfloat[$t=3$]{\includegraphics[width=0.24\columnwidth]{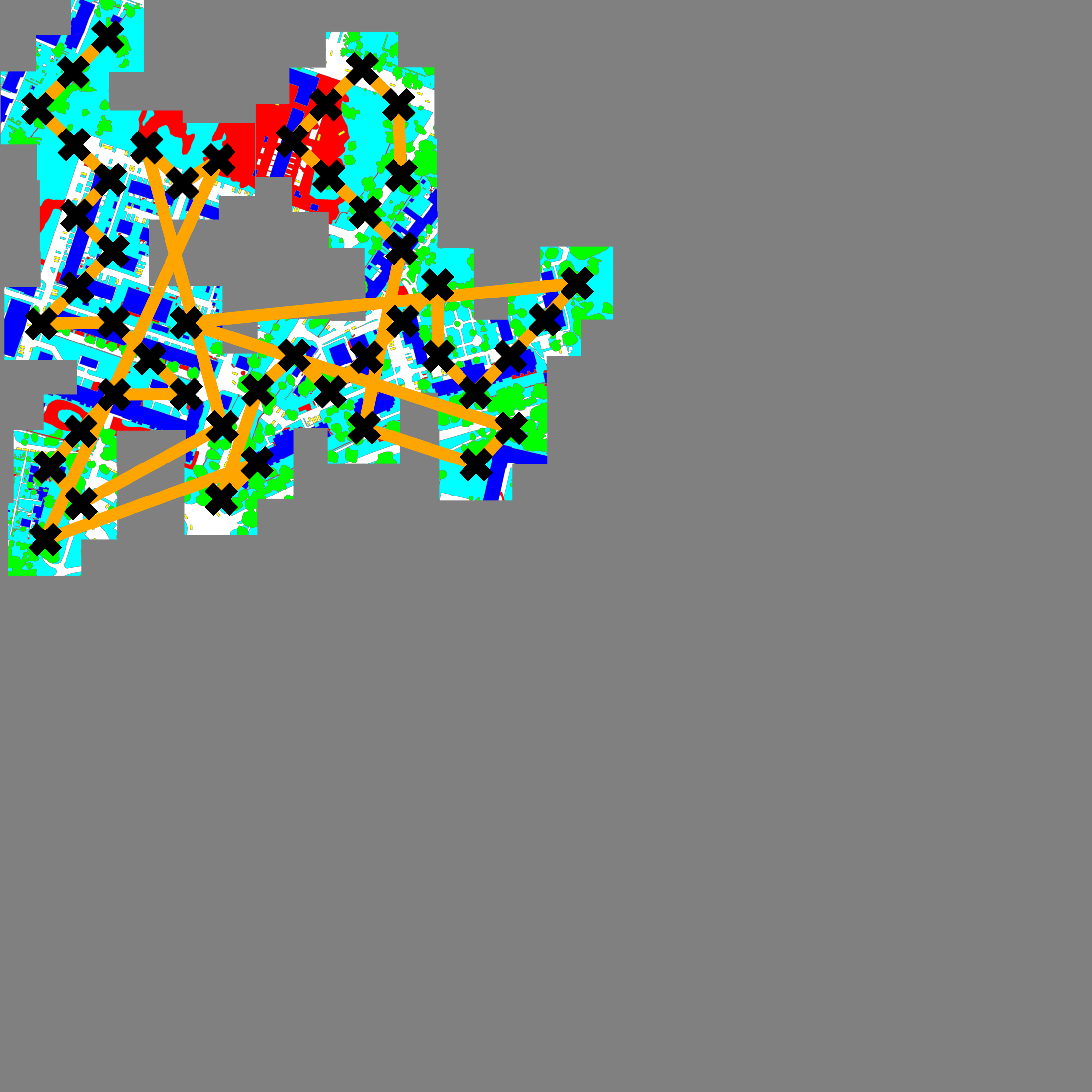}}
    \hfill
    \subfloat[$t=4$]{\includegraphics[width=0.24\columnwidth]{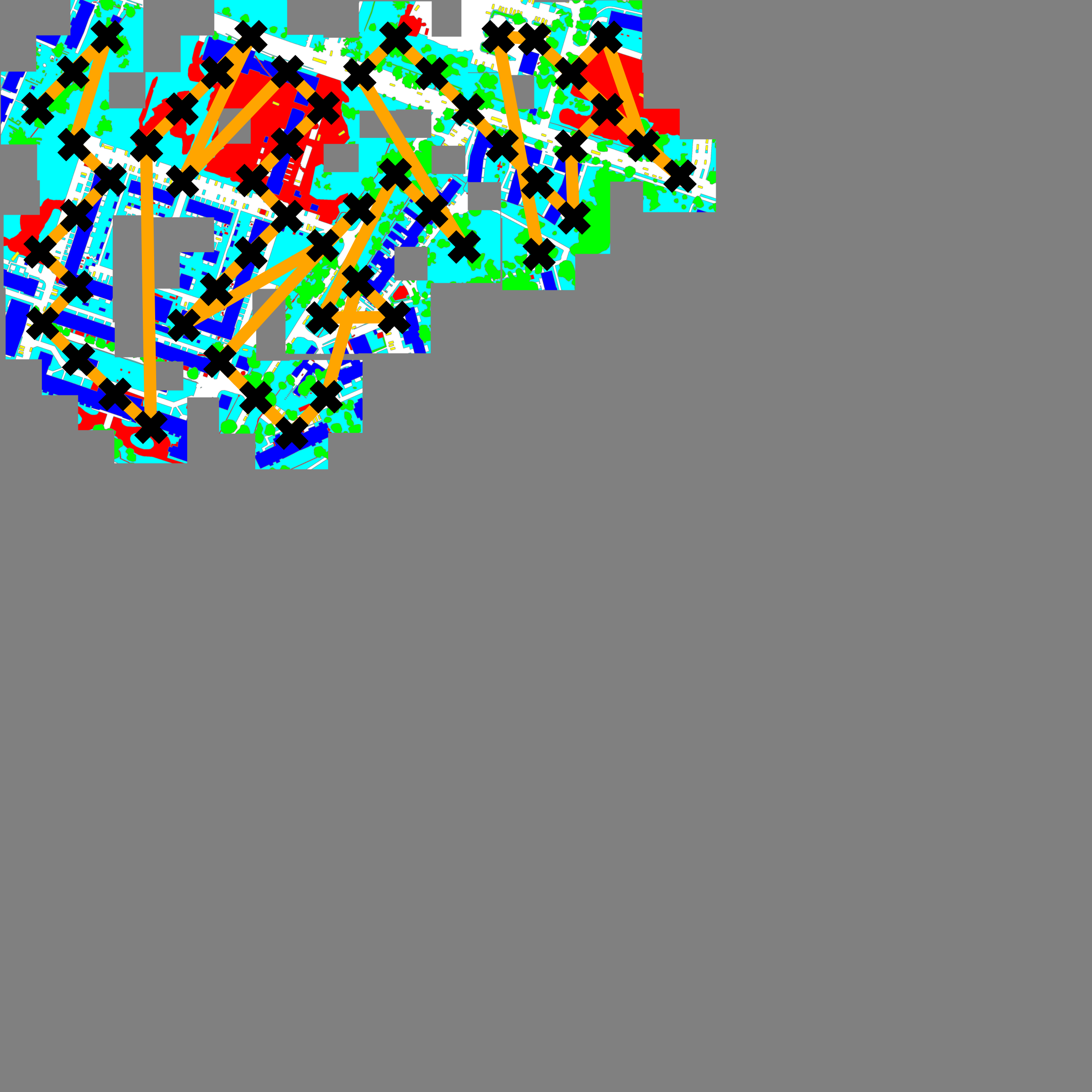}}

\caption{Examples of paths planned on ISPRS Potsdam using the frontier-based planning strategy with (top) and without (bottom) our approach for precomputing informative prior maps. The priors are computed before each of four subsequent missions, with the UAV starting in the top-left corner. As shown by the planned paths (orange lines) and measurement positions (black crosses), using informative priors facilitates spatial exploration across missions and leads to more targeted training data collection within missions.}
\label{F:ablation_mapping_frontier_paths}
\end{figure}

\cref{F:results_potsdam_informed_priors_vs_baselines} summarises the \ac{AL} performance with the informed mapping strategy for each planner. All planners reach higher final prediction performance than the coverage baseline (yellow). This supports the claim that our framework is generally applicable to different planning algorithms. Further, it suggests that active replanning is key to efficiently improving robot vision. Notably, our global map-based planners (orange, blue, green) exceed the coverage baseline's maximum prediction performance ($\approx 55\%$ \ac{mIoU}, black dashed line) after $\leq 250$ labelled images (dashed green line), while the baseline requires $\approx 500$ labelled images (yellow dashed line) to reach this performance. Particularly, for the uncertainty-based objective, our map-based planners show stronger \ac{AL} performance than the local planner (purple) proposed by Blum et al.~\cite{Blum2019} as the map-based planners' performances upper-bound the local planner's performance for any fixed number of labelled images. In contrast to local planning, map-based planners drastically reduce training data requirements and tend to achieve higher final prediction performance.

\begin{figure*}[!t]
    \centering
    \subfloat[Frontier-based planner \label{SF:ablation_mapping_frontier}]{\includegraphics[width=0.32\textwidth]{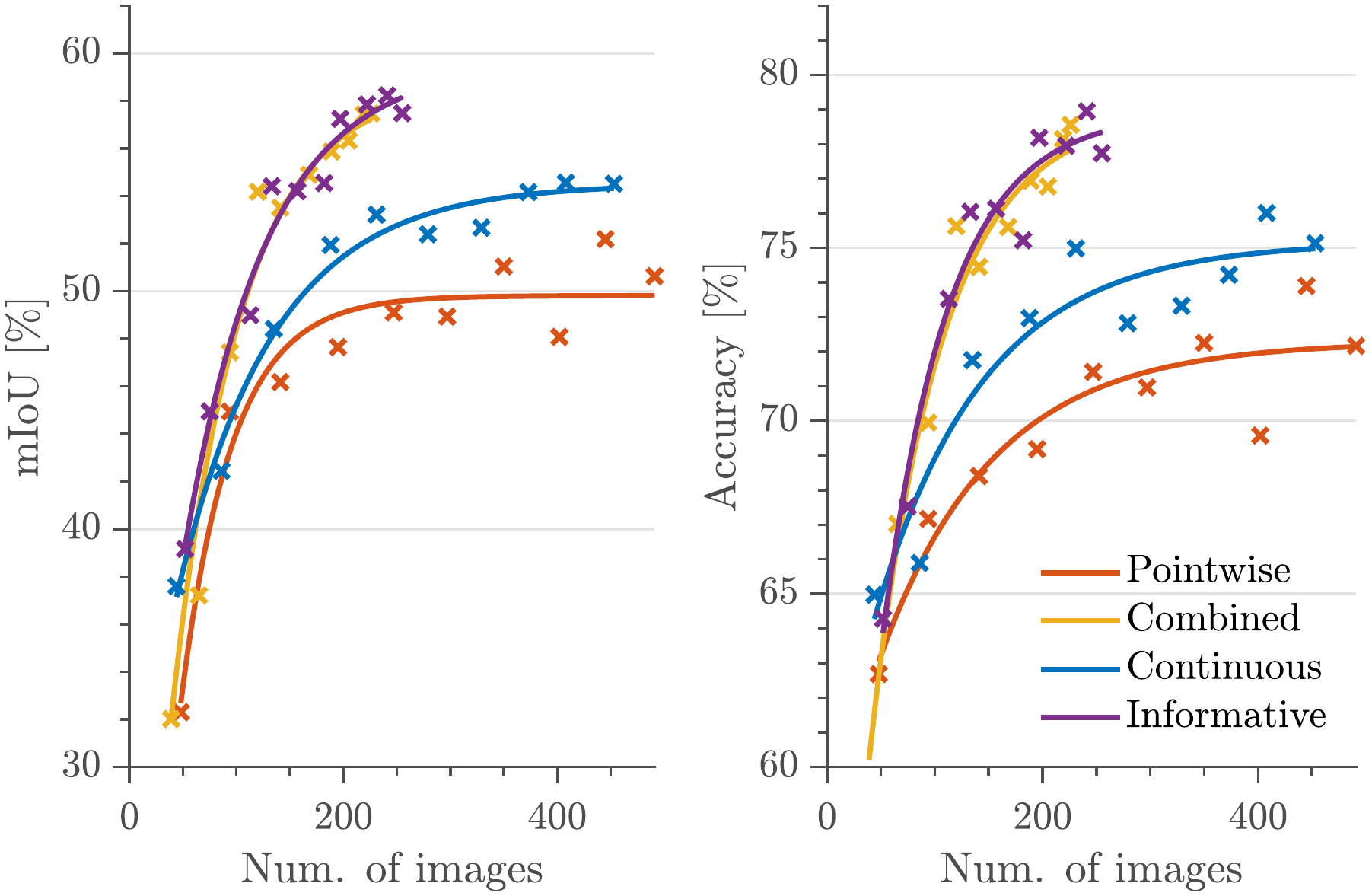}}
    \hfill
    \subfloat[Optimisation-based planner \label{SF:ablation_mapping_cmaes}]{\includegraphics[width=0.32\textwidth]{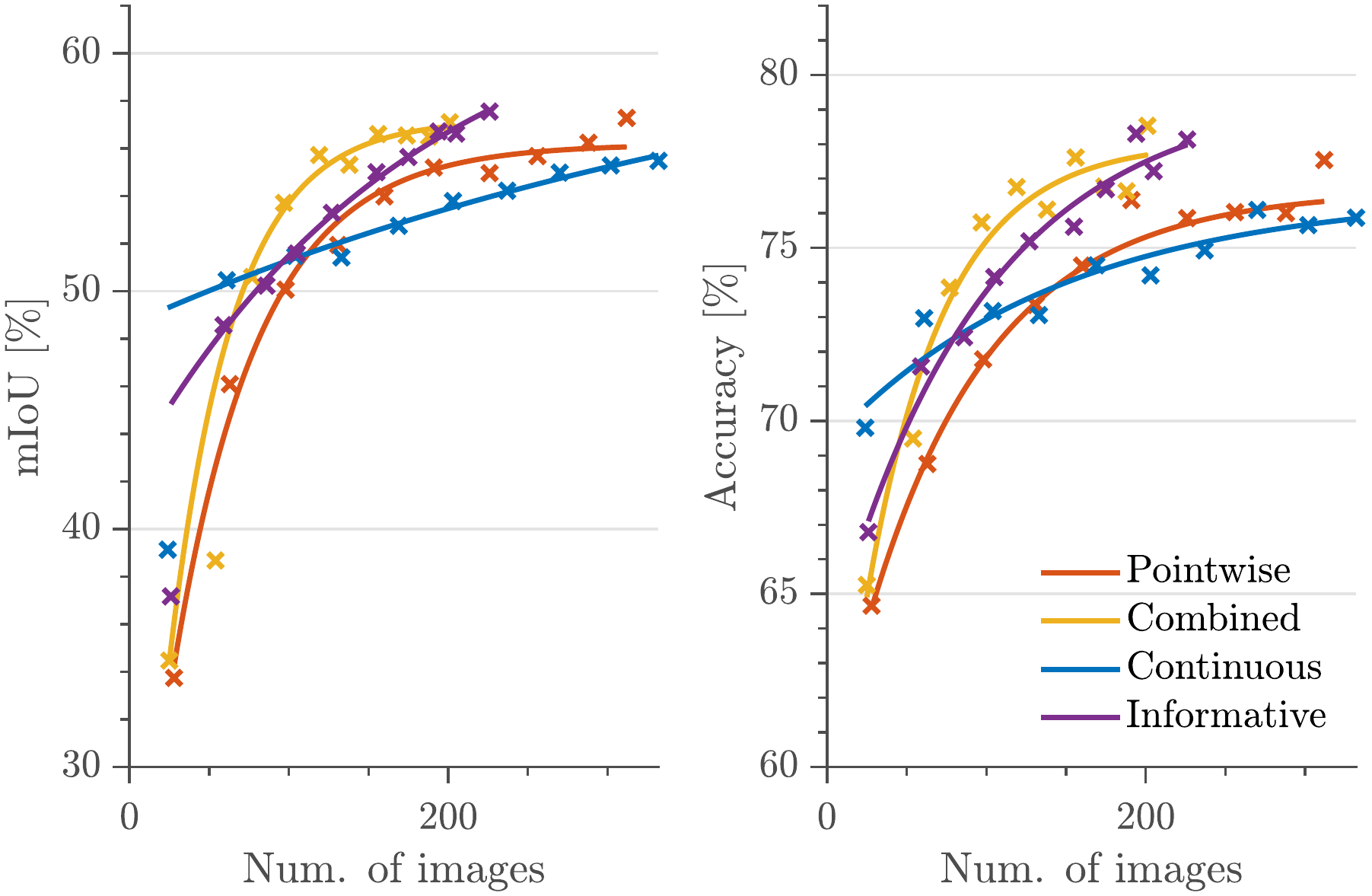}}
    \hfill
    \subfloat[Sampling-based planner \label{SF:ablation_mapping_mcts}]{\includegraphics[width=0.32\textwidth]{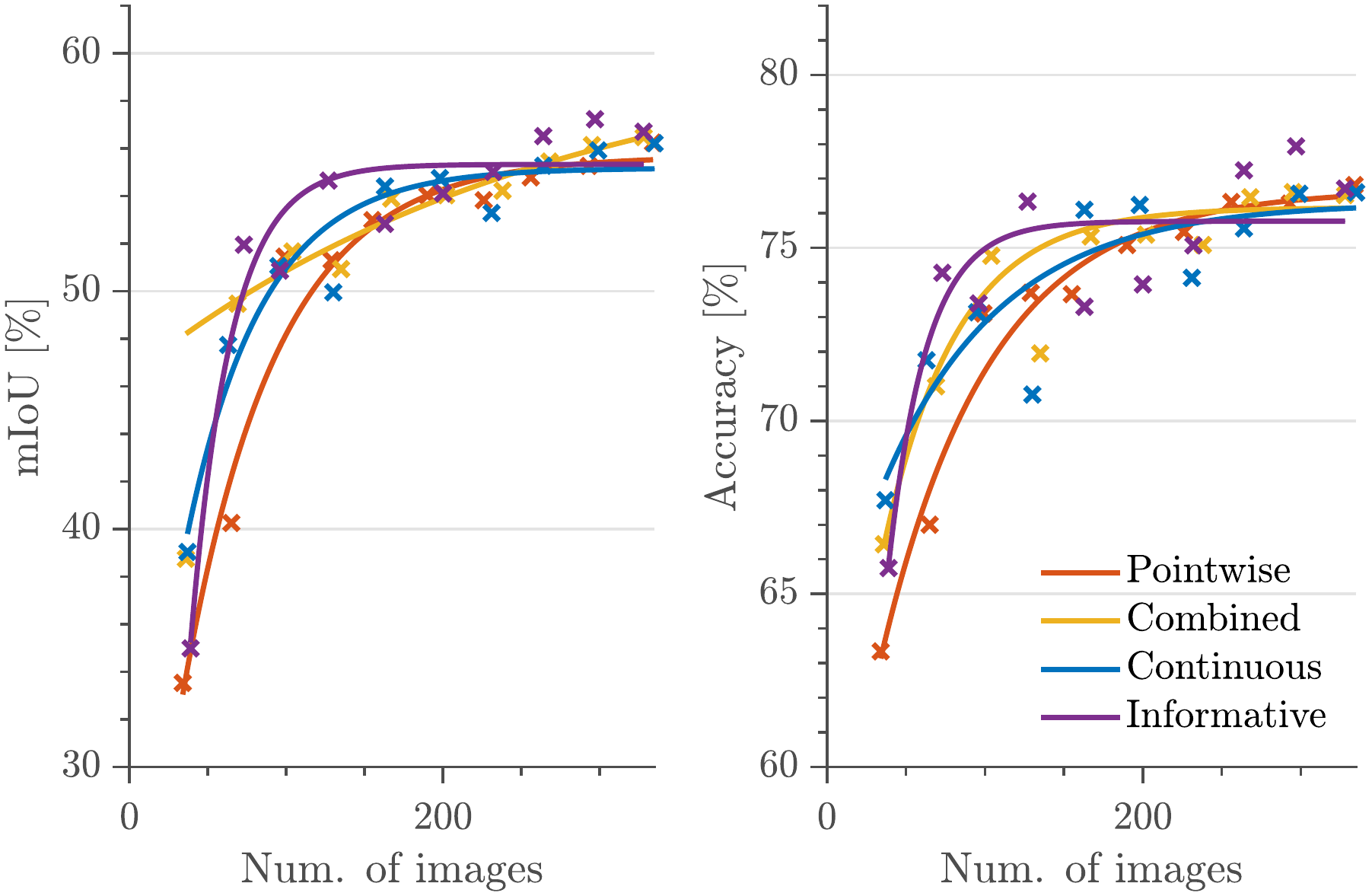}}

    \caption{Comparison of effects of our mapping module using our framework. The planners consistently benefit from recomputing informative map priors before a mission starts (purple). The performance gain of mapping a continuous RGB sensor stream (blue) instead of only mapping images at planned measurement positions (orange) is less significant. Combining both continuous sensor streams and informative map priors (yellow) also leads to consistent performance improvements. In particular, our informative mapping approach drastically improves the greedy frontier-based strategy.}
    \label{F:ablation_mapping}
\end{figure*}

\begin{figure}[!t]
    \includegraphics[width=\columnwidth]{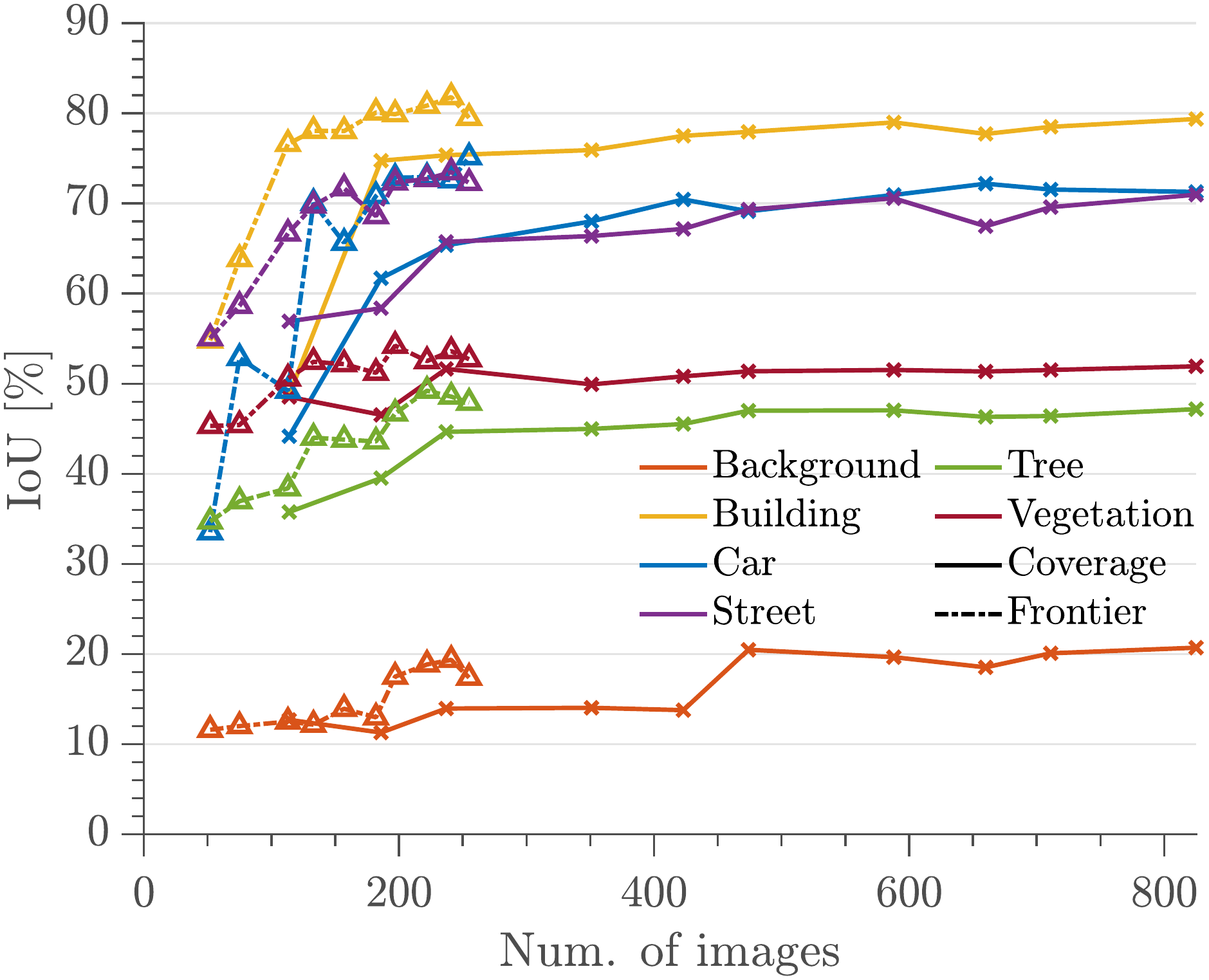}
    \caption{Comparison of per-class \ac{AL} performance of map-based frontier vs. coverage planning with a Bayesian model uncertainty-based planning objective estimated by \ac{MC} dropout and computing informative prior maps before each mission starts. The frontier planner outperforms the coverage baseline (yellow) in almost all classes as our framework can capture complex task-dependent inter-class and intra-class model uncertainties.}
    \label{F:results_potsdam_informed_priors_vs_baselines_per_class}
\end{figure}

To better understand the benefits of our active planning framework, \cref{F:results_potsdam_informed_priors_vs_baselines_per_class} exemplarily compares the per-class \ac{AL} performance of the map-based frontier planner (dashed lines) to the coverage baseline (solid lines) in the ISPRS Potsdam scenario. Our active frontier-based planning strategy shows higher \ac{AL} performance in almost all classes, irrespective of their training data support. Interestingly, the ‘car’ class (blue) has lower training data support than the ‘tree’ (green) and ‘vegetation’ (red) classes but shows stronger IoU performance, even with non-targeted coverage planning. However, active planning improves the ‘car’ prediction performance even faster than the non-targeted baseline showing the benefit of our framework for classes with little training data support. Further, although the ‘tree’, ‘background’ (orange), and ‘vegetation’ classes have high training data support, they are difficult to distinguish as their visual appearance from a top-down view depends on the image resolution, altitude, and season. This leads to challenging predictions, which may be partially attributed to data instead of model uncertainty, which cannot be explained away with more training data~\cite{kendall2017uncertainties}. Thus, not all classes with high training data support benefit to the same extent from active planning. At the same time, although the ‘building’ class (yellow) has high training data support and is reliably detected by both planners, the frontier-based planner still shows faster performance improvement as our framework can account for the differing visual appearance and geometry of office buildings, historical buildings, and townhouses. Overall, the results suggest that our framework can capture complex task-dependent inter-class and intra-class model uncertainties, which are too complex to capture with a single training data support heuristic, leading to superior \ac{AL} performance over non-targeted baselines. 

To support the claim that our new mapping module is important for the planning framework's performance, we perform an ablation study to measure its effect on our map-based planners. We consider two mapping setups where the \ac{UAV} either maps training images at planned measurement positions only (pointwise sensor) or maps the images continuously as it moves (continuous sensor stream). \cref{F:ablation_mapping} displays the \ac{AL} performance of our map-based planners (i) recomputing informative prior maps before each mission starts based on previously collected data and the re-trained network (purple), (ii) mapping a continuous RGB image stream (blue) instead of mapping training images at planned measurement positions only (orange), and (iii) combining both informative prior maps and mapping continuous sensor streams (yellow).

All map-based planners show better \ac{AL} performance with recomputed map priors as they exploit already mapped heterogeneous terrain information. This suggests that mapping and updating knowledge collected across missions with re-trained networks, i.e. changing vision capabilities, is key to strong planning performance. In contrast, mapping more information during a single mission with a fixed network is less crucial. Mapping a continuous image stream instead of mapping training data information at planned measurement positions only leads to performance improvements for the greedy frontier-based planner, while both non-greedy planners do not benefit from mapping more information during a mission. Accordingly, combining both mapping continuous sensor streams and recomputing map priors leads to higher \ac{AL} performance of the frontier-based and optimisation-based planner. The sampling-based planner does not show a performance gain when combining mapping of continuous sensor streams and recomputing map priors. Particularly, our greedy frontier-based planning strategy shows significant improvements by leveraging the informative mapping procedure. Our non-greedy optimisation- and sampling-based planners are more robust to less informed map priors, perhaps because they utilise non-myopic planning along multiple waypoints, while the frontier-based planner only reasons about the next waypoint. Qualitatively, \cref{F:ablation_mapping_frontier_paths} verifies that informative prior maps for frontier-based planning leads to more efficient terrain exploration across missions and targeted data collection within missions resulting in higher model performance with fewer training images.

\subsection{Bayesian ERFNet Ensemble Study} \label{SS:bayesian_ensemble_results}

The second experiment shows that our Bayesian ensemble provides reliable uncertainty estimates for \ac{AL} planning objectives. Moreover, the ensemble achieves higher prediction performance than non-Bayesian and Bayesian ERFNet with \ac{MC} dropout, presented in our prior work~\cite{ruckin2022informative}.

\begin{figure}[!h]
    \captionsetup[subfigure]{labelformat=empty}

    \centering
    \subfloat[]{\includegraphics[width=0.19\columnwidth]{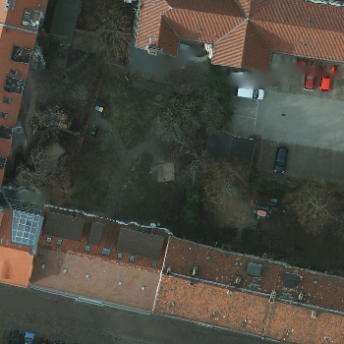}}
    \hfill
    \subfloat[]{\includegraphics[width=0.19\columnwidth]{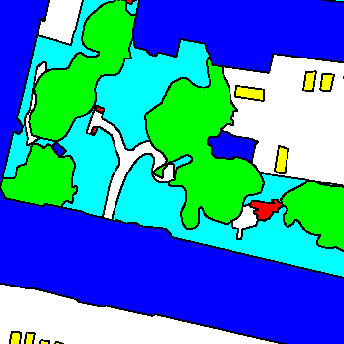}}
    \hfill
    \subfloat[]{\includegraphics[width=0.19\columnwidth]{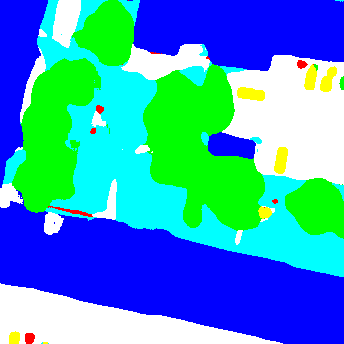}}
    \hfill
    \subfloat[]{\includegraphics[width=0.19\columnwidth]{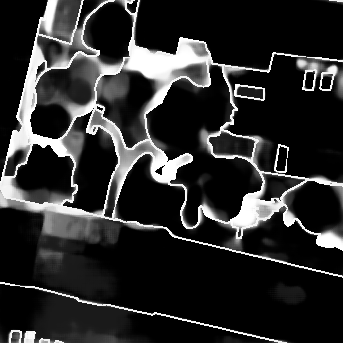}}
    \hfill
    \subfloat[]{\includegraphics[width=0.19\columnwidth]{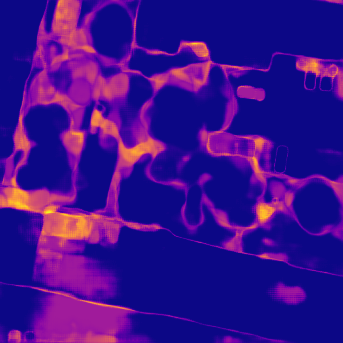}}

    \vspace{-7mm}
    \subfloat[]{\includegraphics[width=0.19\columnwidth]{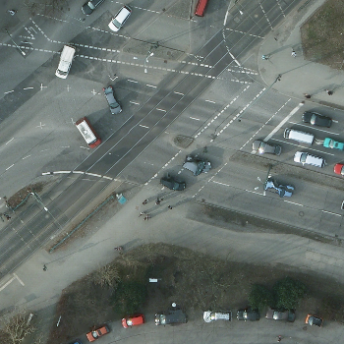}}
    \hfill
    \subfloat[]{\includegraphics[width=0.19\columnwidth]{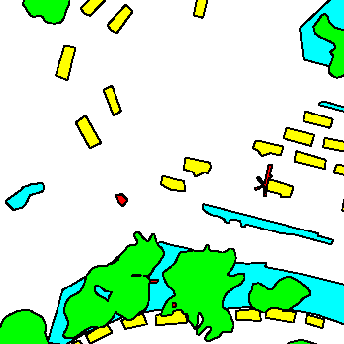}}
    \hfill
    \subfloat[]{\includegraphics[width=0.19\columnwidth]{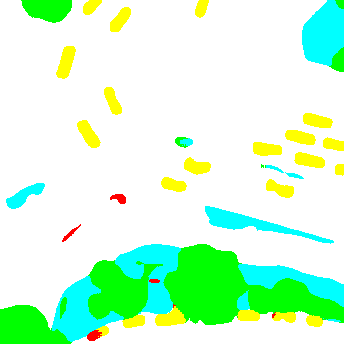}}
    \hfill
    \subfloat[]{\includegraphics[width=0.19\columnwidth]{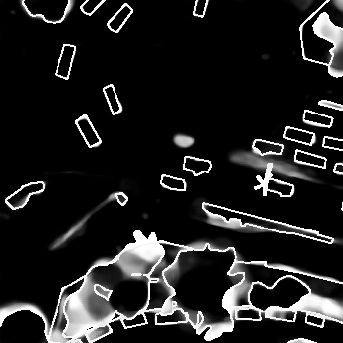}}
    \hfill
    \subfloat[]{\includegraphics[width=0.19\columnwidth]{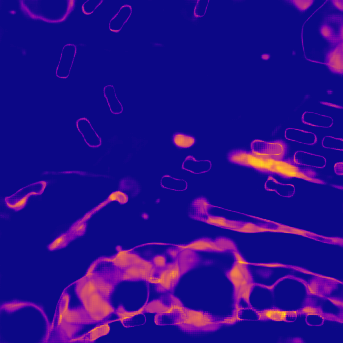}}

    \vspace{-7mm}
    \subfloat[Input]{\includegraphics[width=0.19\columnwidth]{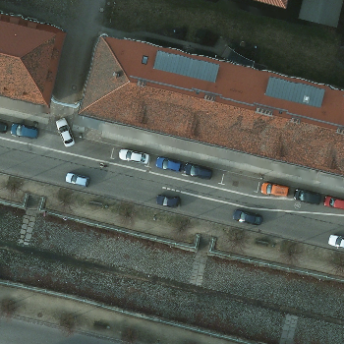}}
    \hfill
    \subfloat[Ground truth]{\includegraphics[width=0.19\columnwidth]{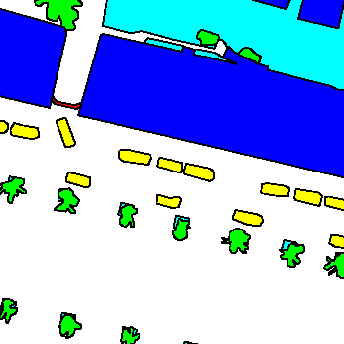}}
    \hfill
    \subfloat[Prediction]{\includegraphics[width=0.19\columnwidth]{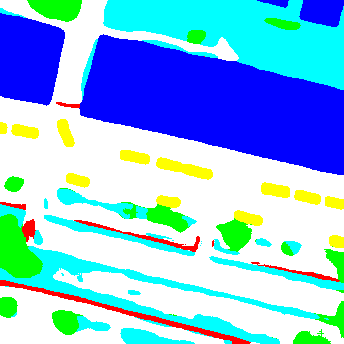}}
    \hfill
    \subfloat[Error]{\includegraphics[width=0.19\columnwidth]{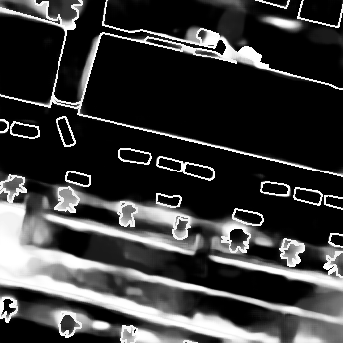}}
    \hfill
    \subfloat[Uncertainty]{\includegraphics[width=0.19\columnwidth]{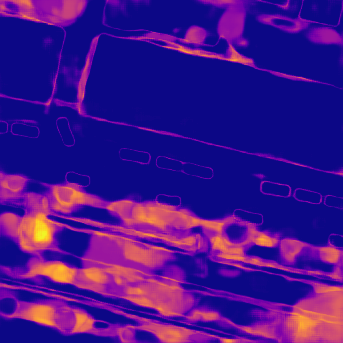}}

\caption{Qualitative results with our ensemble of $T = 8$ ERFNets trained on ISPRS Potsdam. Columns from left to right: RGB input, ground truth, prediction, error image (negative prediction log-likelihood), model uncertainty (\cref{eq:mutual_information}). High model uncertainty (yellower) in misclassified regions (whiter) validates that our ensemble provides consistent uncertainty estimates as a basis for an \ac{AL} planning objective in our framework.}
\label{F:potsdam_ensemble_prediction_examples}
\end{figure}

To confirm that our Bayesian ensemble of ERFNets delivers informative model uncertainties for planning and yields superior prediction performance, we train an ensemble on the ISPRS Potsdam dataset~\cite{Potsdam2018} with $4000$ training images and compare it to the Bayesian ERFNet with \ac{MC} dropout developed in our previous work~\cite{ruckin2022informative} and a deterministic ERFNet. Qualitatively, \cref{F:potsdam_ensemble_prediction_examples} verifies high model uncertainty of our ensemble in misclassified or hard-to-predict regions. Thus, the ensemble's model uncertainties provide reliable information for planning objectives.

\begin{figure}[!t]
    \includegraphics[width=\linewidth]{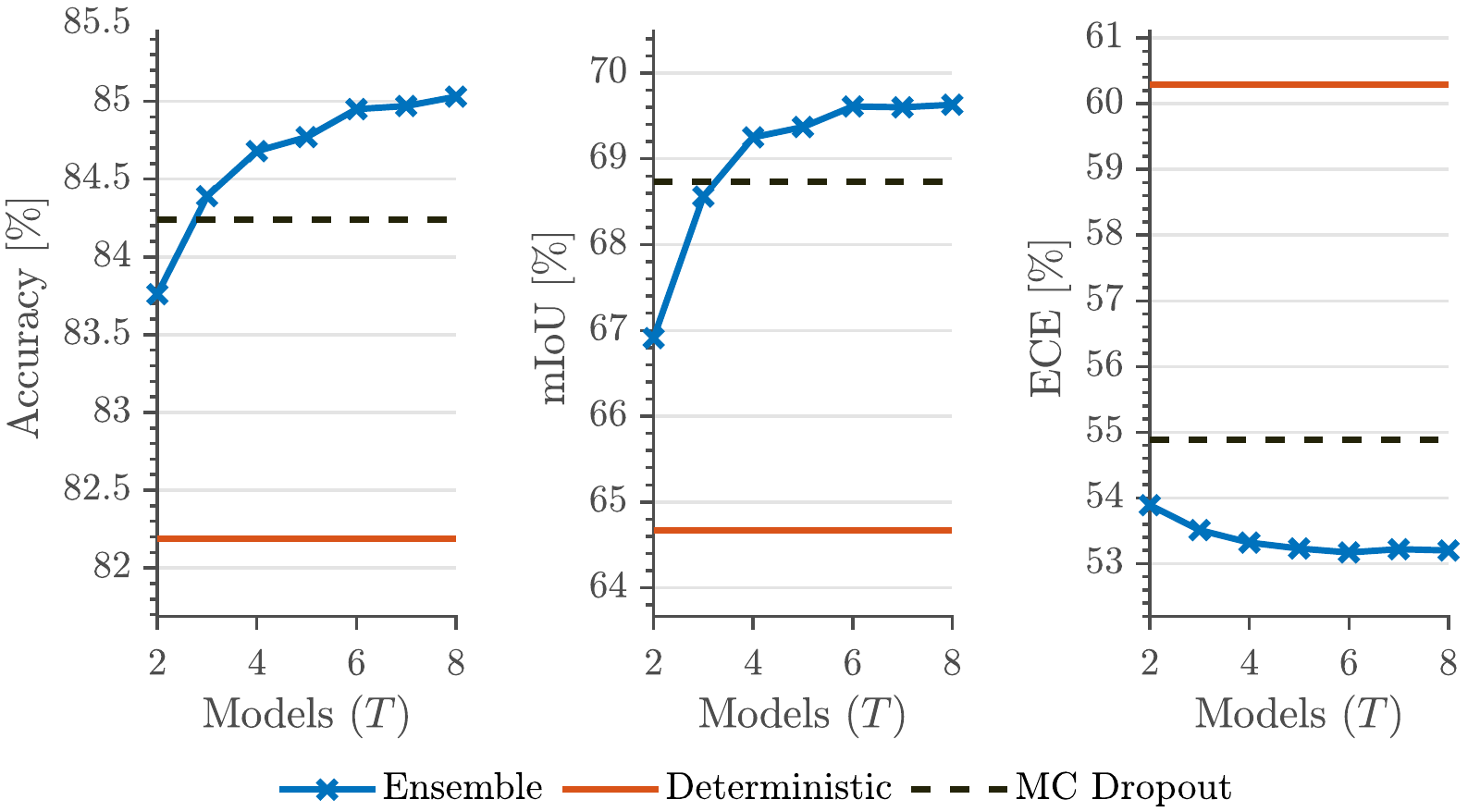}
    \caption{Performance of our Bayesian ensemble with varying number $T$ of ERFNets (blue), deterministic ERFNet (orange), and Bayesian ERFNet with $T = 50$ \ac{MC} dropout samples~\cite{ruckin2022informative}  (black, dashed) on ISPRS Potsdam. For $T = 8$, the ensemble improves mIoU by $4.96 \%$ (middle) and reduces ECE by $7.09 \%$ (right) over a deterministic ERFNet, and improves mIoU by $0.90 \%$ and reduces ECE by $1.68 \%$ over the Bayesian ERFNet with \ac{MC} Dropout.}
    \label{F:results_ablation_ensemble}
\end{figure}

To assess our Bayesian ensemble's prediction capabilities and computational efficiency for online inference on \acp{UAV}, we study its performance with varying numbers of ERFNet models $T = \{2,\, \ldots,\,8\}$ in \cref{F:results_ablation_ensemble}. We compare the ensemble's performance to the deterministic ERFNet~\cite{Romera2018} using a single forward pass and to our Bayesian ERFNet utilising $T = 20$ \ac{MC} dropout samples for converging to maximal performance~\cite{ruckin2022informative}. To quantify the reliability of estimated uncertainties, we measure model calibration using the \ac{ECE} metric~\cite{guo2017calibration}. Intuitively, model calibration is high, i.e. \ac{ECE} low, when the model's probabilistic predictions match its accuracy on a test set.

For $T=8$ models, our ensemble (blue) improves segmentation performance by $4.96 \%$ \ac{mIoU} and \ac{ECE} by $7.09 \%$ over the deterministic ERFNet (orange). Additionally, for $T=8$ models, our ensemble improves segmentation performance by $0.90 \%$ \ac{mIoU} and \ac{ECE} by $1.68 \%$ compared to the Bayesian ERFNet with \ac{MC} dropout. Overall, as the number of models increases, segmentation performance and calibration both improve. Favourably for online inference, with $T \approx 6$ models, performance gains already converge. Further, the Bayesian ensemble performs on par with the Bayesian ERFNet ($T = 20$ \ac{MC} dropout samples) already with $T = 3$ ERFNet models. Thus, our ensemble requires substantially fewer forward passes ($\approx 6\times$), i.e. compute resources, at deployment to achieve the same performance. At train time, the ensemble's compute requirements scale linearly with the number of models $T$, while the \ac{MC} dropout Bayesian ERFNet has constant compute requirements. However, training is performed offline, hence it is not time-critical. For details about efficient ensemble training, we refer to Huan et al.~\cite{huang2017snapshot}.

\subsection{Comparison of Planning Objectives} \label{SS:planning_objectives_results}

Our third set of experiments shows that Bayesian model uncertainty-based objectives guarantee strong \ac{AL} performance irrespective of the uncertainty estimation technique. Further, it verifies that our general framework supports various \ac{AL} acquisition function paradigms, including representation-based and uncertainty-based objectives.

\begin{figure}[!t]
    \includegraphics[width=\linewidth]{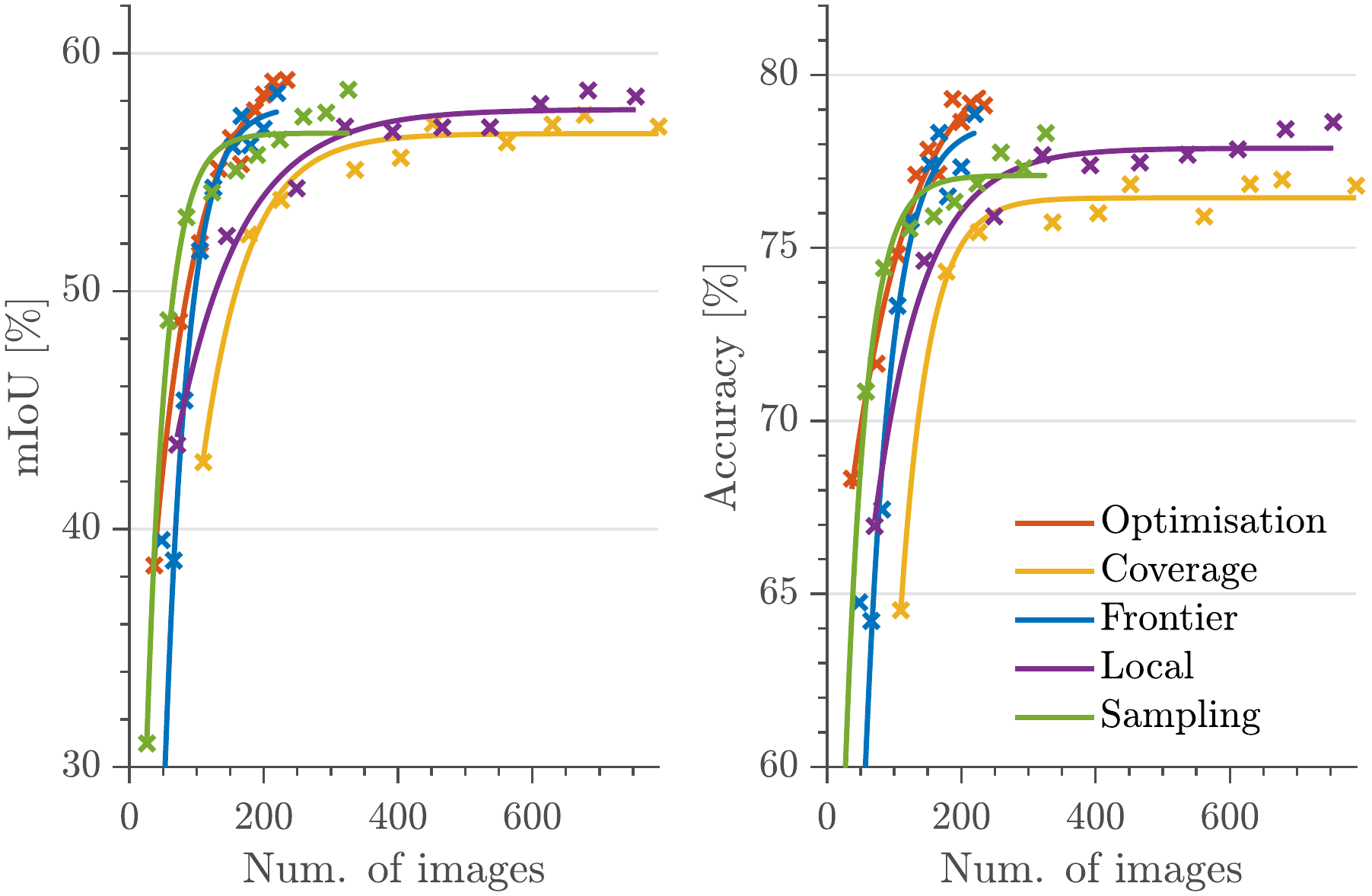}
    \caption{Comparison of \ac{AL} performance with a Bayesian model uncertainty-based planning objective estimated by an ensemble of $T = 4$ ERFNets and computing informative prior maps before each mission starts. All active planners exceed the coverage baseline's performance with less training data. Our global map-based planners outperform the local planning scheme.}
    \label{F:results_potsdam_ensembles_vs_baselines}
\end{figure}

As we show in this experiment, Bayesian model uncertainty-based planning objectives outperform baselines with different uncertainty estimation techniques. We investigate our Bayesian ensemble's \ac{AL} performance on the ISPRS Potsdam dataset. For a fair assessment, we evaluate the coverage baseline with ensemble inference. \cref{F:results_potsdam_ensembles_vs_baselines} summarises the results using our Bayesian ensemble of $T = 4$ ERFNets for all planning approaches. All planners show better performance than the coverage baseline (yellow), which confirms the intuition that active planning for \ac{AL} benefits from Bayesian model uncertainty-based objective functions. Similar to our \ac{MC} dropout-based uncertainty estimation in \cref{F:results_potsdam_informed_priors_vs_baselines}, map-based planners (orange, blue, green) achieve higher prediction performance with fewer training images compared to the local planner (purple), further illustrating the advantage of map-based planning in our framework.

\begin{figure}[!t]
    \captionsetup[subfigure]{labelformat=empty}

    \centering
    \subfloat[]{\includegraphics[width=0.19\columnwidth]{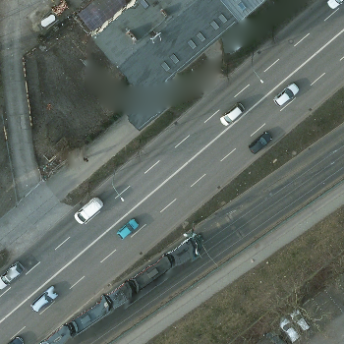}}
    \hfill
    \subfloat[]{\includegraphics[width=0.19\columnwidth]{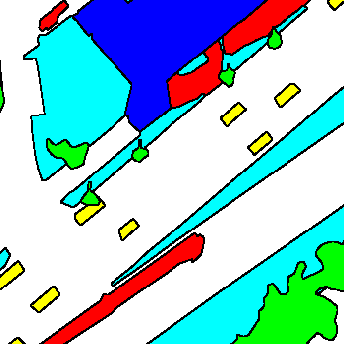}}
    \hfill
    \subfloat[]{\includegraphics[width=0.19\columnwidth]{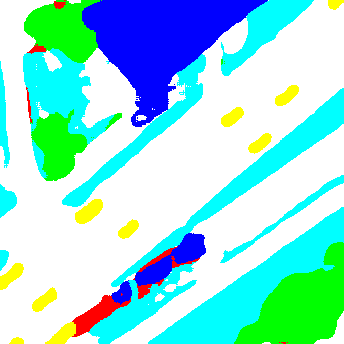}}
    \hfill
    \subfloat[]{\includegraphics[width=0.19\columnwidth]{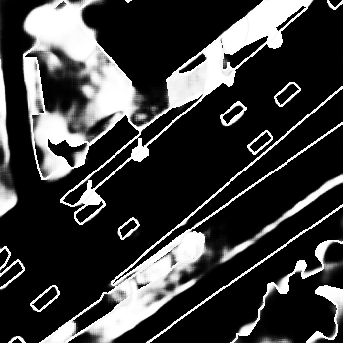}}
    \hfill
    \subfloat[]{\includegraphics[width=0.19\columnwidth]{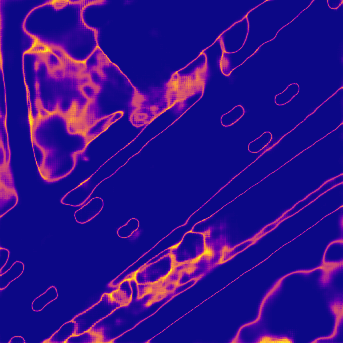}}

    \vspace{-7mm}
    \subfloat[]{\includegraphics[width=0.19\columnwidth]{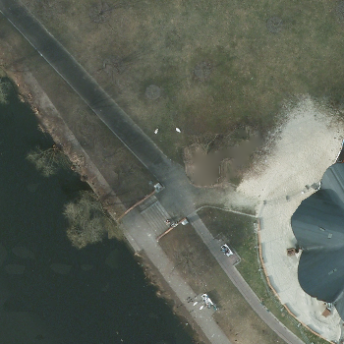}}
    \hfill
    \subfloat[]{\includegraphics[width=0.19\columnwidth]{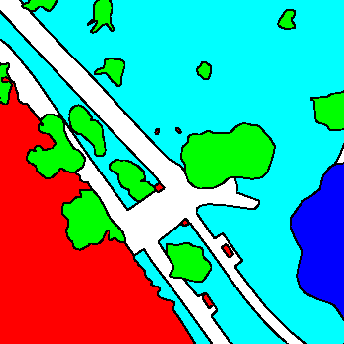}}
    \hfill
    \subfloat[]{\includegraphics[width=0.19\columnwidth]{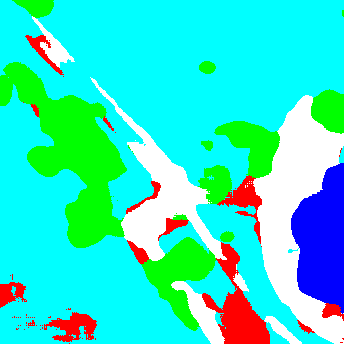}}
    \hfill
    \subfloat[]{\includegraphics[width=0.19\columnwidth]{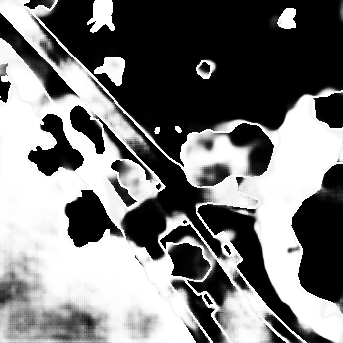}}
    \hfill
    \subfloat[]{\includegraphics[width=0.19\columnwidth]{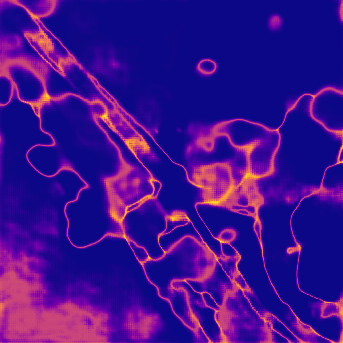}}

    \vspace{-7mm}
    \subfloat[Input]{\includegraphics[width=0.19\columnwidth]{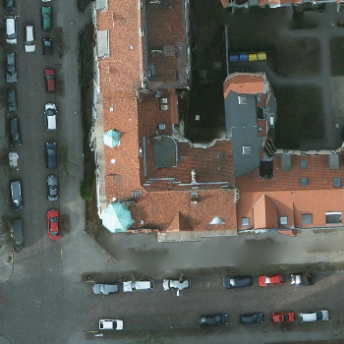}}
    \hfill
    \subfloat[Ground truth]{\includegraphics[width=0.19\columnwidth]{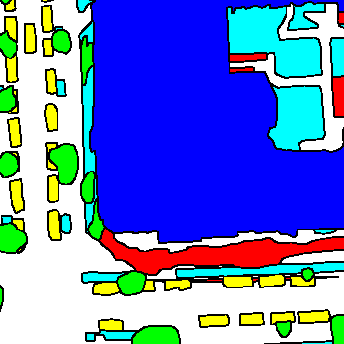}}
    \hfill
    \subfloat[Prediction]{\includegraphics[width=0.19\columnwidth]{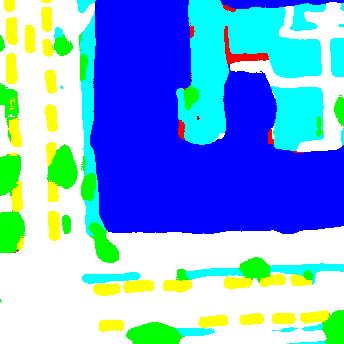}}
    \hfill
    \subfloat[Error]{\includegraphics[width=0.19\columnwidth]{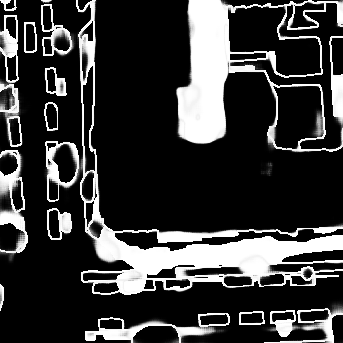}}
    \hfill
    \subfloat[Uncertainty]{\includegraphics[width=0.19\columnwidth]{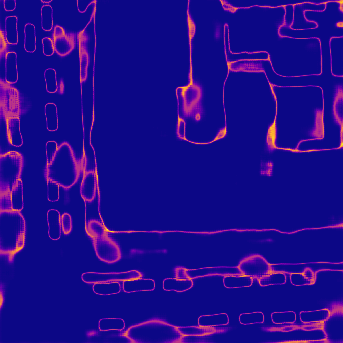}}

\caption{Qualitative results of a deterministic ERFNet trained on ISPRS Potsdam. Columns from left to right: RGB input, ground truth, prediction, error image (negative prediction log-likelihood), prediction entropy. High entropy (yellower) is only weakly correlated with high errors (whiter).}
\label{F:potsdam_deterministic_uncertainty_prediction_examples}
\end{figure}

To further support our framework's generality under various uncertainty-based objective functions, we investigate its performance using a classical non-Bayesian entropy-based acquisition function~\cite{gal2017deep, postels2021practicality}. Given an image $\bm{z}$ and a model with deterministic parameters $\bm{W}$, $p(\bm{y}\,|\,\bm{z}, \bm{W})$ is the maximum likelihood estimate over labels $\bm{y}$. Then, the prediction entropy

\begin{equation}
    \mathbb{H}(\bm{y}) = -p(\bm{y}\,|\,\bm{z}, \bm{W})^{\top} \text{log}\big(p(\bm{y}\,|\,\bm{z}, \bm{W})\big)
\end{equation}

is highest when the prediction is uniform, i.e. most uncertain. Qualitatively, \cref{F:potsdam_deterministic_uncertainty_prediction_examples} shows that non-Bayesian entropy is weakly correlated with prediction errors as it fails to estimate globally calibrated uncertainties.

\begin{figure}[!t]
    \includegraphics[width=\linewidth]{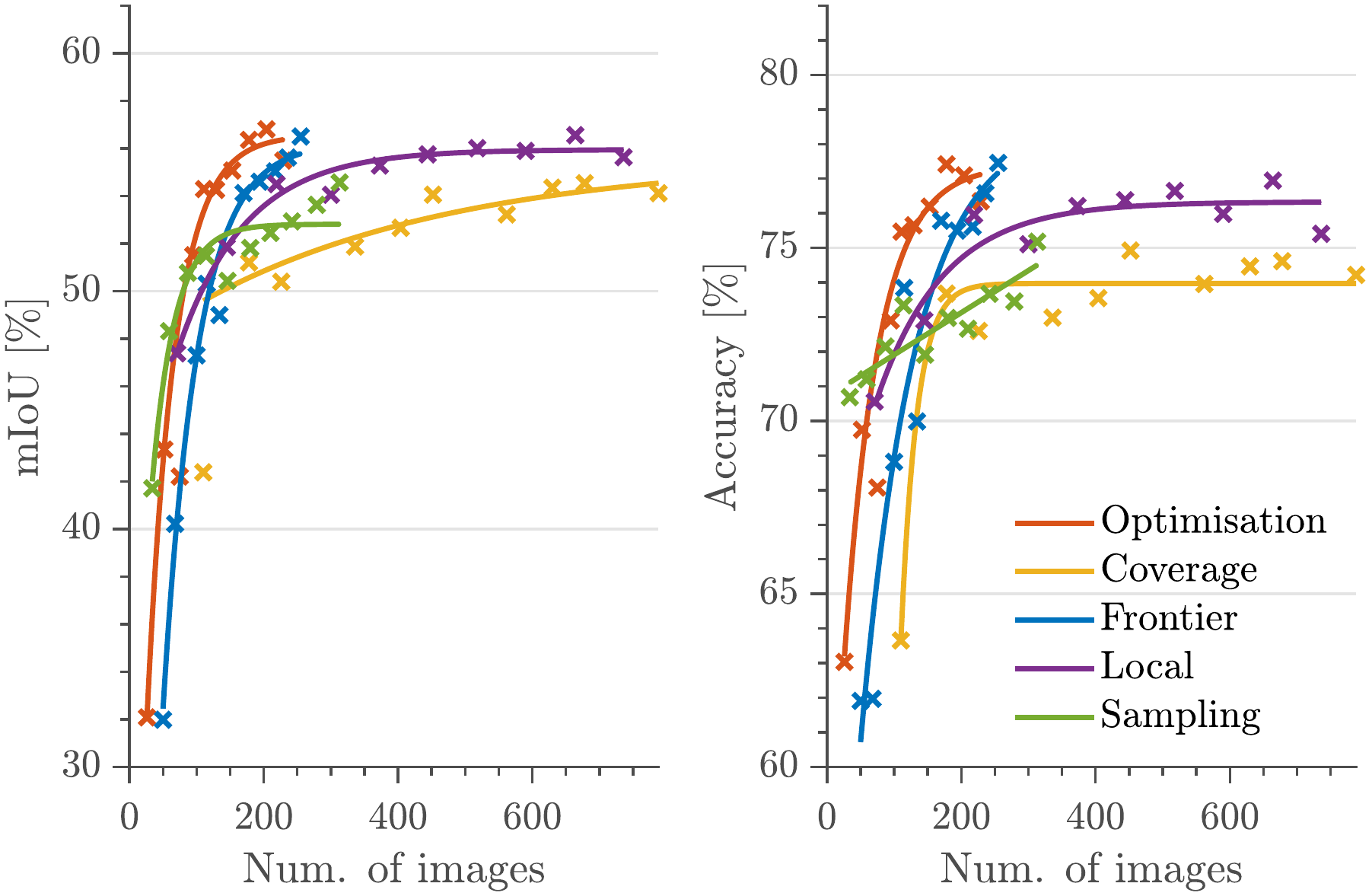}
    \caption{Comparison of \ac{AL} performance with a non-Bayesian entropy-based planning objective over a deterministic forward pass and computing informative prior maps before each mission starts. The frontier-based, optimisation-based, and local planner outperform the coverage baseline's \ac{AL} performance.}
    \label{F:results_potsdam_deterministic_vs_baselines}
\end{figure}

\begin{figure}[!t]
    \centering

    \subfloat[Frontier \label{F:ablation_uncertainty_frontier}]{\includegraphics[width=0.32\columnwidth]{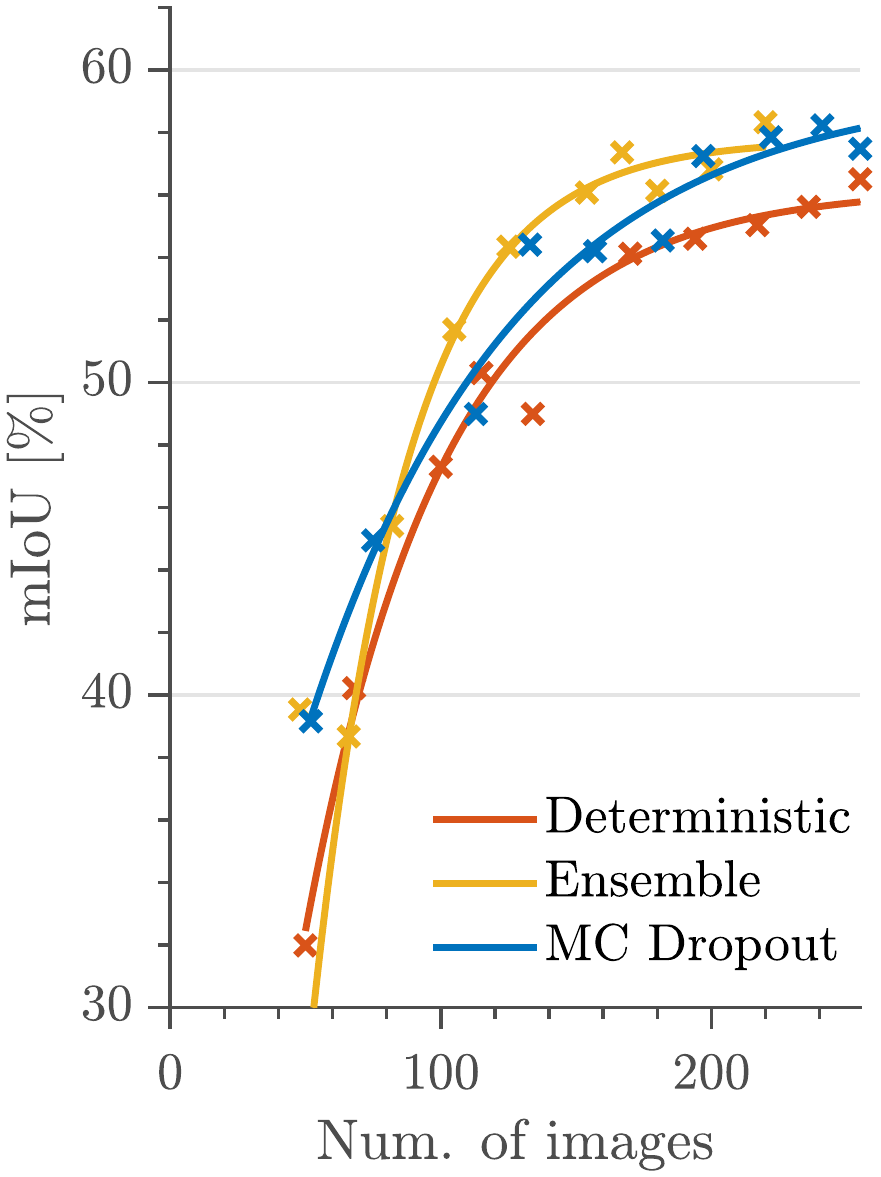}}
    \hfill
    \subfloat[Optimisation \label{F:ablation_uncertainty_cmaes}]{\includegraphics[width=0.32\columnwidth]{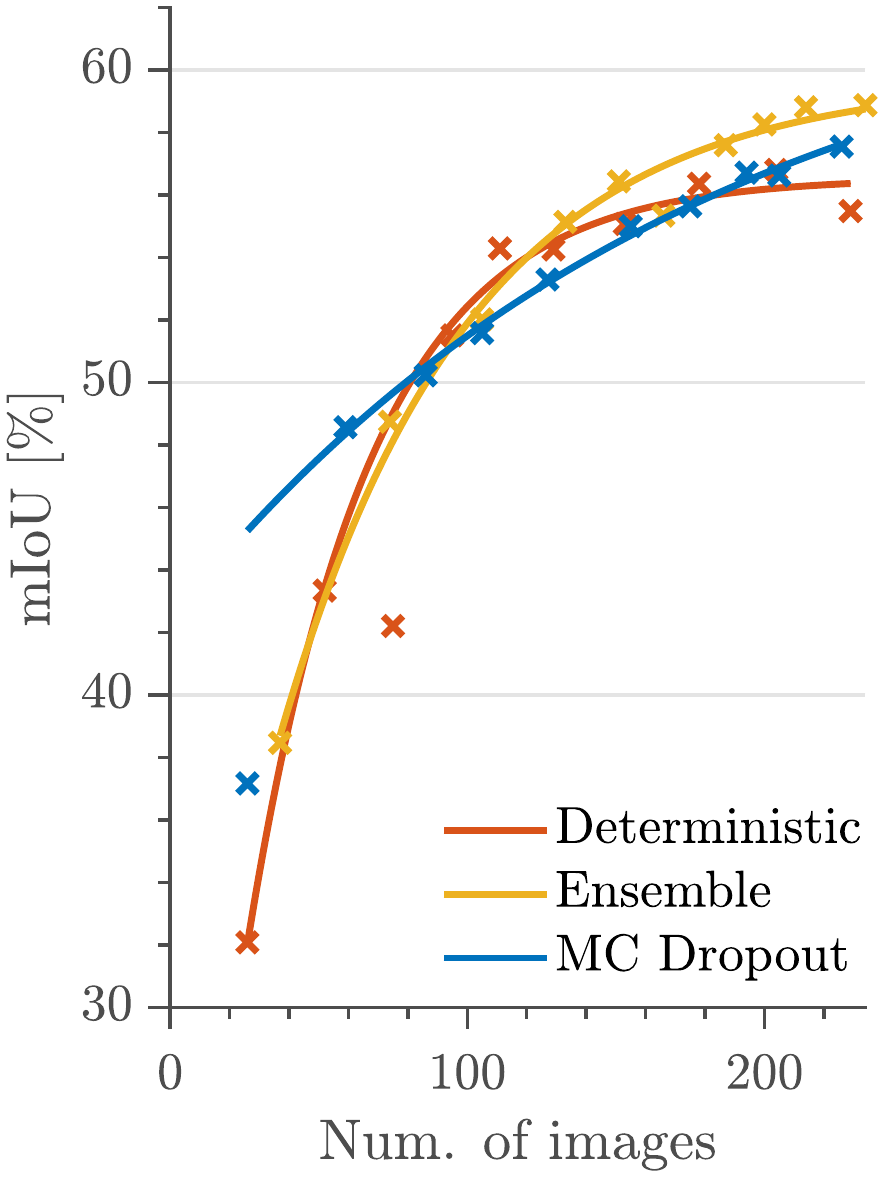}}
    \hfill
    \subfloat[Sampling \label{F:ablation_uncertainty_mcts}]{\includegraphics[width=0.32\columnwidth]{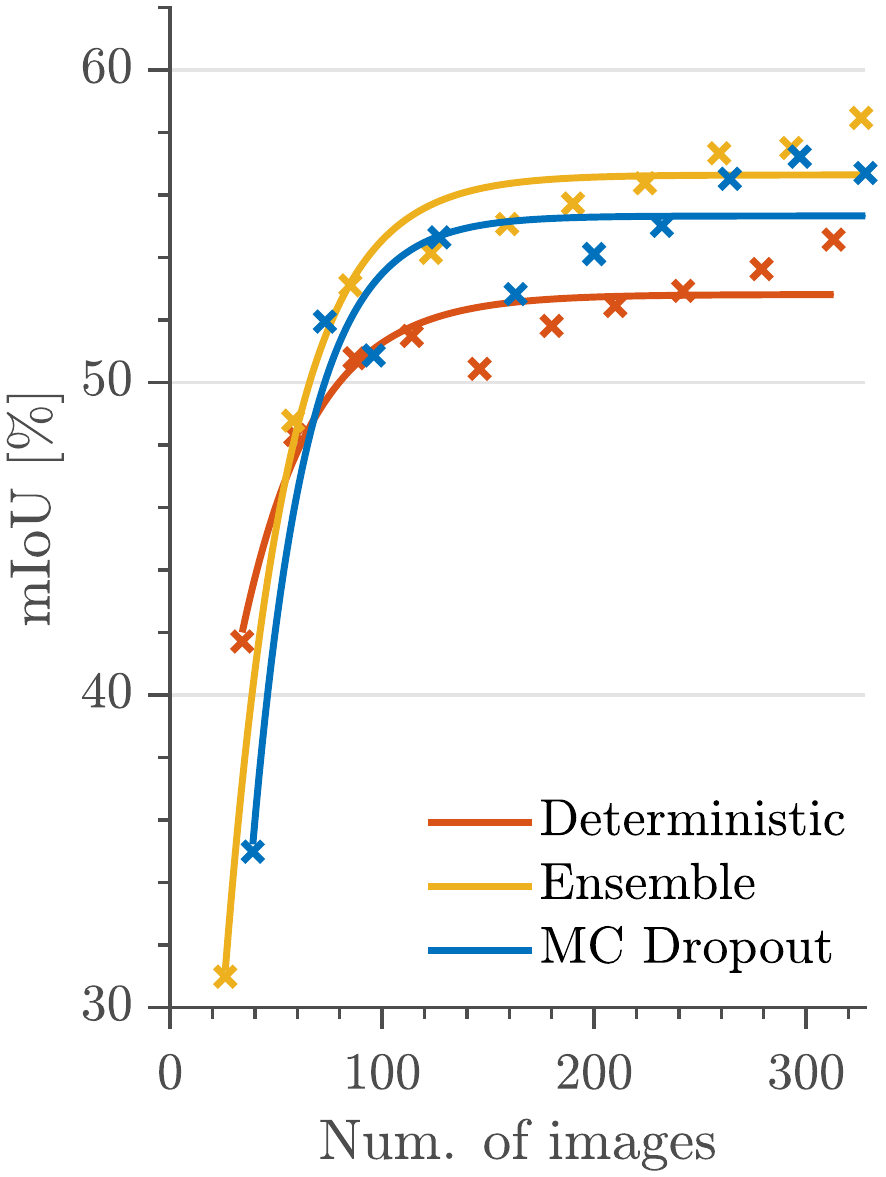}}

\caption{Comparison of \ac{AL} performance of uncertainty-based planning objectives on ISPRS Potsdam. Our Bayesian uncertainty-based objectives (blue, yellow) tend to perform better than the non-Bayesian entropy-based objective (orange). The Bayesian ensemble (yellow) achieves the highest \ac{AL} performance across the planning strategies.}
\label{F:ablation_uncertainty}
\end{figure}

We replace the Bayesian model uncertainty, see \cref{eq:mutual_information}, with the entropy of a deterministic forward pass. For a fair comparison, the coverage baseline uses a deterministic forward pass as well. As shown in \cref{F:results_potsdam_deterministic_vs_baselines}, the optimisation-based, frontier-based and local planners outperform the baseline, while the sampling-based planner performs similarly to the baseline. In line with results for Bayesian model uncertainty-based objectives, the optimisation-based and frontier-based planners show high prediction performance with substantially fewer training images compared to the local planner.

\cref{F:ablation_uncertainty} shows the effect of non-Bayesian entropy-based (orange) and Bayesian model uncertainty-based planning objectives estimated by either \ac{MC} dropout (blue) or an ensemble (yellow) on the planners' performances. Particularly, the map-based planners achieve higher \ac{AL} performance using Bayesian model uncertainty-based objectives irrespective of the uncertainty estimation technique. Although the non-Bayesian objective yields competitive performance with multiple planners in early missions, generally, the Bayesian ensemble method leads to the best \ac{AL} results. This could be due to two reasons. First, the ensemble shows higher prediction power (\cref{F:results_ablation_ensemble}). Second, as suggested by our qualitative results (\cref{F:potsdam_deterministic_uncertainty_prediction_examples}), non-Bayesian uncertainty is weakly calibrated, which results in a less informative planning objective.

\begin{figure}[!t]
    \captionsetup[subfigure]{labelformat=empty}

    \centering
    \subfloat[]{\includegraphics[width=0.19\columnwidth]{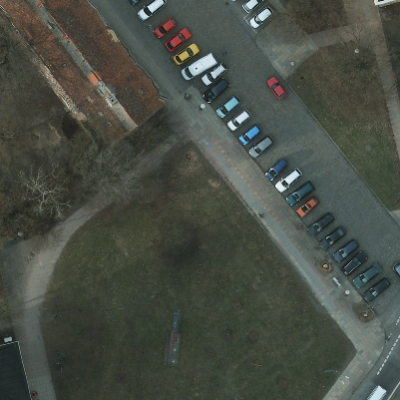}}
    \hfill
    \subfloat[]{\includegraphics[width=0.19\columnwidth]{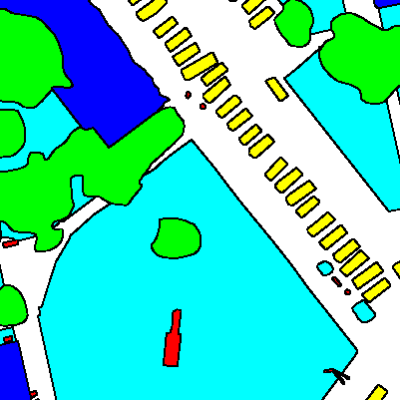}}
    \hfill
    \subfloat[]{\includegraphics[width=0.19\columnwidth]{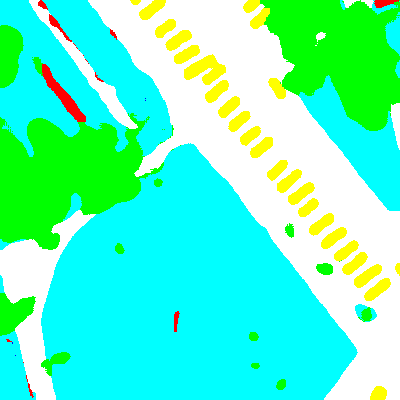}}
    \hfill
    \subfloat[]{\includegraphics[width=0.19\columnwidth]{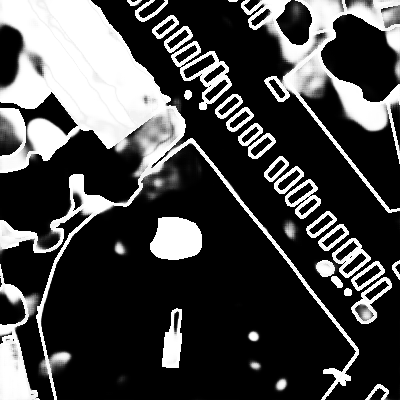}}
    \hfill
    \subfloat[]{\includegraphics[width=0.19\columnwidth]{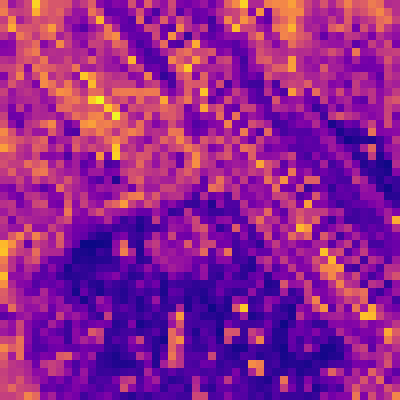}}

    \vspace{-7mm}
    \subfloat[]{\includegraphics[width=0.19\columnwidth]{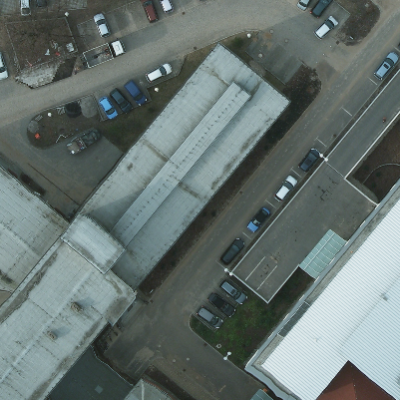}}
    \hfill
    \subfloat[]{\includegraphics[width=0.19\columnwidth]{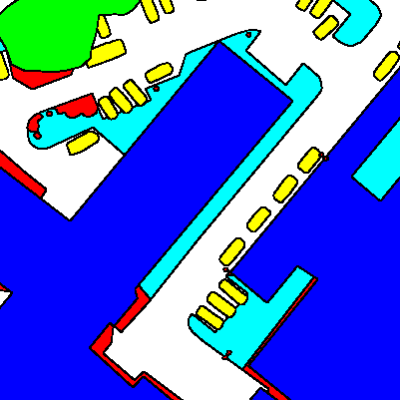}}
    \hfill
    \subfloat[]{\includegraphics[width=0.19\columnwidth]{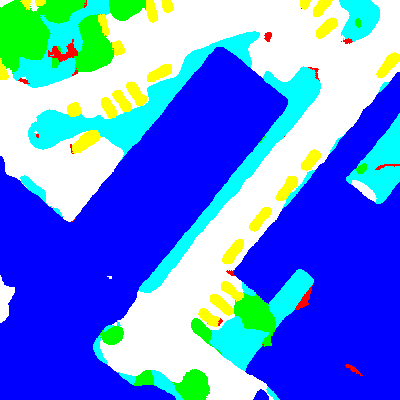}}
    \hfill
    \subfloat[]{\includegraphics[width=0.19\columnwidth]{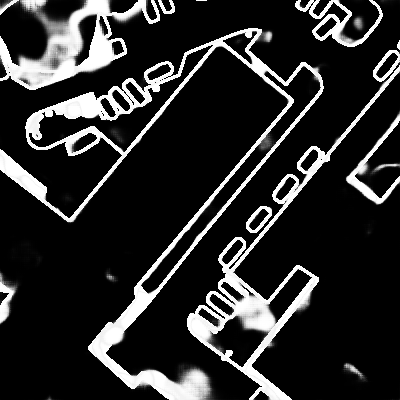}}
    \hfill
    \subfloat[]{\includegraphics[width=0.19\columnwidth]{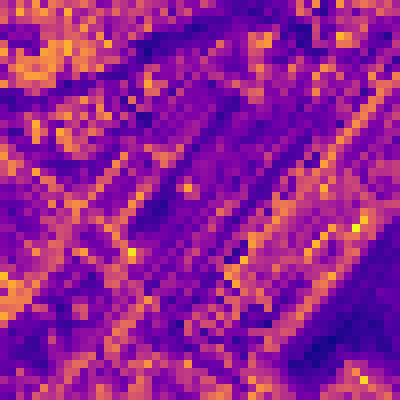}}

    \vspace{-7mm}
    \subfloat[Input]{\includegraphics[width=0.19\columnwidth]{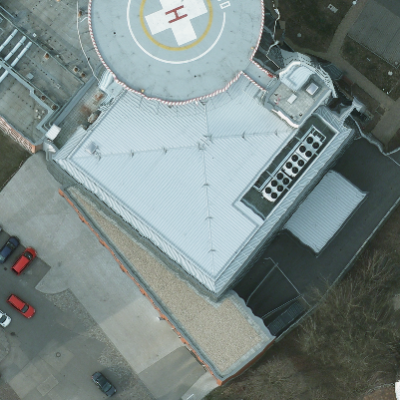}}
    \hfill
    \subfloat[Ground truth]{\includegraphics[width=0.19\columnwidth]{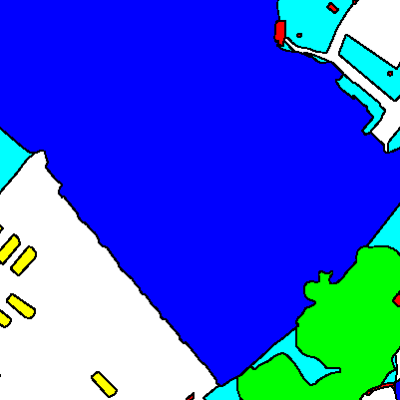}}
    \hfill
    \subfloat[Prediction]{\includegraphics[width=0.19\columnwidth]{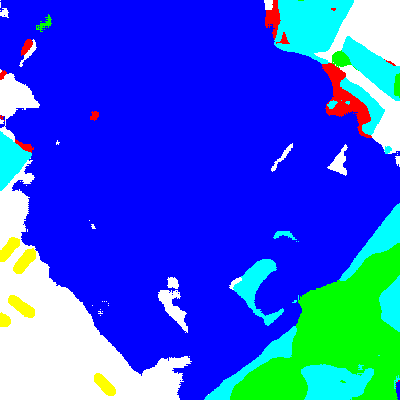}}
    \hfill
    \subfloat[Error]{\includegraphics[width=0.19\columnwidth]{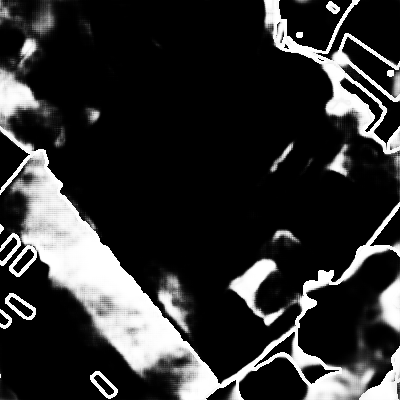}}
    \hfill
    \subfloat[Novelty]{\includegraphics[width=0.19\columnwidth]{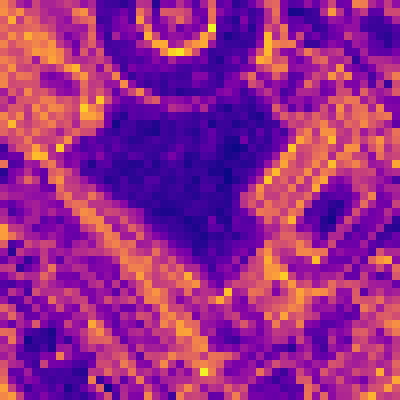}}

\caption{Qualitative results of a deterministic ERFNet trained on the ISPRS Potsdam dataset~\cite{Potsdam2018}. Columns from left to right: RGB input, ground truth, prediction, error image (negative prediction log-likelihood), representation novelty (\cref{eq:novelty_score}). High novelty scores (yellower) in case of rare visual cues, such as the helipad (bottom row), suggest that our representation-based objective provides useful information for \ac{AL} planning scenarios.}
\label{F:potsdam_representation_prediction_examples}
\end{figure}

To confirm that our framework is applicable to representation-based acquisition functions, we utilise the novelty score shown in \cref{eq:novelty_score} computed over the latent space of a deterministic ERFNet in our planning objective. For a fair assessment, we also utilise a deterministic ERFNet for the coverage baseline. Qualitatively, \cref{F:potsdam_representation_prediction_examples} visualises the representation novelties of a network trained and tested on disjoint areas of ISPRS Potsdam. Although the novelties do not correlate strongly with prediction errors (whiter), high novelty (yellower) is assigned to rare visual cues, such as the helipad (bottom row), which could be an informative objective to collect diverse training images.

\begin{figure}[!t]
    \includegraphics[width=\linewidth]{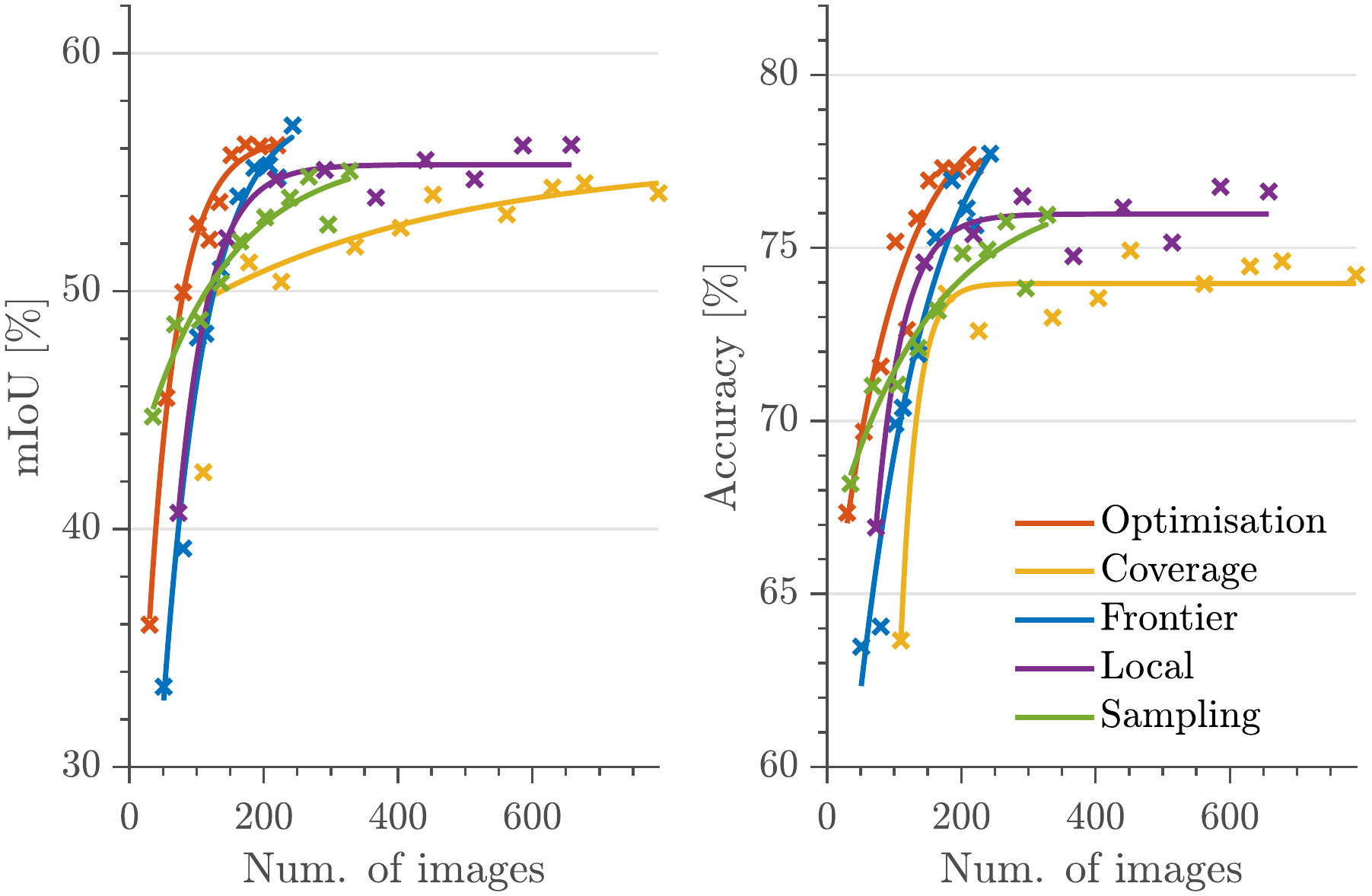}
    \caption{Comparison of \ac{AL} performance with representation-based novelty objective and computing informative prior maps before missions start. All planners outperform the baseline (yellow) with fewer training images.}
    \label{F:results_potsdam_representation_based_vs_baselines}
\end{figure}

\cref{F:results_potsdam_representation_based_vs_baselines} depicts the \ac{AL} results using representation novelties in the planning objective. All adaptive planners achieve higher segmentation performance than the coverage baseline (yellow). Further, our map-based optimisation (orange) and frontier (blue) planners require fewer training images than the local planner (purple) to reach high prediction performance. This validates that our framework generally supports various acquisition function paradigms and ensures higher \ac{AL} performance than the baseline approaches, irrespective of the planning objective. Our experiments suggest that the map-based planners outperform the local planner more significantly using Bayesian uncertainty objectives. This could be due to the better-calibrated Bayesian uncertainty estimates (\cref{F:results_ablation_ensemble}) leading to more informative planning objectives.

\subsection{Other Scenarios} \label{SS:rit18_and_flightmare_results}

The fourth set of experiments suggests that (i) our planning framework reduces the number of labelled images required to maximise segmentation performance across substantially different environments, and (ii) our global map-based planning strategies outperform state-of-the-art local planning in most cases, irrespective of the chosen planning objective.

We support these claims with an evaluation of our framework on the RIT-18 dataset~\cite{kemker2018algorithms} and in the Flightmare simulator~\cite{song2020flightmare}. The RIT-18 semantics cover large areas leading to challenging exploration. The Flightmare simulator resembles real-world \ac{UAV} control over an easy-to-explore photorealistic industrial terrain with strong random walk baseline performance. We access the framework's performance using the Bayesian model uncertainty estimated with \ac{MC} dropout, see \cref{eq:mutual_information}, and the representation novelty score given by \cref{eq:novelty_score}.

\begin{figure}[!t]
    \centering

    \subfloat[Bayesian model uncertainty-based objective \label{F:results_rit18_mc_dropout_vs_baselines}]{\includegraphics[width=\columnwidth]{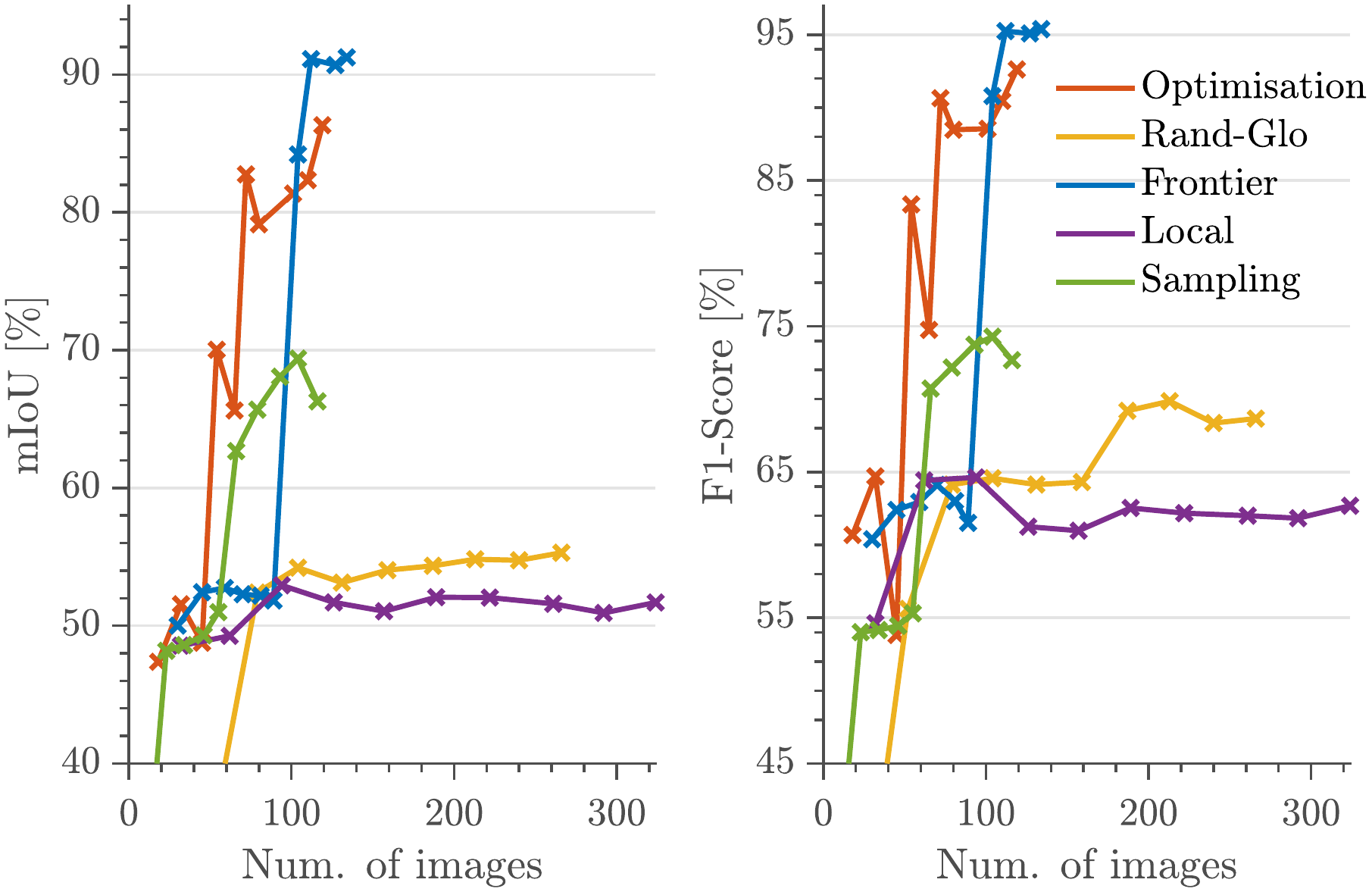}}

    \subfloat[Representation-based novelty objective \label{F:results_rit18_representation_vs_baselines}]{\includegraphics[width=\columnwidth]{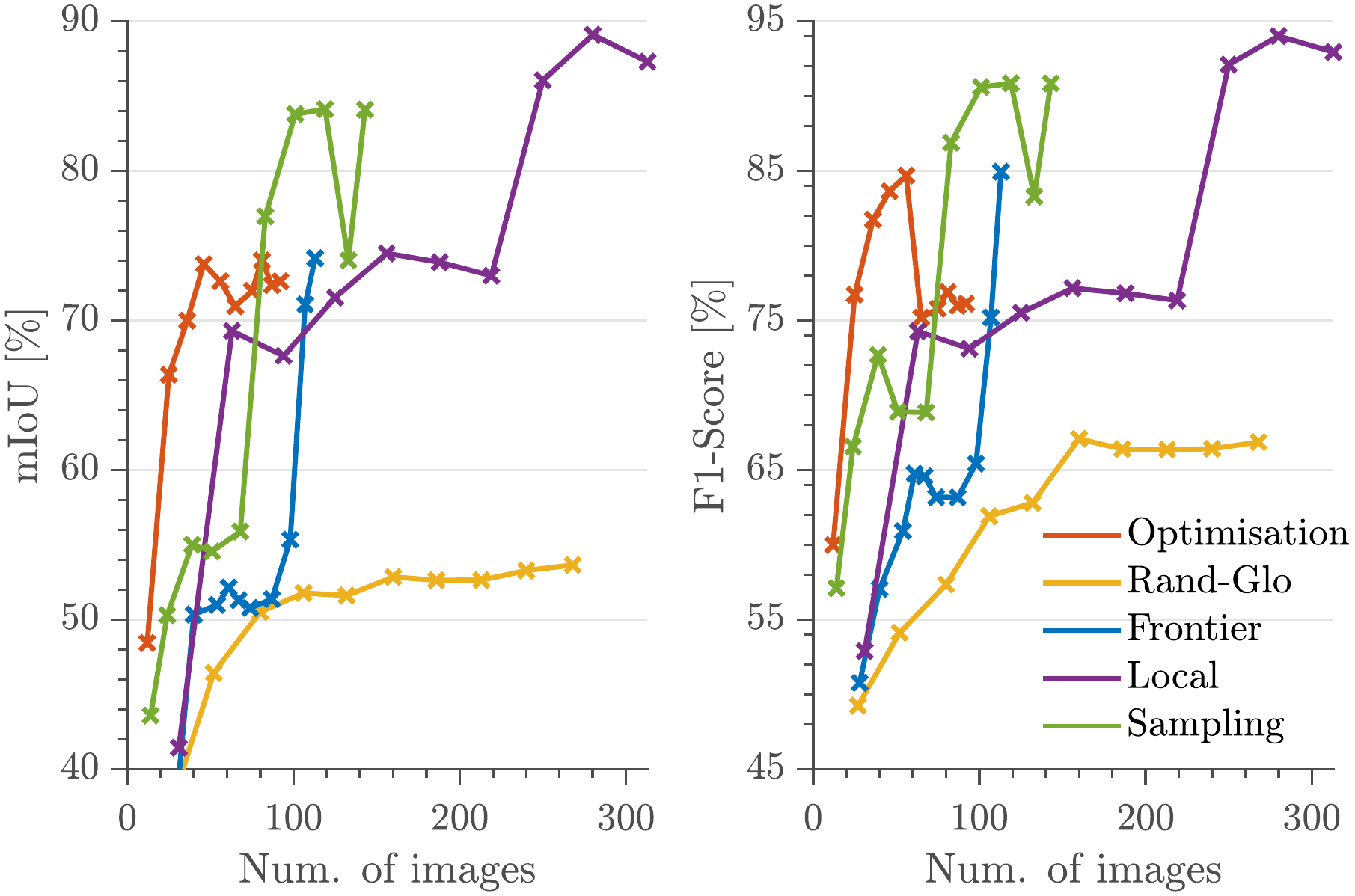}}

    \caption{\ac{AL} results on the RIT-18 dataset~\cite{kemker2018algorithms} using informative prior maps with (a) the Bayesian model uncertainty objective estimated by \ac{MC} dropout (\cref{eq:mutual_information}), and (b) the representation novelty objective (\cref{eq:novelty_score}). All map-based planners significantly outperform the random walk baseline (yellow). In most cases, our map-based planners lead to substantially higher \ac{AL} performance than the local planning strategy (purple).}
    \label{F:results_rit18_vs_baselines}
\end{figure}

\cref{F:results_rit18_vs_baselines} summarises our planning results on the RIT-18 dataset~\cite{kemker2018algorithms}. Note that non-monotonic model performance improvements on RIT-18 are expected as semantics cover large areas leading to challenging exploration influencing the training class distribution. All map-based planning strategies show significantly higher final segmentation performance than the random walk baseline (yellow), irrespective of the chosen planning objective. This confirms that our framework reduces human labelling effort while maximising segmentation performance over vastly differing terrains. Particularly, in most cases, our map-based planners require fewer training images to achieve segmentation performance on par or higher than the local planner (purple). Notably, the local planner performs worse than the baseline using Bayesian model uncertainty showing that our map-based planners are more generally applicable than the local planner.

\begin{figure}[!t]
    \centering

    \subfloat[Bayesian model uncertainty-based objective \label{F:results_flightmare_mc_dropout_vs_baselines}]{\includegraphics[width=\columnwidth]{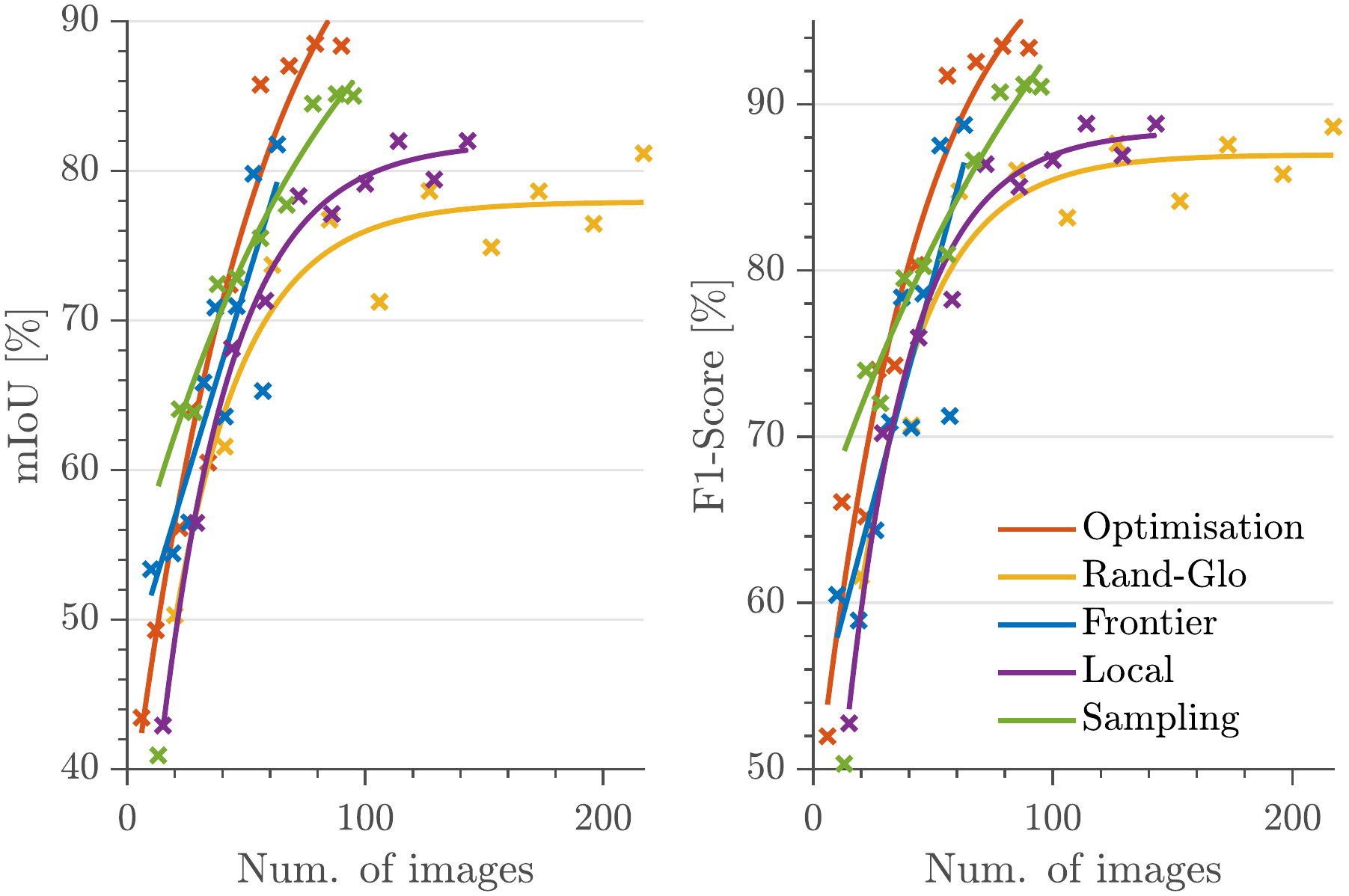}}

    \subfloat[Representation-based novelty objective \label{F:results_flightmare_representation_vs_baselines}]{\includegraphics[width=\columnwidth]{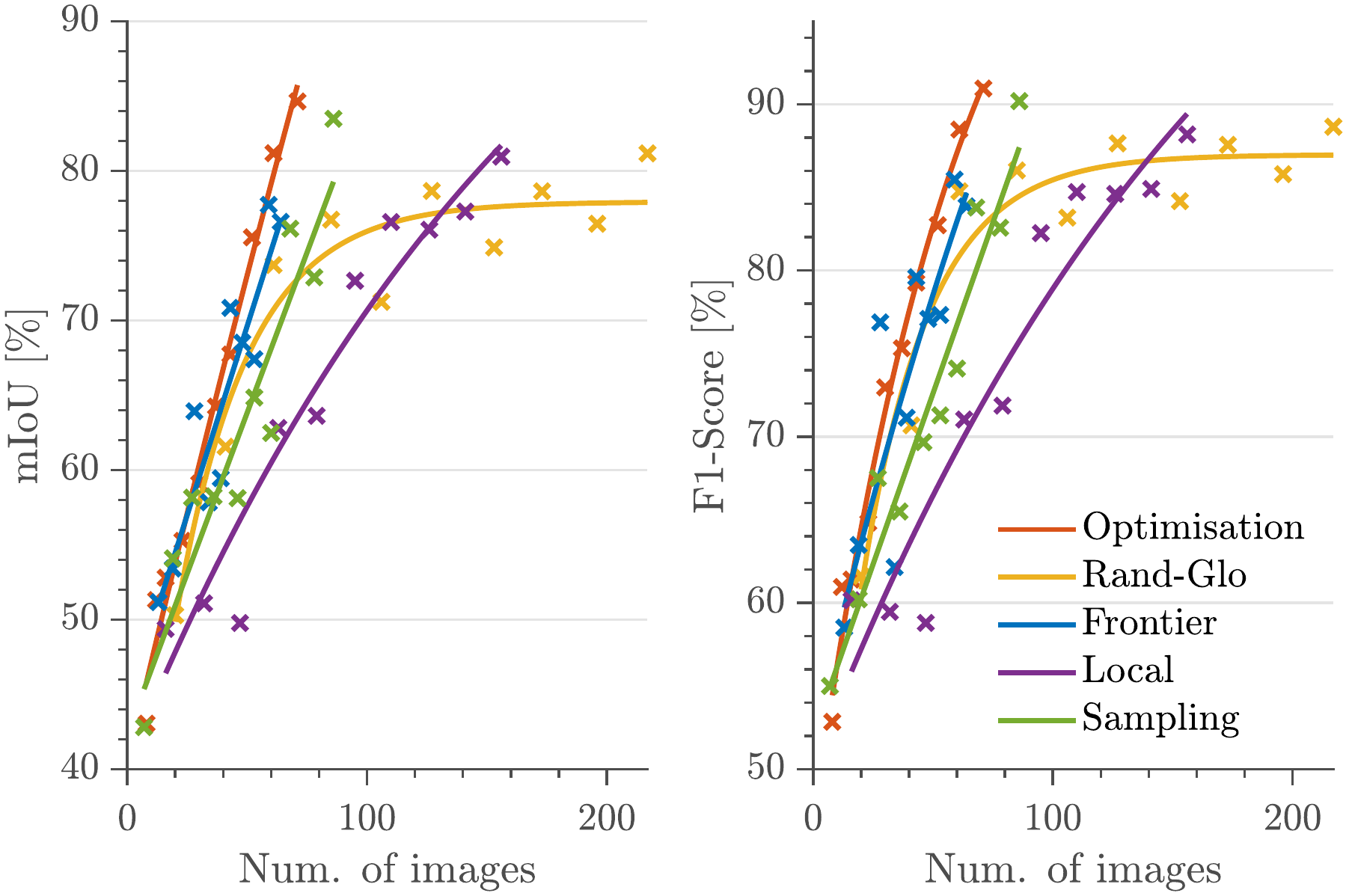}}

    \caption{\ac{AL} results in the Flightmare simulator~\cite{song2020flightmare} using informative prior maps with (a) the Bayesian model uncertainty objective (\cref{eq:mutual_information}), and (b) the representation novelty objective (\cref{eq:novelty_score}). All planners outperform the random walk baseline (yellow) using the Bayesian model uncertainty objective. Using the representation novelty objective, only our map-based optimisation and sampling planners show higher final prediction performance than the baseline.}
    \label{F:results_flightmare_vs_baselines}
\end{figure}

\cref{F:results_flightmare_vs_baselines} illustrates our planning results in the Flightmare simulator~\cite{song2020flightmare}. All planners using the Bayesian model uncertainty objective show higher \ac{AL} performance than the random walk baseline (yellow). Using the representation-based objective, only our two map-based optimisation (orange) and sampling (green) planners result in higher final prediction performance than the baseline. Combined with the RIT-18 results (\cref{F:results_rit18_vs_baselines}), this suggests that our Bayesian model uncertainty-based objectives are more robustly applicable across varying terrains compared to the representation novelty score proposed by Blum et al.~\cite{Blum2019}. One possible explanation could be that Bayesian model uncertainty is more strongly correlated with the prediction errors, as indicated by our qualitative results in \cref{F:potsdam_ensemble_prediction_examples} and \cref{F:potsdam_representation_prediction_examples}. In most cases, our map-based planners show higher \ac{AL} performance in both terrains than local planning. This verifies that our map-based planners are crucial for informative data collection, while local planning is not robustly applicable to varying terrains and planning objectives.

\subsection{Sensitivity Analysis} \label{SS:sensitivity_analysis}

The fifth set of experiments analyses our framework under various task-dependent design choices. It (i) verifies our framework's \ac{AL} performance with varying \ac{UAV} starting positions; (ii) validates our framework's robustness to different pre-training schemes; and (iii) showcases our framework's applicability and superior performance over baselines with different model architectures. The experiments are evaluated on the ISPRS Potsdam~\cite{Potsdam2018} and RIT-18~\cite{kemker2018algorithms} datasets using the Bayesian model uncertainty-based planning objective estimated by \ac{MC} dropout. If not stated otherwise, we utilise the Bayesian ERFNet (\cref{SS:al_acquisition_functions}) pre-trained on Cityscapes~\cite{cordts2016cityscapes}.

\begin{figure}[!t]
    \centering

    \subfloat[ISPRS Potsdam dataset \label{F:results_potsdam_informed_priors_vs_baselines_all_starting_positions}]{\includegraphics[width=\columnwidth]{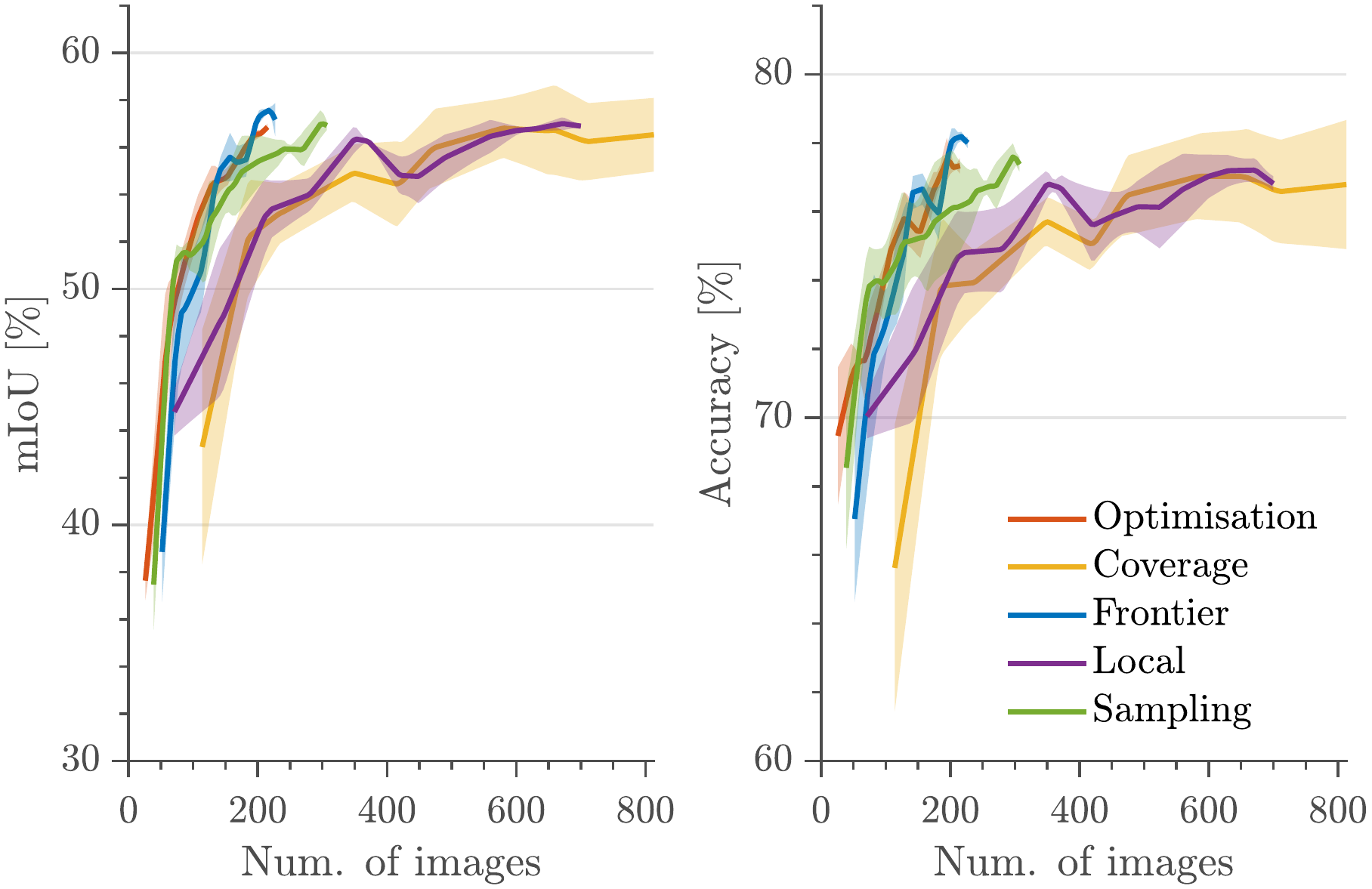}}

    \subfloat[RIT-18 dataset \label{F:results_rit18_informed_priors_vs_baselines_all_starting_positions}]{\includegraphics[width=\columnwidth]{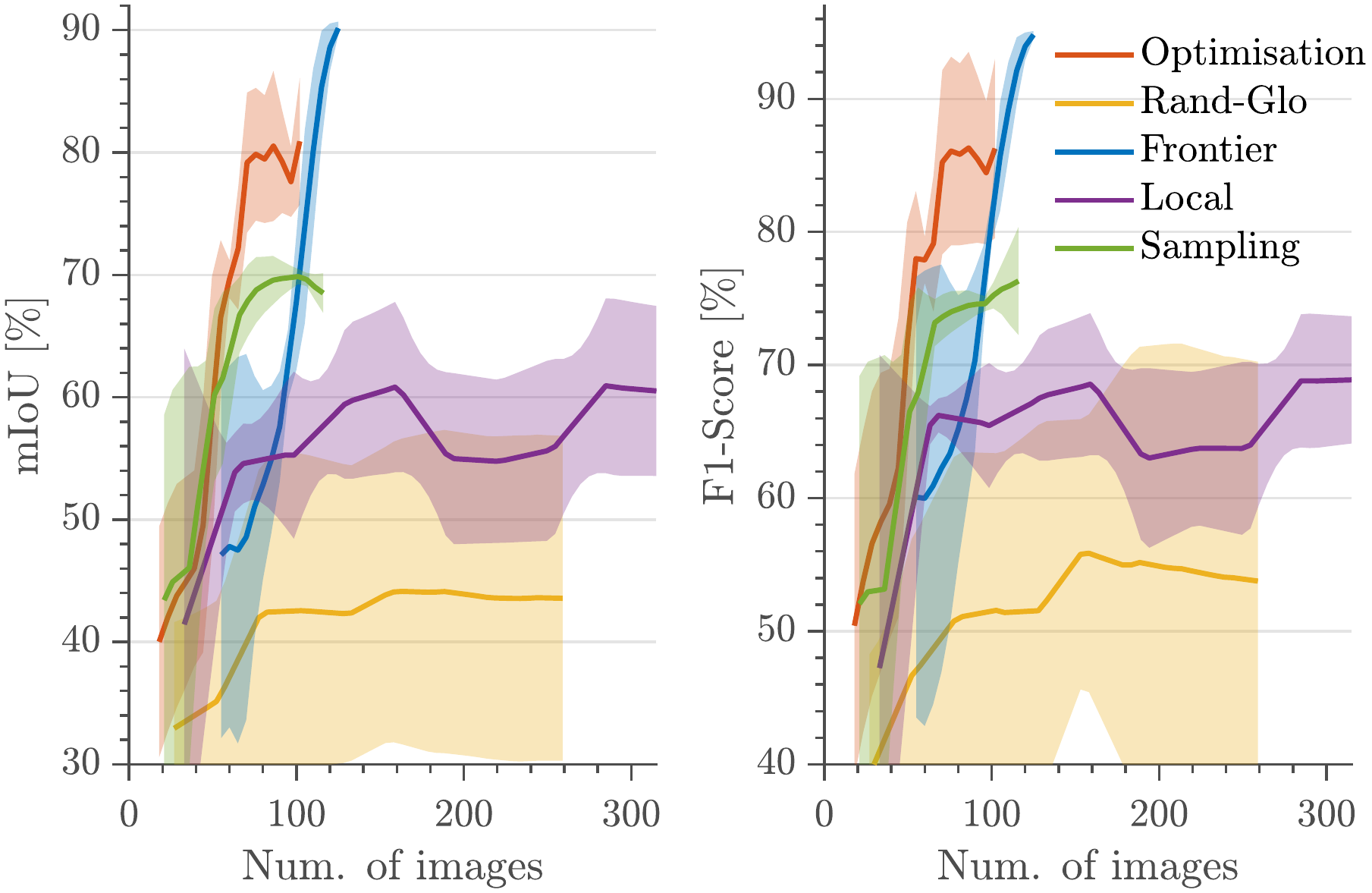}}
    
    \caption{Comparison of \ac{AL} performance on the (a) ISPRS Potsdam dataset~\cite{Potsdam2018} and (b) RIT-18 dataset~\cite{kemker2018algorithms} with the Bayesian model uncertainty-based planning objective estimated by \ac{MC} dropout and computing informative prior maps before each mission starts. Results are averaged over three different \ac{UAV} starting positions. Shaded regions indicate one standard deviation. Our map-based planners consistently outperform the baseline approaches (yellow) and local planning (purple) on both datasets with less training data while showing less sensitivity to the \ac{UAV} starting position.}
    \label{F:results_informed_priors_vs_baselines_all_starting_positions}
\end{figure}

\cref{F:results_informed_priors_vs_baselines_all_starting_positions} summarises the \ac{AL} performance for each planner averaged over three different starting positions at the top-left, top-right, and bottom-right corners of the ISPRS Potsdam and RIT-18 datasets. All our map-based planners, on average, reach higher \ac{AL} performance than the coverage baseline (yellow) and local planner (purple) on both datasets. In contrast, the local planner, on average, does not perform better than the coverage baseline on the ISPRS Potsdam dataset, as indicated by their largely overlapping means and standard deviations. Further, as indicated by the large standard deviations of the local planner and random walk baseline (yellow) on the RIT-18 dataset, the local planning and random walk \ac{AL} performances heavily depend on the \ac{UAV} starting position in challenging to explore terrains. This verifies that our map-based planners are robust to varying \ac{UAV} starting positions, while local planning and the baselines are sensitive to the \ac{UAV} starting position.

 \begin{figure}[!t]
    \centering

    \subfloat[ISPRS Potsdam dataset \label{F:results_potsdam_informed_priors_vs_baselines_all_checkpoints}]{\includegraphics[width=\columnwidth]{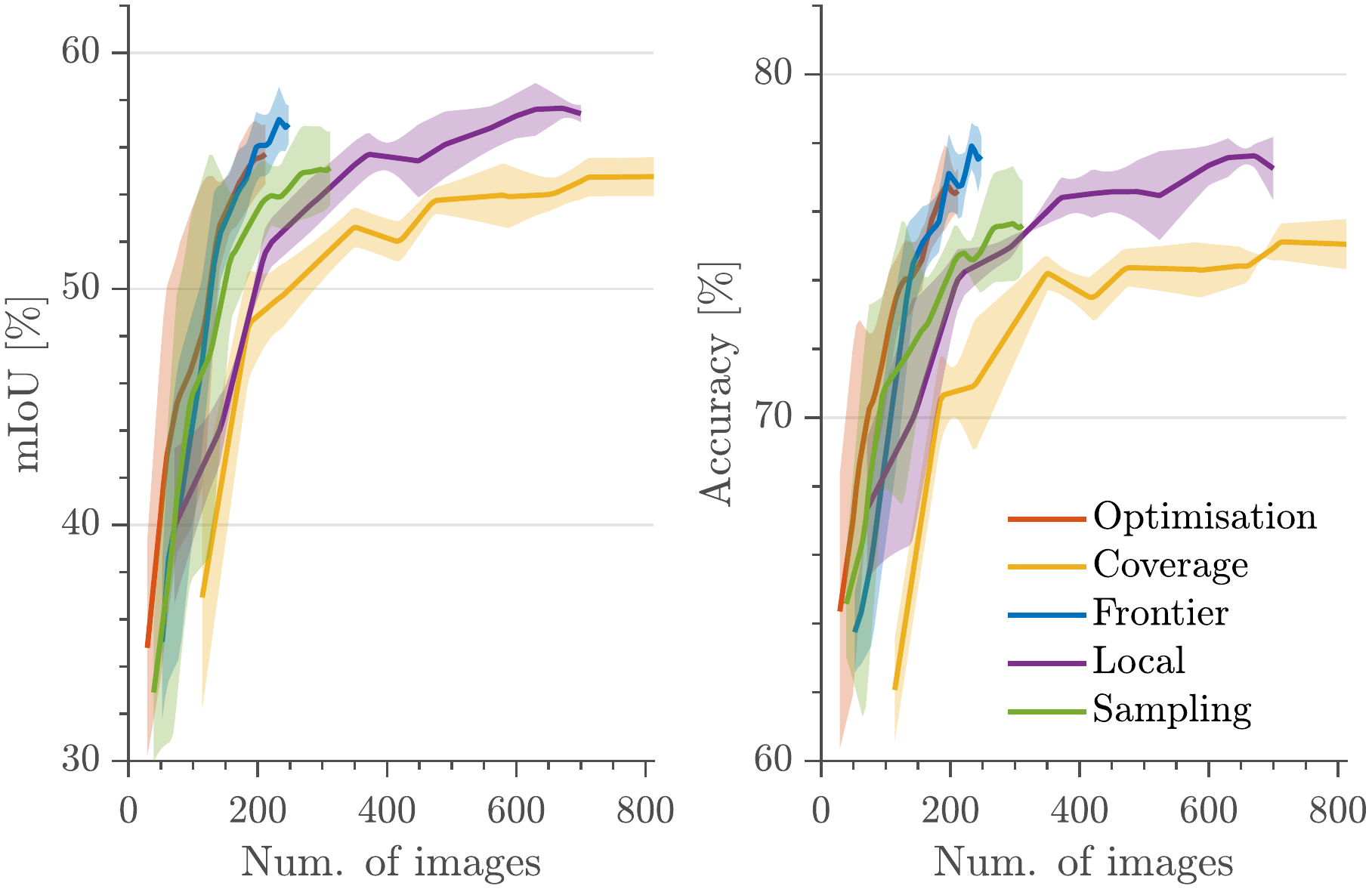}}

    \subfloat[RIT-18 dataset \label{F:results_rit18_informed_priors_vs_baselines_all_checkpoints}]{\includegraphics[width=\columnwidth]{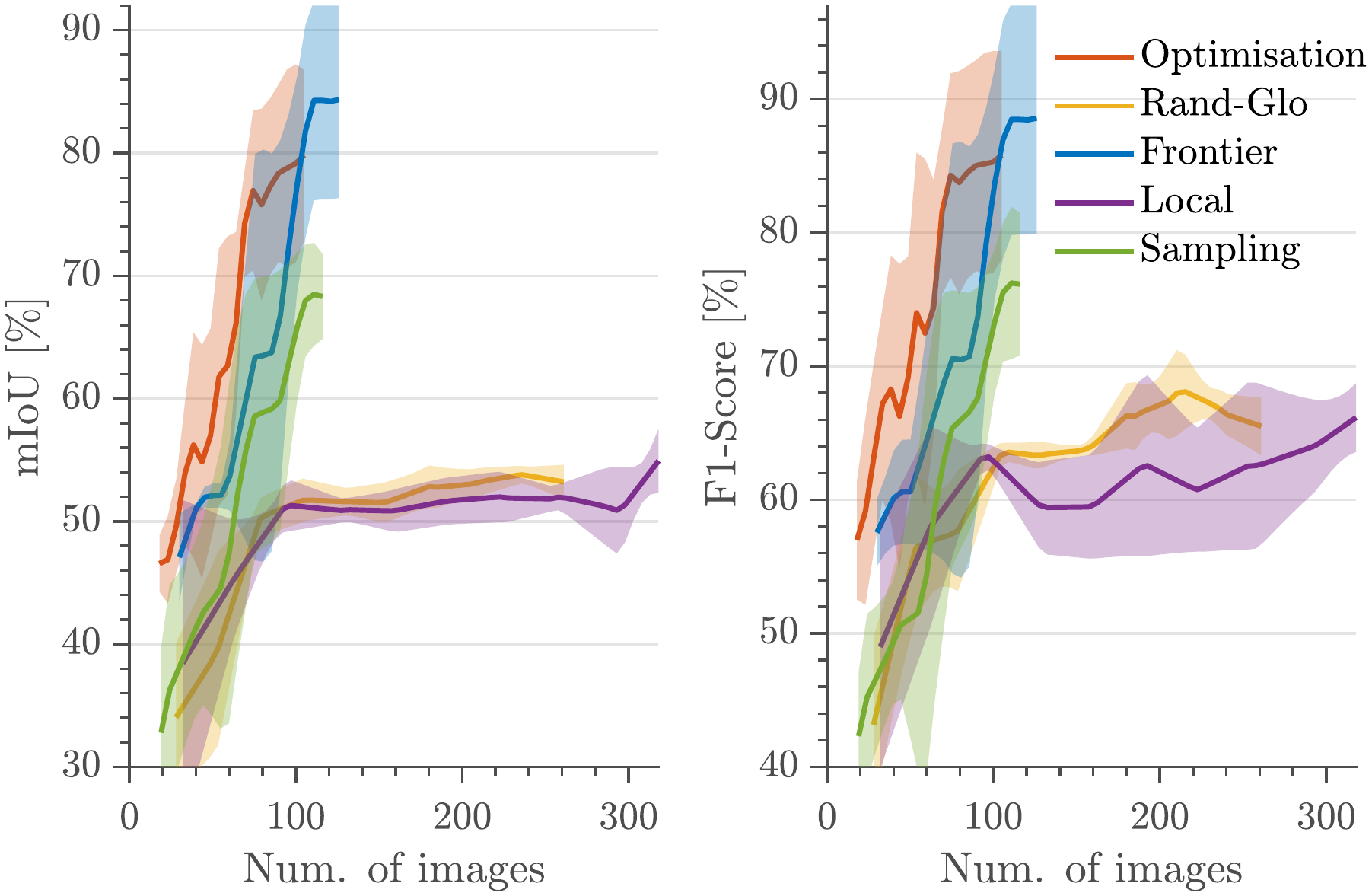}}

    \caption{Comparison of \ac{AL} performance on the (a) ISPRS Potsdam dataset~\cite{Potsdam2018} and (b) RIT-18 dataset~\cite{kemker2018algorithms} with the Bayesian model uncertainty-based planning objective estimated by \ac{MC} dropout and computing informative prior maps before each mission starts. Results are averaged over three differently pre-trained Bayesian ERFNets. Shaded regions indicate one standard deviation. Our map-based planners, on average, outperform the baseline approaches (yellow) and local planning (purple) on both datasets with less training data, irrespective of the pre-training scheme.}
    \label{F:results_informed_priors_vs_baselines_all_checkpoints}
\end{figure}

\cref{F:results_informed_priors_vs_baselines_all_checkpoints} summarises the \ac{AL} performance for each planner averaged over three differently pre-trained Bayesian ERFNets. Each mission starts from the top-left corner of the ISPRS Potsdam and RIT-18 datasets with Bayesian ERFNet being randomly initialised, pre-trained on the Cityscapes dataset~\cite{cordts2016cityscapes}, or pre-trained on the Flightmare dataset~\cite{song2020flightmare}. Note that the standard deviations are mainly a result of the randomly initilised models having, as expected, weaker prediction performance than the pre-trained models irrespective of the planning approach. All our map-based planners, on average, show stronger \ac{AL} performance than the baseline approaches (yellow) and the local planner (purple) on both datasets. Particularly, on the RIT-18 dataset, the local planner fails to outperform the random walk (yellow) irrespective of the pre-training scheme. These findings validate our map-based planners' robustness to varying model pre-training schemes.

\begin{figure}[!t]
    \includegraphics[width=\linewidth]{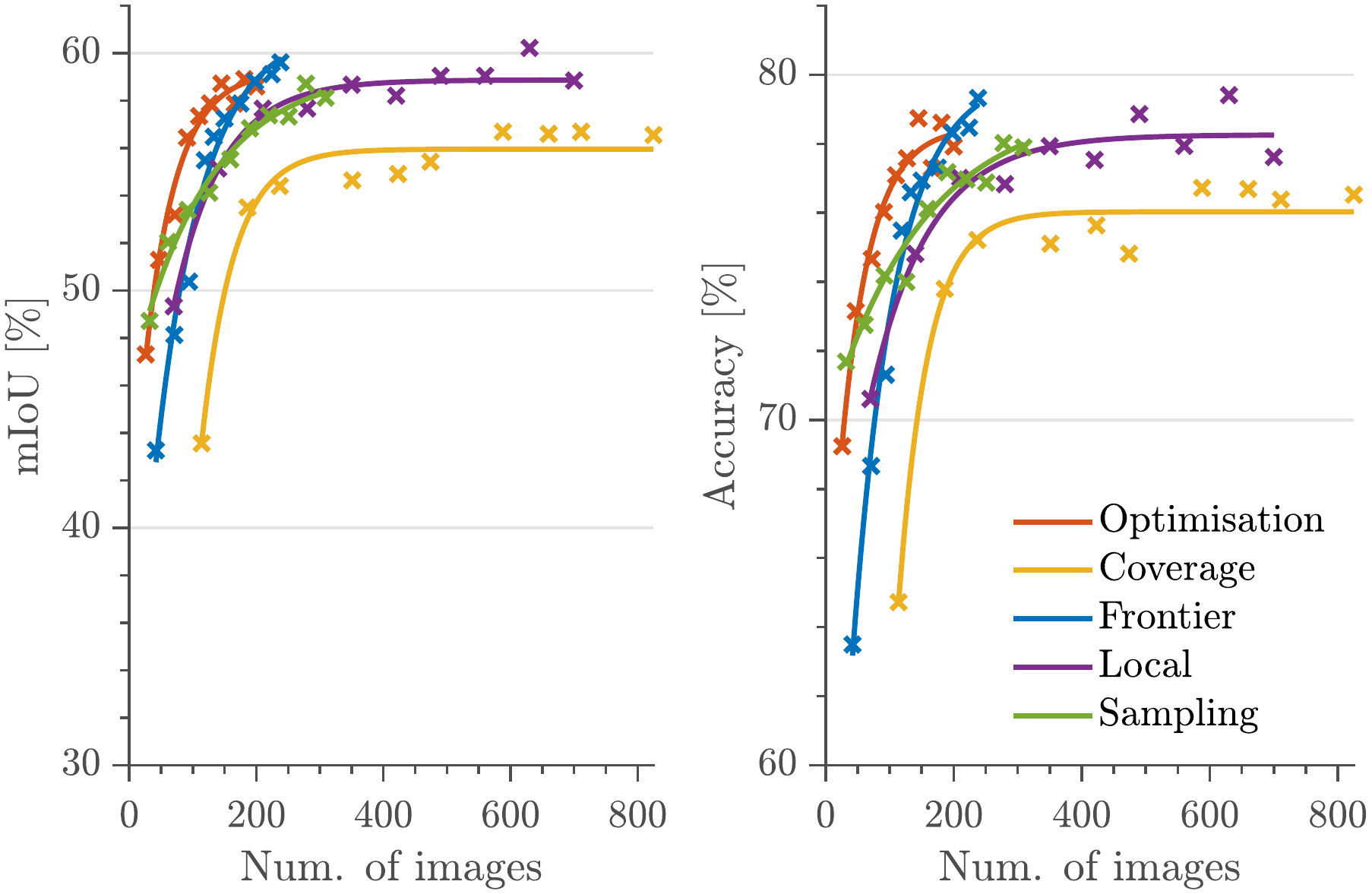}
    \caption{Comparison of \ac{AL} performance on the ISPRS Potsdam dataset~\cite{Potsdam2018} using a Bayesian version of U-Net~\cite{ronneberger2015u} pre-trained on the Flightmare dataset~\cite{song2020flightmare}.  The Bayesian model uncertainty-based planning objective is estimated by \ac{MC} dropout, and informative prior maps are computed before each mission starts. Our map-based optimisation (orange) and frontier (blue) planners outperform coverage (yellow) and local planning (purple), while the sampling planner (green) performs on par with local planning.}
    \label{F:results_potsdam_informed_priors_vs_baselines_unet}
\end{figure}

\cref{F:results_potsdam_informed_priors_vs_baselines_unet} summarises the \ac{AL} performance of our planning framework utilising a Bayesian variant of U-Net~\cite{ronneberger2015u}. We extend the U-Net architecture by adding dropout layers after each convolutional block with a dropout probability of $10\%$ to perform \ac{MC} dropout for computing the Bayesian uncertainty-based planning objective. We conduct experiments with the Bayesian U-Net pre-trained on the Flightmare dataset~\cite{song2020flightmare} using the ISPRS Potsdam dataset starting each mission from the top-left corner. All active planners exceed the maximum semantic segmentation performance of the coverage baseline (yellow) with less than half of the training images. This confirms the effectiveness of active planning for \ac{AL} irrespective of the chosen model architecture. Further, our map-based frontier (blue) and optimisation (orange) planners outperform local planning (purple), while the sampling planner (green) performs on par with local planning. This showcases strong \ac{AL} performance of our map-based planners and the applicability of our framework to different model architectures.

\section{Conclusion and Future Work} \label{S:conclusion}

This paper proposed a novel and unified planning framework for \acl{AL} in aerial semantic mapping to improve a robot's semantic perception with minimal expert guidance. A key aspect of our work is to link our planning objectives to \acl{AL} acquisition functions, enabling us to adaptively replan the robot's paths towards regions of informative training data. To ensure maximally informed online decision-making, our global planning algorithms leverage a sequentially updated probabilistic terrain map capturing semantics and acquisition function information. The framework is generally applicable to aerial robotic missions as it provides diverse acquisition functions, proposes various planning algorithms, is agnostic to the model architecture, and can be easily extended to other acquisition functions and planners.

Our experimental results show that our framework reduces the human labelling effort and maximises segmentation performance across varying terrains compared to traditionally used coverage and random walk data collection. Further, our map-based planners outperform state-of-the-art local planners used in \acl{AL}. The results also verify the benefit of our mapping module for the \acl{AL} performance. Overall, our findings demonstrate how active learning combined with online planning enables efficient training data collection to improve robotic perception in initially unknown environments.

Future work concerns integrating varying altitudes into the planning algorithms and estimating the resulting data uncertainty to select multi-view consistent informative training data. To further reduce human labelling effort, combining the supervised \ac{AL} paradigm with self-supervised training and continual learning across different terrains could be a promising avenue for future research.

\bibliographystyle{IEEEtran}
\footnotesize
\bibliography{2023-tro-rueckin}

\vspace{-11mm}
\begin{IEEEbiography}[{\includegraphics[width=1in,height=1.25in,clip,keepaspectratio]{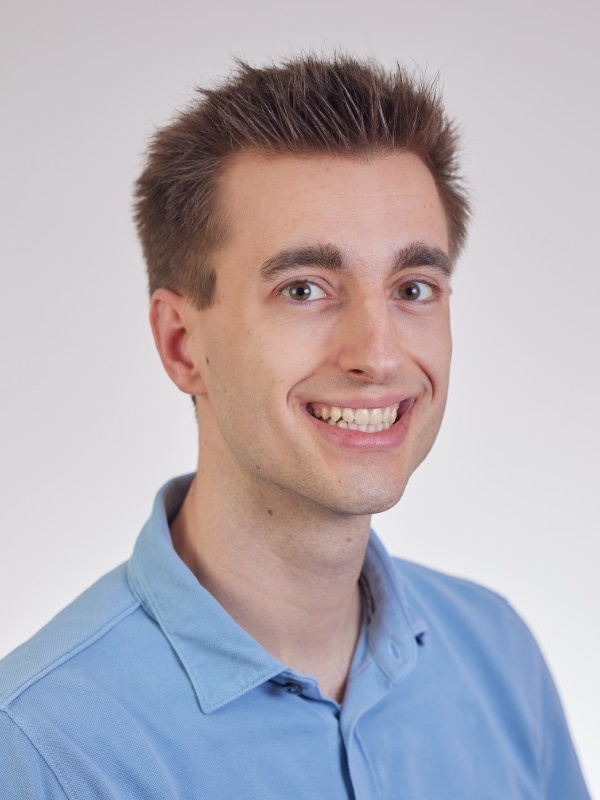}}]{Julius R\"{u}ckin} received a MSc degree in ”Mathematics in Data Science” from the Technical University of Munich in 2020. Previously, he received a B.Sc. degree in IT Systems Engineering from the University of Potsdam in 2018. He is currently a doctoral student in the Decision-Making for Autonomous Mobile Robots Group at the University of Bonn.  His research focuses on integrated planning and learning for autonomous robots at the intersection of computer vision, active learning and reinforcement learning.
\end{IEEEbiography}

\vspace{-11mm}
\begin{IEEEbiography}[{\includegraphics[width=1in,height=1.25in,clip,keepaspectratio]{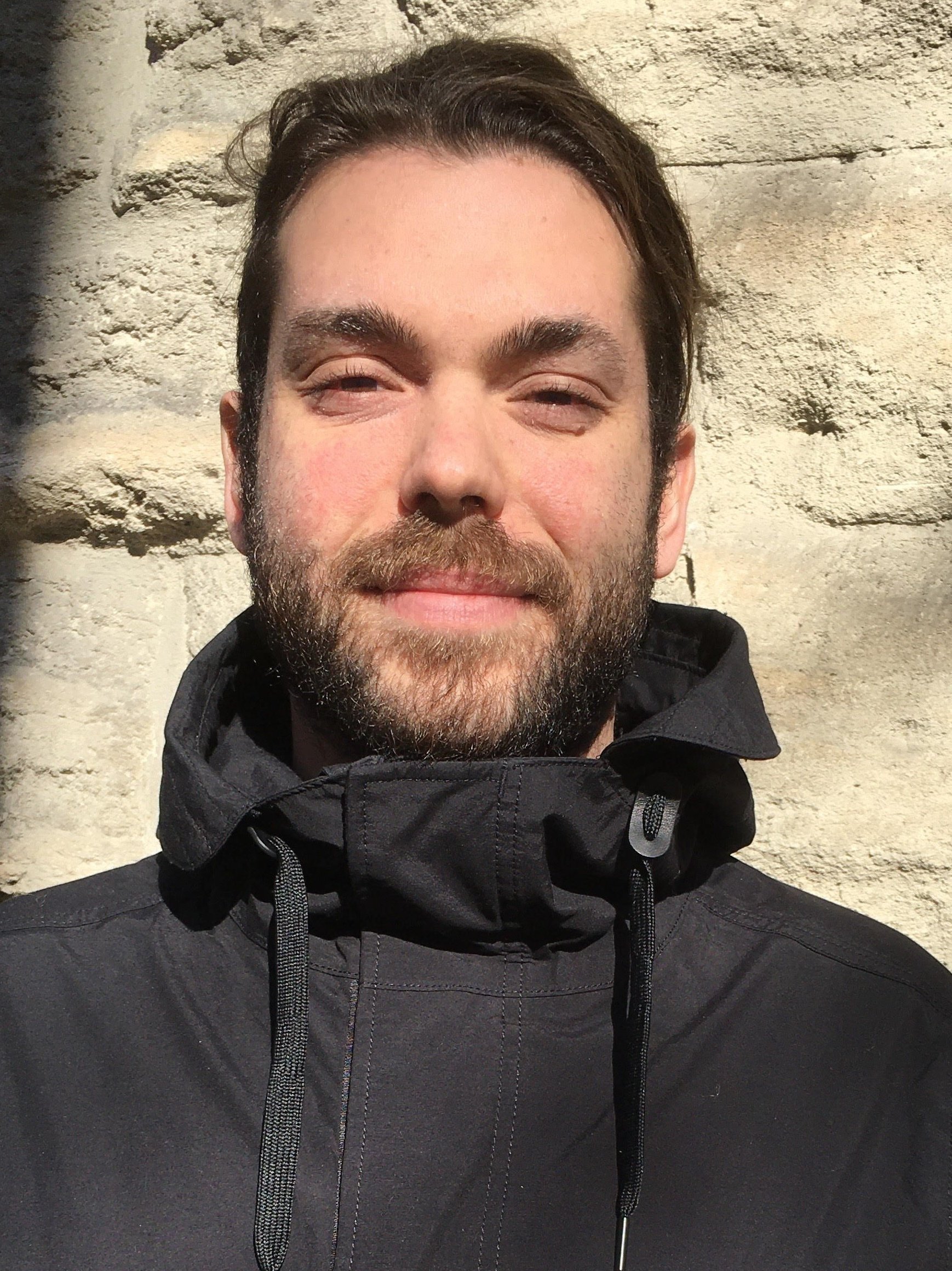}}]{Federico Magistri} is a Ph.D. student at the Photogrammetry Lab at the Rheinische Friedrich-Wilhelms-Universität Bonn since November 2019. He received his M.Sc. in Artificial Intelligence and Robotics from “La Sapienza” University of Rome with a thesis on Swarm Robotics for Precision Agriculture in collaboration with the National Research Council of Italy and the Wageningen University and Research. During his master, he spent one semester at the Albert-Ludwigs Universität Freiburg as an Erasmus student.
\end{IEEEbiography}

\vspace{-11mm}
\begin{IEEEbiography}[{\includegraphics[width=1in,height=1.25in,clip,keepaspectratio]{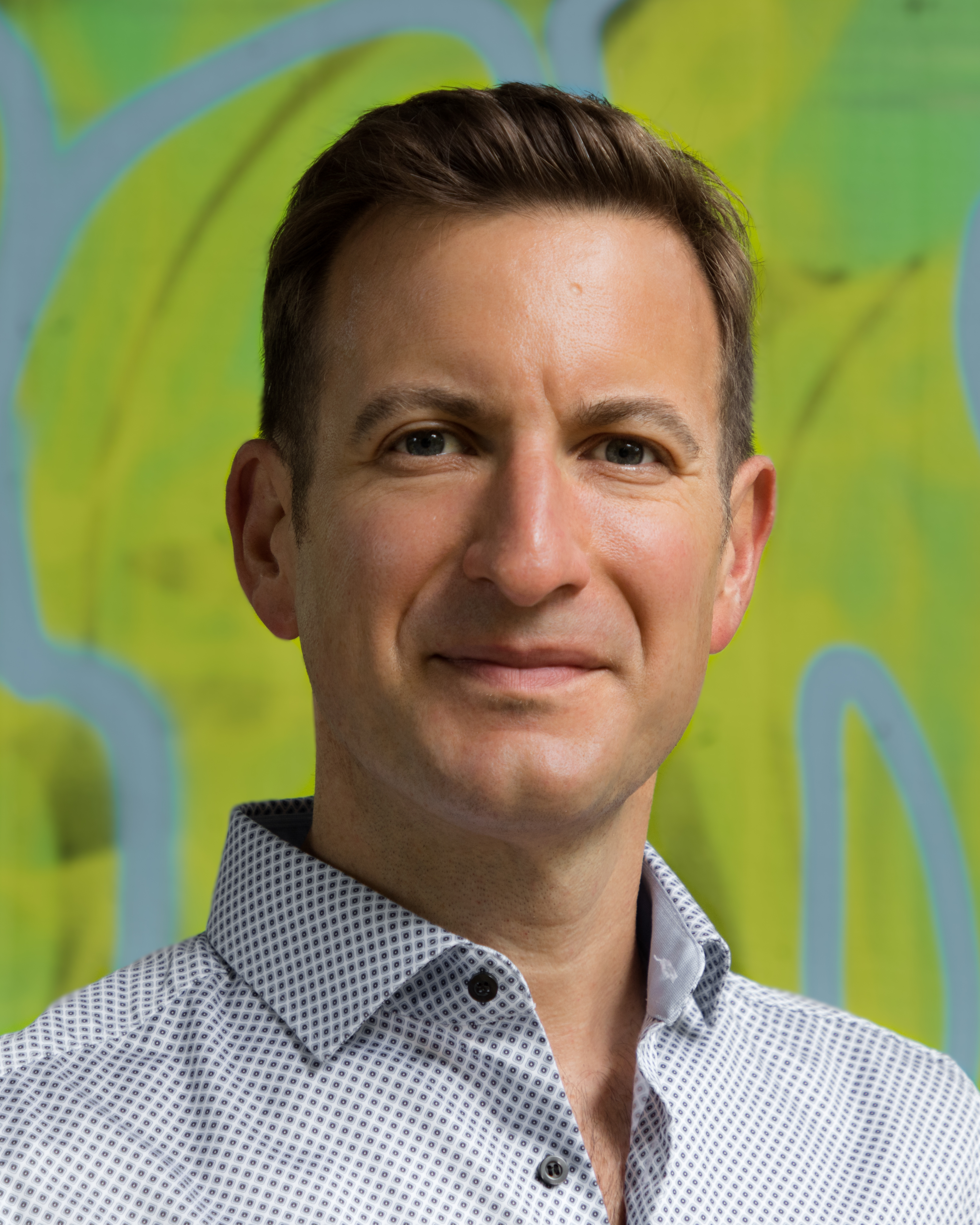}}]{Cyrill Stachniss} is a full professor at the University of Bonn, a Visiting Professor in Engineering at the University of Oxford, and is with the Lamarr Institute for Machine Learning and AI. He is the Spokesperson of the DFG Cluster of Excellence PhenoRob at the University of Bonn. Before his appointment in Bonn, he was with the University of Freiburg and ETH Zurich. His research focuses on probabilistic techniques and learning approaches for mobile robotics, perception, and navigation. The main application areas of his research are agricultural robotics, self-driving cars, and service robots. He has co-authored over 300 peer-reviewed publications and has coordinated multiple large-scale research projects on the national and European levels.
\end{IEEEbiography}

\vspace{-11mm}
\begin{IEEEbiography}[{\includegraphics[width=1in,height=1.25in,clip,keepaspectratio]{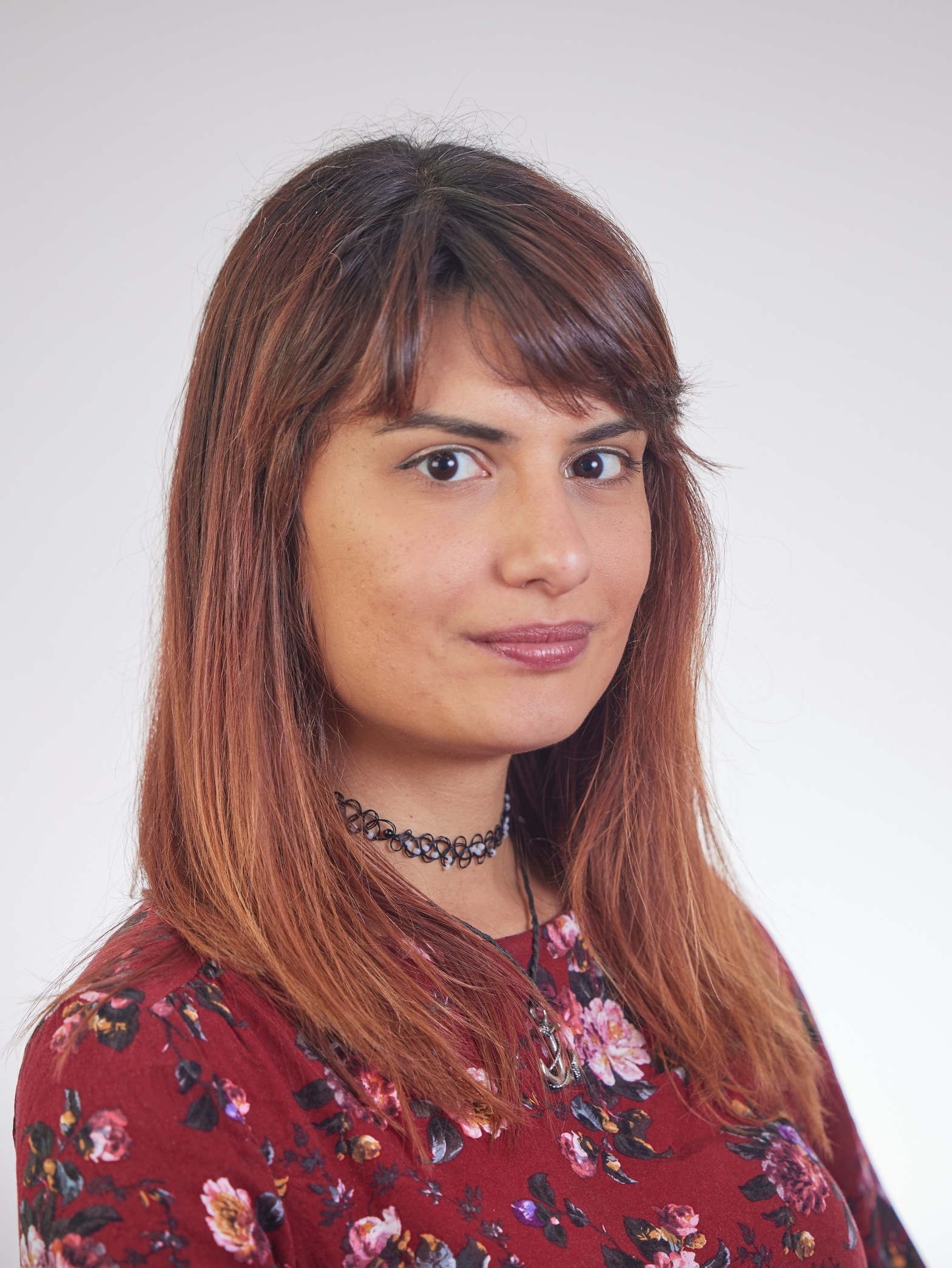}}]{Marija Popovi\'{c}} is a Junior Research Group Leader at the University of Bonn and Cluster of Excellence "PhenoRob". Before starting in Bonn, she was a postdoctoral research associate at the Smart Robotics Lab at Imperial College London, UK. She received her Ph.D. from the Autonomous Systems Lab at ETH Zurich, Switzerland (2019) and did her Master in Engineering in Integrated Mechanical \& Electrical Engineering at the University of Bath, UK (2015). Her research centres around developing algorithms that enable intelligent robotic decision-making, including in planning/coordination, environmental mapping, and computer vision.
\end{IEEEbiography}

\end{document}